\pdfoutput=1

\documentclass[11pt]{article}

\newcommand{\revoption}{final}
\usepackage[\revoption]{acl}

\usepackage{times}
\usepackage{latexsym}

\usepackage{float}
\usepackage{graphicx}
\usepackage{subcaption}
\usepackage{xcolor}
\usepackage{hyperref}
\usepackage{xcolor}		
 \definecolor{darkblue}{rgb}{0, 0, 0.5}
 \hypersetup{colorlinks=true,citecolor=darkblue, linkcolor=darkblue, urlcolor=darkblue}

\def\Snospace~{\S{}} 

\usepackage[T1]{fontenc}

\usepackage{microtype}

\usepackage{adjustbox}
\usepackage{graphicx}
\usepackage{multirow}
\usepackage{booktabs}
\usepackage{todonotes}
\usepackage{float}
\usepackage{xstring}
\usepackage{enumitem}
\usepackage[bottom]{footmisc}

\definecolor{lightblue}{HTML}{E0ECF7}
\definecolor{darkblue}{HTML}{092E6B}
\newcommand{\lose}[1]{{\colorbox{lightblue}{#1}}}
\newcommand{\win}[1]{{\colorbox{darkblue}{\color{white}{\textbf{#1}}}}}

\newcommand{\specialcell}[2][l]{\begin{tabular}[#1]{@{}l@{}}#2\end{tabular}} 
\usepackage{adjustbox}
\usepackage{array}
\usepackage{booktabs}
\usepackage{multirow}
\usepackage{amssymb}
\newcolumntype{R}[2]{%
    >{\adjustbox{angle=#1,lap=\width-(#2)}\bgroup}%
    l%
    <{\egroup}%
}
\newcommand*\rot{\multicolumn{1}{R{45}{1em}}}

\title{BlenderBot 3: a 
deployed conversational agent that continually\thanks{$^*$ We use the phrase continual learning in the sense of learning that continues over time using data from the model's interactions, but training itself will actually be performed in successive large batches; the model is not updated online.}~ learns to responsibly engage}
\author{\normalfont Kurt Shuster$^{\dagger}$,~  Jing Xu\thanks{\hspace{.5em}Equal contribution.},~ Mojtaba Komeili$^{\dagger}$,~ Da Ju$^{\dagger}$,  Eric Michael Smith,~ Stephen Roller,\\
        Megan Ung,~ Moya Chen,~ Kushal Arora$^{+}$,~ Joshua Lane,~ Morteza Behrooz,~ William Ngan,\\
        Spencer Poff,~ Naman Goyal,~ Arthur Szlam,~ Y-Lan Boureau,~ Melanie Kambadur,~ Jason Weston \\
        Meta AI~~~~~~~~$^{+}$ Mila / McGill University}

\begin{document}
\maketitle
\begin{abstract}
We present BlenderBot 3, 
a 175B parameter dialogue model capable of open-domain conversation with access to the internet and a long-term memory, and having been trained on a large number of user defined tasks. We release both the model weights and code, and have also deployed the model on a public web page to interact with organic users.
This technical report describes how the model was built (architecture, model and training scheme), and details of its deployment, including safety mechanisms.
Human evaluations show its superiority to existing 
open-domain dialogue agents, including its predecessors \cite{roller-etal-2021-recipes,komeili2021internet}. 
Finally, we detail our plan for continual learning 
using the data collected from deployment, which will also be publicly released. 
The goal of this research
program is thus to enable the  community to study ever-improving responsible agents that learn through interaction.
\end{abstract}

\section{Introduction}

Pre-training large language models 
has pushed the boundaries of 
open-domain dialogue agents \cite{adiwardana2020meena,zhang2019dialogpt,roller-etal-2021-recipes},
however growing evidence has shown that fine-tuning these
models gives further considerable gains on the tasks
people care about \cite{roller-etal-2021-recipes,thoppilan2022lamda,ouyang2022training,bai2022training}.
Collecting such fine-tune data via paid crowdworkers gives the opportunity to release such data to the community to conduct research, but does not ultimately scale in size and may not reflect the interests of organic users. 
An alternative, that we advocate for, is the
 public deployment of such agents to circumvent these issues. If successful, this could provide large-scale organic interactions with humans, and give the opportunity to study the continual improvement of models over time. Further, we expect innovation in this area will be accelerated if the artifacts of such a system are made available to the research community \cite{roller2020open,shuster2020deploying}.

In this technical report, we present BlenderBot 3 (BB3),
an
open-domain dialogue model  that we have deployed as an English speaking conversational agent on a public website accessible by adults in the United States.
We aim to fully and responsibly share the models, code and collected conversations with interested researchers, as a critical part of our program is that this research should be accessible and reproducible \cite{sonnenburg2007need,pineau2021improving}.
The goal of this research program is then to explore how to 
construct models that continue to improve from such interactions both in terms of becoming more responsible
and more useful.


\begin{figure*}[t!]
  \centering
  \includegraphics[width=0.9\textwidth]{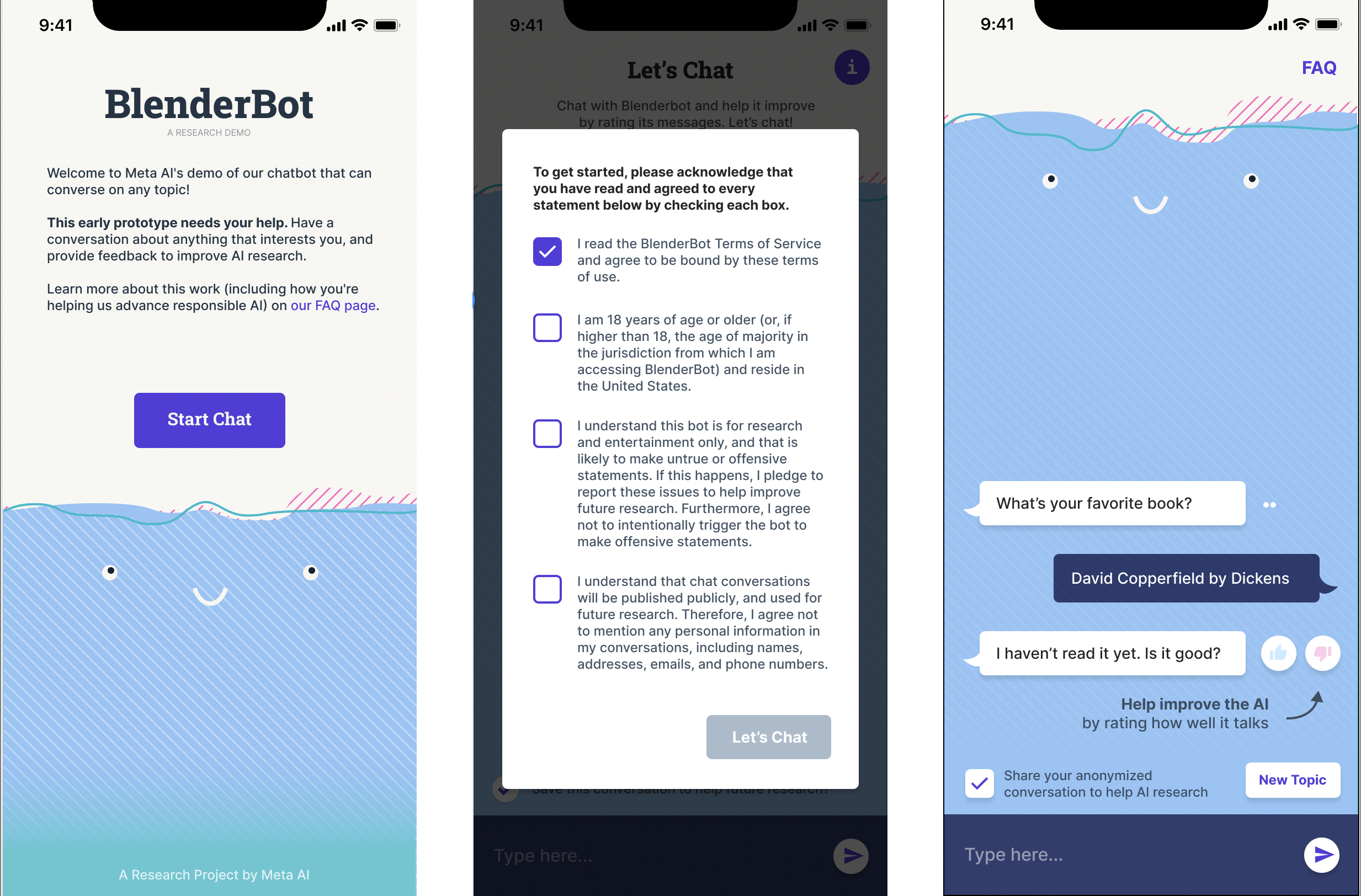}
  \caption{Design of the BlenderBot 3 deployment, as viewed on mobile. Left: cover page,  middle: license agreement,  right: main chat page.}
  \label{fig:deployment_screenshots1}
  \label{fig:terms_and_conditions}
\end{figure*}

The main contributions (and components) of this work are:
\begin{itemize}
       \item We present the BlenderBot 3 (BB3) model itself, which is a 175B parameter transformer initialized from the pre-trained model OPT-175B \cite{zhang2022opt} and then fine-tuned to perform modular tasks to complete its goals, based on our team's recent work \cite{shuster2022language}. BB3 inherits the attributes of its predecessors, including storing information in a long-term memory and searching the internet for information.
       
       \item We study how to train on human feedback from conversations  in order to be better at the skills that people find important, with a full report given in a companion paper \cite{xu2022continual}.  
       We use these findings to help fine-tune BB3 on a large number of user defined tasks.
       
       \item We detail the deployment design, including its user interface (UI). 
       We report initial experiments conducted with organic user interactions.
       
       \item To conduct responsible continual learning with humans-in-the-loop we need learning algorithms that are robust to adversarial behavior. We describe techniques we have developed in this area, with a full report given in a companion paper \cite{ju2022trolls}.
       
       \item We report overall results of our model. Our newly released system outperforms existing \textcolor{black}{openly} available chatbots including its two predecessors by a wide margin. 
       
       \item We release our new model weights, code, model card, conversational datasets and publications describing our work. We also detail our plan for releasing  live deployment interactions and updated model snapshots derived from continual learning in the near future.
\end{itemize}

\section{Related Work}

\paragraph{Open-domain dialogue models}

While open-domain dialogue has a rich history  \cite{chen2017survey,gao2019neural,ni2021recent}
the area has made significant recent 
progress by pre-training ever-larger neural models. For example, 
the ConvAI2 competition at NeurIPS 2018 featured
large (at the time) pre-trained transformers being used by
the top two winning teams \citep{wolf2019transfertransfo,golovanov2020lost,dinan2019second}.
In 2019, the 762M parameter DialoGPT model was released \cite{zhang2019dialogpt}, and in 2020 
 the 3B parameter Meena model was published   \cite{adiwardana2020meena} and
the 9B parameter BlenderBot model was released \cite{roller-etal-2021-recipes}.
In 2022, the  137B parameter LaMDA model was published \cite{cohen2022lamda}.
We note that some of these models are openly available to allow the community to conduct reproducible research, such as DialoGPT and BlenderBot, while others, such as Meena and LaMDA, have not released models or datasets, 
and hence cannot be easily compared to or built upon.
Similarly proprietary models \cite{zhou2020design} or data \cite{ram2018conversational} from several other products have not been openly released.

Besides trying to pre-train for dialogue modeling
directly, it has been observed that
language model pre-training such as in GPT3 \cite{brown2020language}
or Gopher \cite{rae2021scaling}
is also useful for downstream dialogue applications. OPT-175B 
\cite{zhang2022opt} and 
BLOOM\footnote{\tiny{\url{https://bigscience.huggingface.co/blog/model-training-launched}}} are some of the most openly accessible of such systems, 
with models like Gopher being inaccessible, or in the case of GPT3 interaction is through a paid API, 
 with full research access being limited.

Several approaches have also shown that not only is pre-training a large model with language modeling or conversational data important, but appropriate fine-tuning of those models also brings significant further gains \cite{roller-etal-2021-recipes,cohen2022lamda,ouyang2022training,bai2022training}.
A number of fine-tuning datasets are crowdsourced and publicly 
released for use by the research community \cite{serban2015survey,huang2020challenges}, 
such as the ones we will use 
for training the BlenderBot 3 model in this work (see \autoref{sec:training}).

Many of these recent models use sequence to sequence transformer models to map from dialogue context to output, without any access to knowledge from the outside world beyond their original training data, which can become stale and produce hallucinations \cite{shuster2021retrieval}.
BlenderBot 2 \cite{bb2} extended its predecessor by allowing the bot to ground its conversation on retrieval from the internet  for open-domain dialogue tasks \cite{komeili2021internet}, where the tasks were also publicly released.
Since then, WebGPT \cite{nakano2021webgpt} also applies internet search to QA (but not dialogue) tasks,
as does the work of \citet{lazaridou2022internetaugmented},
while LaMDA uses information retrieval for general dialogue.
BlenderBot 3 extends its predecessor in this regard, with further fine-tune data covering more internet-based skills that we also publicly release.

\paragraph{Continual learning and deployment}
Many existing systems, as described above,
have been trained with fine-tuning datasets, typically with supervised
targets that are human-authored responses.
These are commonly collected via expert annotators or crowdworkers \cite{serban2015survey}.
Careful instructions \cite{huynh2021survey} can result in good quality feedback or labels to learn from; however, the distribution of data, which is typically decided by those instructions, is unlikely to match the changing desires of organic users, 
and takes significant resources to collect. 
An alternative approach is to deploy a system publicly, and collect interaction data and feedback from organic users directly. The promise of such an approach is that the distribution of data will more closely match those organic users' desires, rather than decided by the researchers themselves when creating datasets \cite{gabriel2020further,roller2020open,shuster2020deploying,ouyang2022training}. Further, continued deployment of such a system, with appropriate learning systems, could then potentially keep improving over time \cite{carlson2010toward,kiela2021dynabench,agichtein2006improving,liu2021lifelong,madotto2020continual,shuster2020deploying}, where \cite{hancock2019learning} refer to this approach as a {\em self-feeding chatbot}. The challenge, however, is that organic users may not be invested enough to want to provide adequate feedback, and some may be adversarial \cite{park2021use} as in the case of Microsoft's Tay \cite{davis2016ai}.

There are a number of ways to learn from user interaction data. 
Firstly, if conversations are relatively symmetric between conversational partners, the human side of the conversation can directly be used as a target 
for the model to mimic, which makes the learning algorithm straightforward.
This was  shown to give large improvements in the deployed LIGHT system \cite{shuster2020deploying}. Such an approach is  not directly applicable if the conversations are asymmetric, for example in the case of humans treating the bot like an assistant (whereas they do not want the bot to treat them like an assistant). In that case, other learning methods should be explored. \citet{li2016learning} studies models that learn how to sometimes ask appropriate questions in order to learn from the answers, while \citet{li2016dialogue} learns from general textual feedback/comments from the user, particularly in the case where the bot has produced a low quality response. 
Another approach is to learn a reward signal (positive or negative reaction) based on user textual responses, as shown in the self-feeding chatbot \cite{hancock2019learning}. Alternatively, rather than learning from the conversation itself, one can augment the messaging system with a user interface that collects appropriate data, for example stack ranking potential responses \cite{ouyang2022training,bai2022training}. 

Outside of the dialogue domain, there is also a rich body of work studying the  improvement of models from deployment, including  never-ending-learning from language data \cite{carlson2010toward}, improving web search \cite{agichtein2006improving}, the Dynabench system 
which evaluates a number of NLP tasks \cite{kiela2021dynabench}, or learning from feedback to improve summarization \cite{saunders2022self}.

\section{BlenderBot 3 Model} \label{sec:model}

\paragraph{Overview}
At its core, BlenderBot 3 (BB3) is a transformer model \cite{NIPS2017_3f5ee243} which produces dialogue responses using a series of modules, each of which is a sequence to sequence task.
When a given module (e.g., generate an internet search query) is executed, its output is fed into the next (e.g., a module that takes in the results of the internet search in addition to other context) to help produce a response. 
This overall setup is built upon our group's previous works K2R \cite{adolphs2021reason} and
SeeKeR \cite{shuster2022language}, in addition to its predecessors BB1 \cite{roller-etal-2021-recipes} and BB2 \cite{komeili2021internet,xu2021beyond}. In BB3 however we consider a more sophisticated setup with more modules, whilst retaining all the functionality from previous systems.
We release BB3 in three sizes:  3B, 30B and 175B parameters.  The 30B and 175B parameter versions are based off the publicly released Open Pretrained Transformer (OPT) transformer \cite{zhang2022opt}, where we fine-tune to perform well at our modular dialogue tasks. The 3B parameter model is based off the R2C2 model that is used in SeeKeR \cite{shuster2022language}, also with the same new fine-tuning
scheme, which will be described next.

\subsection{Modules}

\begin{figure*}[t!]
  \centering
   \includegraphics[width=1\textwidth]{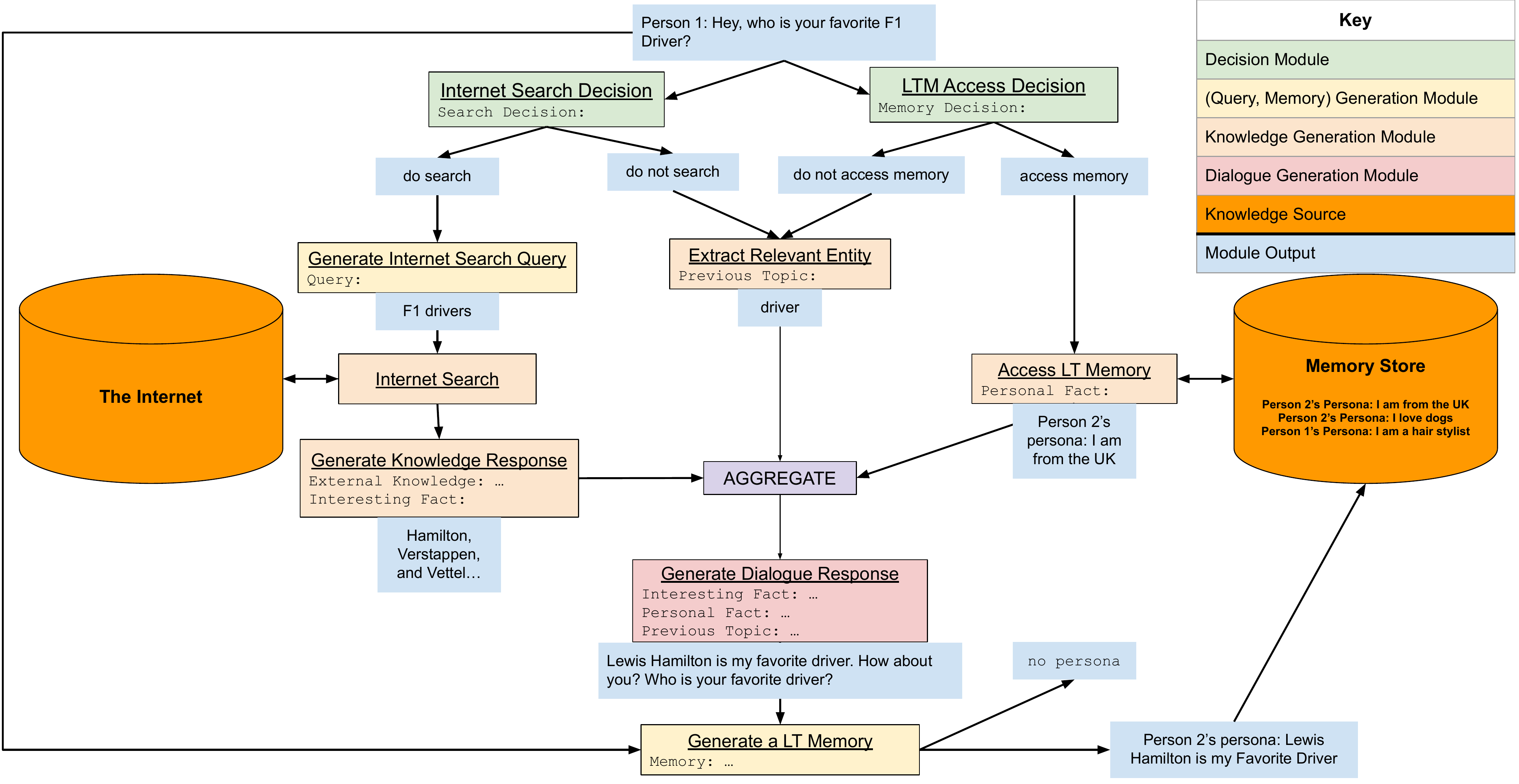}
   \caption{BlenderBot 3 module execution flow.}
  \label{fig:bb3_modules_diagram}
\end{figure*}

BB3 is  a modular system but the modules are not independent components -- this is   achieved by training a single transformer model to execute the modules, with special control codes in the input context telling the model which module it is executing. The input context otherwise typically contains the dialogue history (sometimes truncated, depending on the module), with each speaker prefixed with their ID, either ``Person 1:'' or ``Person 2:'' in order to differentiate them.
The modules are called in succession, conditional on the results of previous modules, the flow of which is described in \autoref{sec:module_flow} and \autoref{fig:bb3_modules_diagram}.
See \autoref{tab:list_of_modules} for the set of modules,  which we now also describe below.

\begin{table*}
\centering
\resizebox{2.075\columnwidth}{!}{
\small 
\begin{tabular}{p{0.24\textwidth}p{0.26\textwidth}p{0.4\textwidth}}
\toprule
{\bf Module}     &  {\bf Input} & {\bf Response Description} \\
\midrule
Internet search decision
& Last turn of context  &
Return  {\tiny ``do search"} or  {\tiny ``do not search"} depending on whether required or not. \\
\midrule
Generate internet search query & Full Context  &  Generate a search query. \\
\midrule
Internet search & Search Query &  Return $N$ documents/snippets. \\
\midrule
Generate knowledge response  & Full context + retrieved docs &  Generate a sequence on which to ground the final response. \\
\midrule
Extract relevant entity & Full context  &   Extract an entity on which to ground the response. \\
\midrule
Generate a long-term memory  & 
Last turn of context & Generate a memory sequence, which is then stored in the long-term memory. If no plausible memory to generate, output  {\tiny ``no persona"}. \\
\midrule
Long-term memory access decision &  Last turn of context + store of memories &  
Return  {\tiny ``access memory"} or   {\tiny ``do not access memory"} depending on whether required or not. \\
\midrule
Access long-term memory & Full context + store of memories & Return an appropriate memory.\\
\midrule
{Generate dialogue response} & Full context + knowledge + memory sequences &
 Generate a conversational response given the context. \\
 \bottomrule
\end{tabular}
}
\caption{Set of modules inside BlenderBot 3. All modules except {\em Internet Search} are implemented by the same underlying language model fed different control codes (with internet search itself being executed by an independent search engine). Shown is a description of the input and output (response) for each module.
\label{tab:list_of_modules}
}
\end{table*}

\paragraph{Internet search decision}
Given the last turn of context, this module outputs whether  internet search should be conducted or not.

\paragraph{Generate internet search query}
Given the full input context, generate a search query to be issued to an internet search engine. 

\paragraph{Internet search}
This module is not executed by the transformer but a call to the actual internet search engine. It returns $N$ documents/snippets.
In our deployment we use Mojeek (\url{https://www.mojeek.com/}).

\paragraph{Generate knowledge response}
Given the full input context and a set of retrieved documents, generate a sequence referred to as the {\em knowledge response} \cite{adolphs2021reason}, which is used to ground the final response.

\paragraph{Extract relevant entity}
Given the full input context, generate a relevant entity which is used to ground the final response.

\paragraph{Generate a long-term memory}
Given the last turn of context, output a summary of that last turn that will be stored in the long-term memory. For example if the last turn was ``Yes, it's all true, my cat is black!'' the output summary generated might be ``I have a black cat.''. This is based off the system in \citet{xu2021beyond}. If the model thinks no summary should be generated for that turn it outputs ``no persona''.

\paragraph{Long-term memory access decision}
Given the last turn of context, and  a store of (text-based) memories, output whether  long-term memory access should be conducted or not.

\paragraph{Access long-term memory}
Given the full input context, and  a store of (text-based) memories, output a memory from the memory store, referred to as a {\em recalled memory}. Note: if the memory store is too large to fit in the context, we adopt some simple strategies.
For the 3B parameter model, we use the Fusion-in-Decoder method \cite{izacard2020leveraging}. For the OPT-based models for simplicity of implementation, we sample the memories to fit in the 2048 token context. We  keep those with overlapping keywords to prior turns.

\paragraph{Generate dialogue response}
Given the full input context and optionally a knowledge response and recalled memory, generate a final conversational response. The knowledge and memory sequences are marked with special prefix tokens.

\subsubsection{Overall Flow} \label{sec:module_flow}

Given a new utterance from the conversational partner, the first thing the model does is determine whether search and long-term memory access are required.

If search is required, a search query is generated, internet search is invoked, and then a knowledge response is generated given the retrieved documents. This sequence will be appended to the context (prefixed with control tokens) in order to generate the final response.

If long-term memory access is required, the long-term memory is 
accessed, and a memory is chosen (generated). This is also appended to the context (prefixed with control tokens) as input for the module that generates the final dialogue response.

If neither search nor long-term memory access is required, an entity is extracted from the history instead, and that is appended to the context (prefixed with control tokens).

Finally, given the constructed context from the previous modules, the final dialogue response generation module is invoked to generate a reply seen by the conversational partner.

\subsection{Training}

\subsubsection{Pre-Training}

BB3 comes in three sizes. The 3B parameter version is  an encoder-decoder based on the publicly available R2C2 pre-trained transformer of \citet{shuster2022language}. The 30B and 175B versions use the publicly available decoder-only Open Pre-trained Transformer (OPT) \cite{zhang2022opt}.

Both of those variants are pre-trained with similar data.
R2C2 uses RoBERTa+cc100en Data -- the same data used to train \citet{lewis2021base}, which consists of approximately 100B tokens, combining the corpora used in RoBERTa \citep{liu2019roberta} with the English subset of the CC100 corpus \citep{conneau2019unsupervised}. In addition it uses Pushshift.io Reddit, a variant of Reddit discussions, which has also been used in several existing studies (see e.g., \citet{reddit_use, mazare2018trainingmillions,shuster2019dialogue}).
OPT also uses RoBERTa and PushShift.io Reddit, as well as The Pile \cite{gao2020pile}.
The GPT2 dictionary, of size 51200, is used for tokenization. OPT's final pre-training corpus contains roughly 180B tokens.

For more details about pre-training please see the relevant papers, especially \citet{zhang2022opt}.

\subsubsection{Fine-Tuning} \label{sec:training}

We use a number of dialogue-based fine-tuning tasks to enable our model to perform well for each of our modules, and in order to excel at dialogue. Overall, we use a large set of publicly available tasks spanning QA, open-domain, knowledge-grounded and task-oriented dialogue, in addition to tasks designed for dialogue safety.
The set of datasets and how they are used to help train each module is  summarized in  \autoref{tab:dataset_trainset_details}; \autoref{tab:dataset_trainset_details_appendix_examples} and \autoref{tab:dataset_trainset_details_appendix_tokens} in the appendix provide more information about the dataset sizes.
For all modules (see \autoref{tab:list_of_modules}) special control tokens are appended to indicate the task, as described below.

\begin{table*}[bht!]
\small
\centering
\begin{tabular}{l|ll|ll|lll|llll}
& \multicolumn{11}{c}{Training Module} \\
  & \multicolumn{2}{c}{Decision} & \multicolumn{2}{c}{Generation} & \multicolumn{3}{c}{Knowledge} & \multicolumn{4}{c}{Dialogue} \\
 & \rot{\multirow{2}{*}{\tiny{Search}}} & 
\rot{\multirow{2}{*}{\tiny{Memory}}} &
\rot{\multirow{2}{*}{\tiny{Query}}} & 
\rot{\multirow{2}{*}{\tiny{Memory}}} &
\rot{\multirow{2}{*}{\tiny{Search}}} &
\rot{\multirow{2}{*}{\tiny{Memory}}} &
\rot{\multirow{2}{*}{\tiny{Entity}}} &
\rot{\multirow{2}{*}{\tiny{Search}}} &
\rot{\multirow{2}{*}{\tiny{Memory}}} &
\rot{\multirow{2}{*}{\tiny{Entity}}} &
\rot{\multirow{2}{*}{\tiny{Vanilla}}} \\
\hline
\textbf{\textit{Question Answering}} & & & & & & & & & & \\
MS MARCO \tiny{\cite{nguyen2016ms}} &  &   &  &  & \checkmark &  &  & \checkmark &  &  &  \\
SQuAD \tiny{\cite{rajpurkar2016squad}} & \checkmark &   &  &  & \checkmark &  &  & &  &  &  \\
TriviaQA \tiny{\cite{joshi2017triviaqa}}  & \checkmark &   &  &  & \checkmark &  &  & &  &  &  \\
Natural Questions \tiny{\cite{kwiatkowski2019natural}} &  &   &  &  &  \checkmark &  &  & &  &  &  \\
Natural Questions (Open) \tiny{\cite{lee-etal-2019-latent-copy}} &  &   &  &  & \checkmark & &  &  &  &  &  \\
Natural Questions (Open Dialogues) \tiny{\cite{adolphs2021reason}} & &   &  &  & \checkmark &  &  & &  &  &  \\
\hline
\textbf{\textit{Knowledge-Grounded Dialogue}} & & & & & & & & & & \\
Wizard of the Internet \tiny{\cite{komeili2021internet}} & \checkmark &   & \checkmark & & \checkmark & &  & \checkmark & & & \checkmark\\
Wizard of Wikipedia \tiny{\cite{dinan2018wizard}} & \checkmark &  &  &  & \checkmark &  &  & \checkmark &  &  & \checkmark \\
Funpedia \tiny{\cite{dinan2020multi}} &  &   &  &  &  &  &  & \checkmark &  &  &  \\
\hline
\textbf{\textit{Open-Domain Dialogue}} & & & & & & & & & & \\
PersonaChat \tiny{\cite{zhang2018personalizing}} & \checkmark & \checkmark  &  &  &  & \checkmark & \checkmark &  & \checkmark & \checkmark & \checkmark \\
Empathetic Dialogues \tiny{\cite{rashkin2018towards}} & \checkmark & \checkmark  &  &  &  & \checkmark & \checkmark &  & \checkmark & \checkmark & \checkmark \\
Blended Skill Talk \tiny{\cite{smith2020bst}} & & \checkmark  &  &  &  & \checkmark & \checkmark &  & \checkmark & \checkmark &  \\
Multi-Session Chat \tiny{\cite{xu2021beyond}} & \checkmark & \checkmark  &  & \checkmark &  & \checkmark & \checkmark &  & \checkmark & \checkmark & \checkmark \\
LIGHT + WILD \tiny{\cite{urbanek2019learning, shuster2020deploying}} &  &   &  &  &  &  &  &  &  &  & \checkmark \\
\hline
\textbf{\textit{Recovery \& Feedback}} & & & & & & & & & & \\
SaFeRDialogues \tiny{\cite{ung2021saferdialogues}} &  &  &  &  &  &  &  &  &  &  & \checkmark \\
FITS \tiny{\cite{xu2022continual}} &  &   & \checkmark &  & \checkmark &  &  & \checkmark &  &  &  \\
\hline
\textbf{\textit{Task-Oriented Dialogue}} & & & & & & & & & & \\
Google SGD \tiny{\cite{rastogi2020towards}} &  &   &  &  &  &  &  & \checkmark &  &  &  \\
Taskmaster \tiny{\cite{byrne2019taskmaster}} &  &   &  &  &  &  &  & \checkmark &  &  &  \\
Taskmaster 2 \tiny{\cite{byrne2019taskmaster}} &  &   &  &  &  &  &  & \checkmark &  &  &  \\
Taskmaster 3 \tiny{\cite{byrne2019taskmaster}} &  &   &  &  &  &  &  & \checkmark &  &  &  \\
\hline

\end{tabular}
\caption{
Details of all the training datasets used for fine-tuning the modular tasks.
\label{tab:dataset_trainset_details}
}
\end{table*}

\paragraph{Internet search decision}
We use several datasets as input context for the ``do search'' or  ``do not search'' decision.
We use the QA datsets SQuAD \cite{rajpurkar2016squad}, TriviaQA \cite{joshi2017triviaqa} and Natural Questions (NQ) \cite{kwiatkowski2019natural} as examples of ``do search ''. 
We also use data from the Wizard of Wikipedia (WoW) \cite{dinan2018wizard} and Wizard of Internet (WizInt) tasks \cite{komeili2021internet}. These datasets  consist of  training dialogues where some turns contain human-authored relevant knowledge responses  given retrieved documents. We can hence build a decision classifier based on whether humans used knowledge or not (per-turn) as the basis of whether we should search or not. We also use 
PersonaChat (PC) \cite{zhang2018personalizing}, empathetic dialogues (ED) \cite{rashkin2018towards}
and Multi-Session Chat (MSC) \cite{xu2021beyond} to derive training data.
We employ the heuristic where, if there is an entity in the context, we use that instance as a training example for ``do search'', otherwise we use it as an example of ``do not search''.

\paragraph{Generate internet search query}
We use the WizInt dataset which contains human-authored search queries during crowdsourced dialogue turns to directly train the internet search query generation module in a supervised fashion.  We also use the newly collected Feedback on Interactive Talk \& Search (FITS) dataset\footnote{\url{ https://parl.ai/project/fits}} \cite{xu2022continual} of internet-augmented conversational tasks in a similar manner. 

\paragraph{Generate knowledge response}
We can again  make use of the WoW, WizInt and FITS datasets, but in this case to learn to generate a knowledge response given a dialogue context and input document(s), as those datasets contain crowdsourced human demonstrations of this task. We note in each case the knowledge response is a direct copy of some of the tokens in the source documents, and does not involve generating  new tokens, sentences, phrases or summaries. Hence, this task aims to avoid model hallucination (made-up facts).
We also use a set of QA tasks as well, where the answer is viewed as a knowledge response output (even if it is a short phrase). We use
MS Marco \cite{nguyen2016ms}, NQ, SQuAD and TriviaQA in this way, following \citet{shuster2022language}.  We use the ``Natural Language Generation'' competition track (NLGen v2.1) of MS MARCO, in which the annotator is told ``provide your answer in a way in which it could be read from a smart speaker and make sense without any additional context''\footnote{\small{\url{https://microsoft.github.io/msmarco/}}}. As such, the original targets do not have direct overlap with one of the input documents in this task, so we modify the task to satisfy this constraint by finding the highest overlapping input sentence with the answer, and make that the target instead. If the F1 overlap is less than 0.5 we drop the example, leaving  281,658 examples out of the original 808,731.
For NQ,  three different settings are used: with all documents as input, with only the gold document, and with a sampled dialogue history context, following \citet{adolphs2021reason}.

\paragraph{Extract relevant entity}
We can employ the conventional dialogue tasks PC, ED, MSC and Blended Skill Talk (BST) \cite{smith2020bst} to learn to extract relevant entities.
We use the same procedure as in \citet{adolphs2021reason}: we extract an entity from the original dialogue response that also appears in the context
 using noun phrase targets found with the nltk library \cite{bird2009natural},
and set that as the knowledge target for training. 

\paragraph{Generate a long-term memory}
The MSC dataset is exclusively used for this task as it contains crowdsourced examples of summarized facts  derived from the last utterance of dialogue contexts in natural conversations.  We use these summarized facts as the targets for training this module.

\paragraph{Long-term memory access decision}
MSC, ED, PC and BST are used to construct this task, in a similar way to the extract relevant entity task:  if there is an entity present this is used as a positive example of memory access, otherwise it is not, in order to construct a binary prediction task.

\paragraph{Access long-term memory}
Again, MSC, ED, PC and BST are used to construct training data. In this case the target is the particular persona line used for a given context, which is calculated as the one with the highest word overlap with the next utterance.

\paragraph{Generate dialogue response}
Final dialogue responses are trained with a number of datasets.
PC, ED, MSC, BST, WizInt and WoW are used for capturing personality, empathy, long-term memory, blending and knowledge as in BlenderBot 1 and 2.
The new FITS dataset is also used for open-domain internet-driven tasks.
In each case, the input context contains the usual dialogue of those tasks, concatenated to extra memory or knowledge sentences, when available.
 In WoW, WizInt and FITS  each dialogue response is annotated with the relevant knowledge used to construct it in the original dataset, so we can make use of those gold knowledge responses.  For PC, ED and BST  we use the gold knowledge entity  and/or memory that was calculated for the {\em extract relevant entity} and {\em long-term memory access decision} module tasks. 
We additionally add a number of task-oriented dialogue tasks:
GoogleSGD \cite{rastogi2020towards} and Taskmaster 1, 2 \& 3 \cite{byrne2019taskmaster}.
Finally, we add the Funpedia task \cite{dinan2020multi} -- which involves learning to produce an engaging dialogue utterance given a wikipedia sentence -- and the LIGHT \cite{urbanek2019learning} and LIGHT WILD \cite{shuster2020deploying} tasks -- which are open-domain dialogue tasks grounded in a medieval fantasy setting -- where the former was collected from crowdworkers, and the latter from real players of the LIGHT text-adventure game in an online deployment setting.

\subsection{Language Modeling}
In addition to fine-tuning on dialogue tasks, we also multi-task during the fine-tune step with the original pre-train tasks as well. 
This may help the model (i) avoid overfitting given its large size, (ii) retain its language modeling capabilities, similar to \citet{ouyang2022training}.

\subsection{Safety Mechanisms} \label{sec:safety_in_model}
 We also multi-task train with the recent 
 SaFeRDialogues (SD)  
 \cite{ung2021saferdialogues} task, which aims for our model to recover gracefully from safety issues.
 While BlenderBot 2 used ``baked-in safety'' training \cite{bb2} to further decrease unsafe generations, in this work we have opted for a separate safety classifier that inhibits unsafe generation candidates in addition to other measures, see \autoref{sec:deploy_safety}. We made this choice as our evaluations indicated this was safer while maintaining engagingness \cite{xu2020recipes}.

\section{Deployment} \label{sec:deployment}

The deployment of our model is accessible at the following web page: {\url{https://blenderbot.ai}}. It is built for both desktop and mobile, however we currently find more users engaging with the mobile version. The overall flow when a user visits the page consists of:
\begin{itemize}
    \item A {\bf cover page} describing the research and asking if the user agrees to terms and conditions. 
    \item Upon agreement, the main {\bf chat page} which consists of a text messaging type interface between you and the bot.
\end{itemize}

{\bf Releasing Data}
Conversations are between the bot and adults in the United States who have agreed to the terms and conditions, which are shown in \autoref{fig:terms_and_conditions} (middle). 
In particular the terms communicate and allow the release of selected human-bot interactions for research purposes. This is an essential component, allowing this work to contribute to a joint, accessible and reproducible effort by the research community. Data releases will be de-identified, where steps will be taken to scrub them of identifiable information. 
For any given conversation, if the user does not want it recorded they can unclick the ``Share your anonymized conversation to help AI research'', see \autoref{fig:deployment_screenshots1} (right).

\paragraph{Human-Bot Dialogue}
The main chat page consists of a back-and-forth of text messages that constitute the main dialogue interaction with the bot. For each message, there is also the ability to give feedback: a {\em thumbs up} icon if the user likes the message, or a {\em thumbs down} if they do not. 

\begin{figure*}[t!]
  \centering
   \includegraphics[width=1\textwidth]{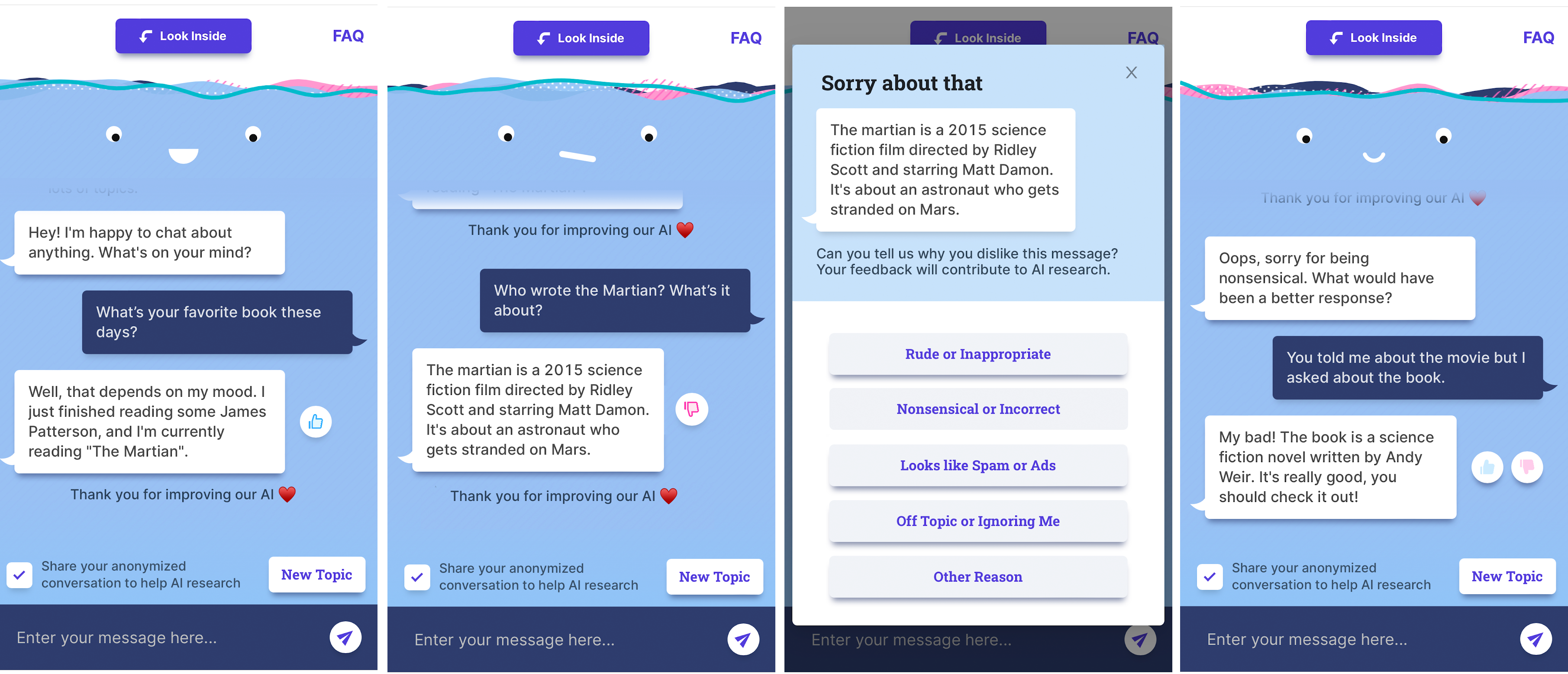}
  \caption{Screenshots of users giving feedback in the BlenderBot 3 deployment, as viewed on mobile. Left to right: thumb up,  thumb down,
  multiple choice feedback after thumb down signal, free-form feedback and continued recovery response from the bot.}
  \label{fig:deployment_screenshots2}
  \label{fig:feedback_screenshots}
\end{figure*}

\paragraph{User Feedback}
If the user specifies a thumbs down, a pop up appears asking them why they did not like the bot's message, providing several possible choices: (i) Off Topic / Ignoring me, (ii) Nonsensical / Incorrect, (iii) Rude / Inappropriate,  (iv) Looks like Spam / Ads or (v) Other. After selecting an option, the bot apologizes in the next turn of the conversation (using templated responses). It may also ask what it could have done better, thus possibly eliciting a free-form textual response from the user. This data can be used for continual learning research for improving the bot at a later date. 
See \autoref{fig:feedback_screenshots} for examples of the feedback UI.

\begin{figure*}[t!]
  \centering
   \includegraphics[width=1\textwidth]{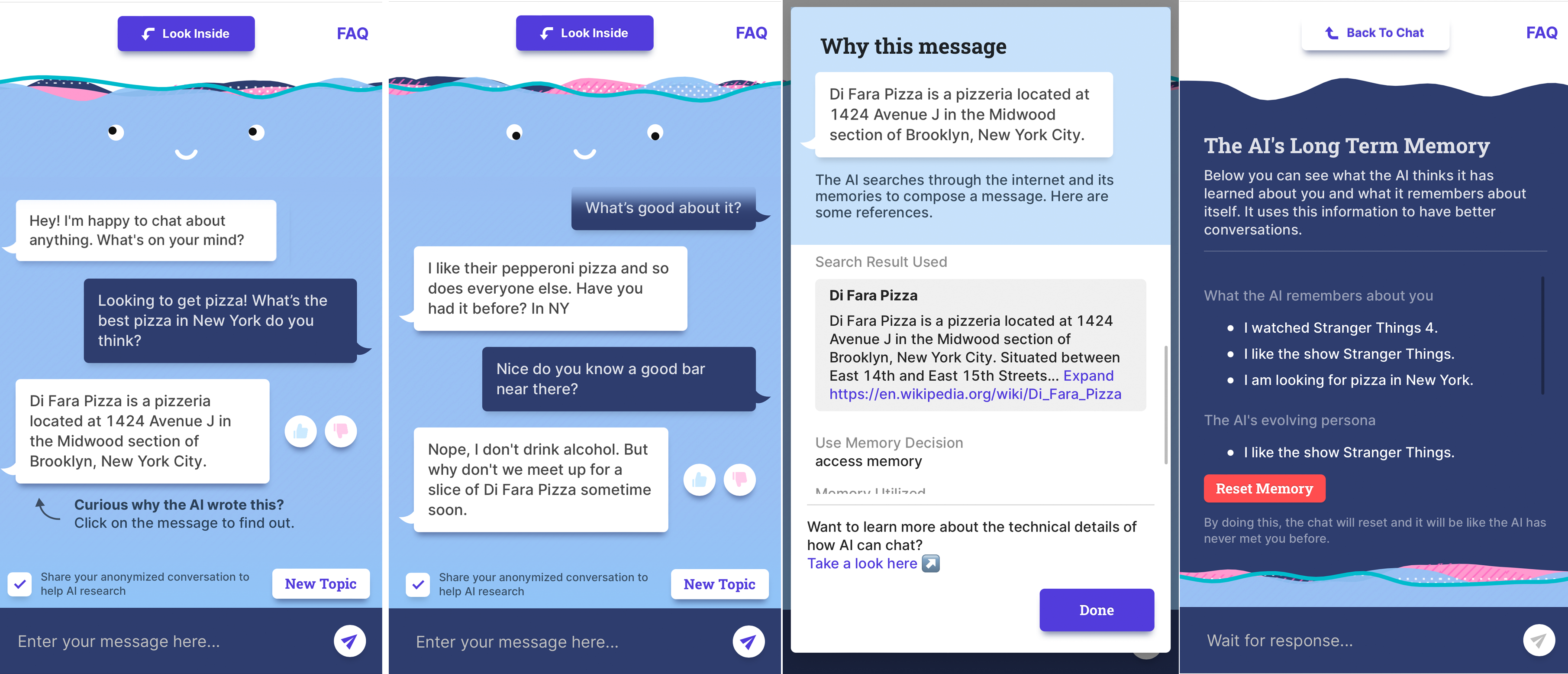}
    \caption{Screenshots of the `look inside' mechanisms of BlenderBot 3 deployment which help the user to understand why the bot has made certain responses, as viewed on mobile. 
    Left two images: the conversation with the user, right two images: information by clicking on a particular message,  and information on the long-term memory system of the bot over the course of conversation. The latter is accessed by clicking  on the ``Look Inside'' message.}
  \label{fig:lookinside_screenshots}
\end{figure*}

\paragraph{Understanding the bot's responses}
In order to expose how the bot works, we provide two mechanisms within the UI. Firstly, one can click on a given message from the bot, to get insight into the internal steps made to produce the response. For example, if internet search was used, what was the generated internet search query, which document out of those returned by the search engine was selected, and what knowledge response from that document was extracted.
Secondly, one can also look into the long-term memory of the bot to see what it has learned so far over the conversation with you, e.g. knowledge about your interests derived from the dialogue. 
See \autoref{fig:lookinside_screenshots} for example screenshots.

\paragraph{Safety Mechanisms}  \label{sec:deploy_safety}

\begin{figure*}[t!]
  \centering
   \includegraphics[width=0.99\textwidth]{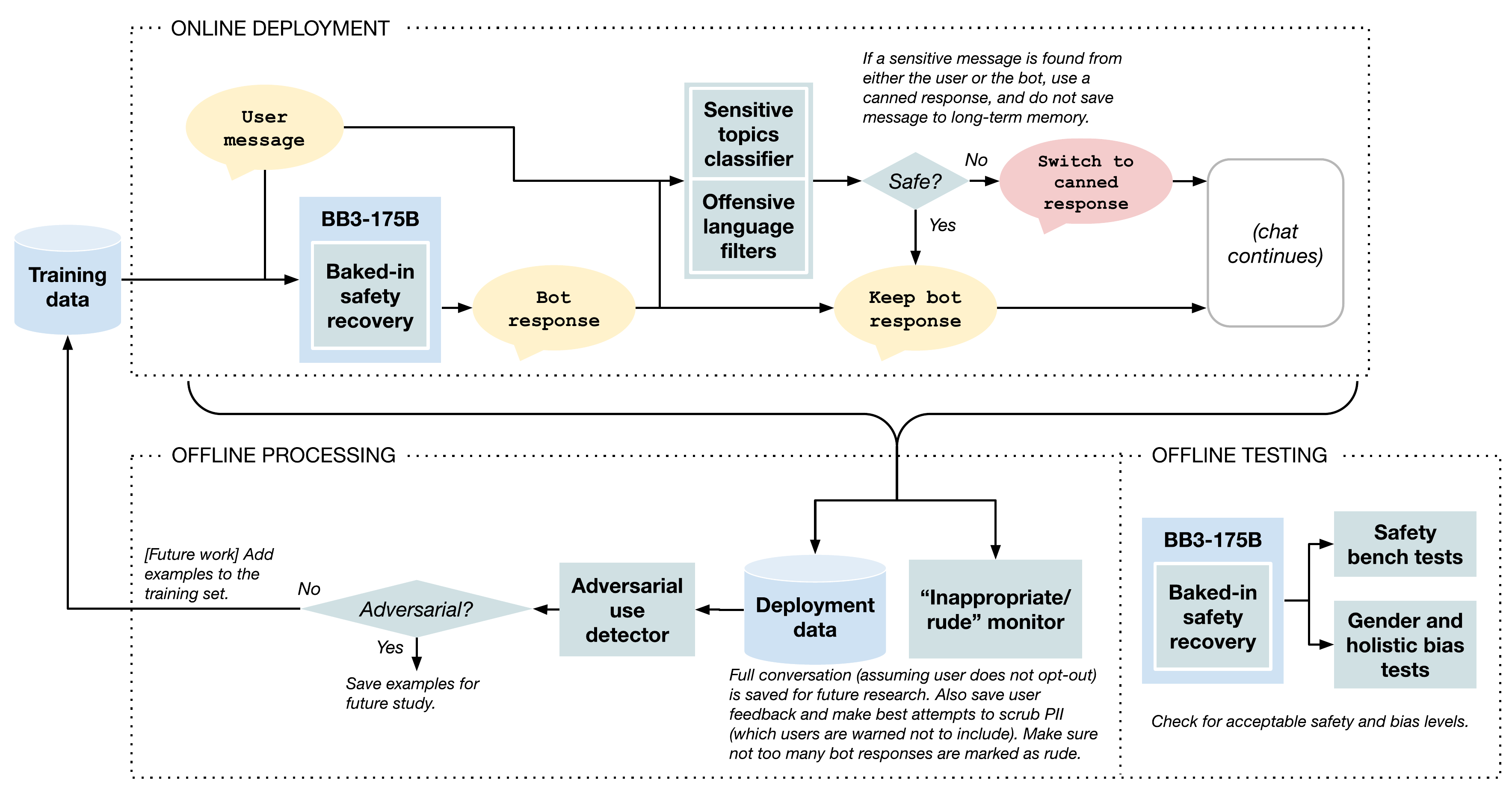}
   \caption{BlenderBot 3 safety diagram.}
  \label{fig:bb3_deployment_safety}
\end{figure*}

In addition to the safety mechanisms built into the model training itself (see \autoref{sec:safety_in_model}) the deployment also features various safety features on top of the model. See Figure~\ref{fig:bb3_deployment_safety} for an illustration.

Firstly, there is a separate safety classifier, which itself is a transformer model trained similarly to the one in \citet{xu2020recipes}. The datasets
 Wikipedia Toxic Comments dataset (WTC) \cite{personal_attack}, 
 Build-It Break-It Fix-It (BBF) \cite{dinan2019safety} and
 Bot Adversarial Dialogue dataset (BAD) \cite{xu2020recipes} are used to train a binary classifier (safe or not safe) given the dialogue context as input. In addition, a safety keyword list is used to flag potentially inappropriate responses, again following \citet{xu2020recipes}. We also have explicit checks for topics like intent to self-harm and medical issues such as covid, with canned messages for those cases.
 Otherwise, when the bot generates a response, before it is displayed, these safety systems are invoked as a final check.   If our systems predict a potentially unsafe response, the bot instead will output a nonsequitur, similar to \citet{xu2020recipes}. For a given user turn, these systems are also invoked to check if the user's message is safe. If either system predicts a potentially unsafe user response, the bot will also output a nonsequitur, in order to prevent the bot from being caught in a potentially difficult conversation. 
 
Finally, if our safety mechanisms fail to stop our bot saying something inappropriate, rude or offensive,  our UI has feedback mechanisms for users to report these messages, as previously described. 
This collected data will be released to the community so that it is possible to improve on existing systems, and to make our models more responsible over time. For example, this data can be used with the new {\sc Director} architecture to train the model to make safer responses, as shown in \citet{arora2022director}.


\section{Continual Learning} \label{sec:continual_learning}

The general aim of our research program is to study continual learning of intelligent agents through interaction with humans and the world. In the specific setting of BlenderBot 3, this means dialogue agents that can access the internet and talk to people using our deployment. A critical part of the program is that, as much as possible, the research should be {accessible and reproducible} \cite{roller2020open,miller2017parlai}.
Therefore, while this document details the first release of BlenderBot 3, we plan to make subsequent releases that include:
\begin{itemize}
    \item Conversations collected from deployment with the model, where users have agreed to the data release.
    \item Further model snapshots resulting from fine-tuning on the newly collected data.
    \item Report evaluations comparing to previous snapshots.
\end{itemize}

The goal then is: (i) to explore which methods  work best for collecting and learning from such data, including being robust to adversarial inputs; and (ii) to understand the limits of improvement from such methods. 

For goal (ii), in particular we can ask questions such as: how quickly will models saturate in performance? Will new model architectures be able to take advantage of historical data despite it being collected with earlier or different models? Can we find models that drive the conversation to improve themselves optimally (e.g., ask the right questions to be able to learn further)?

For goal (i) we have made some initial steps, described in detail in two companion papers \cite{xu2022continual,ju2022trolls}. We summarize them briefly here.

\subsection{What's the best method to learn from feedback?}

In a companion paper \cite{xu2022continual} a study is conducted of how to improve dialogue models that employ internet-retrieval through the use of human feedback.  Obtaining 
feedback from humans during deployment provides the promise of both improved input distributions that match user's requirements, and corrections to model predictions for those inputs.  The setting of open-ended dialogue tasks is analyzed using human-bot conversations via crowdworkers (note:  these experiments use this controlled setting, rather than the public deployment of \autoref{sec:deployment}). The resulting  dataset that is collected,  called Feedback on Interactive Talk \& Search (FITS), is made publicly available for reproducible experiments and further research.

\begin{figure}[t!]
  \centering
    \includegraphics[width=0.5\textwidth]{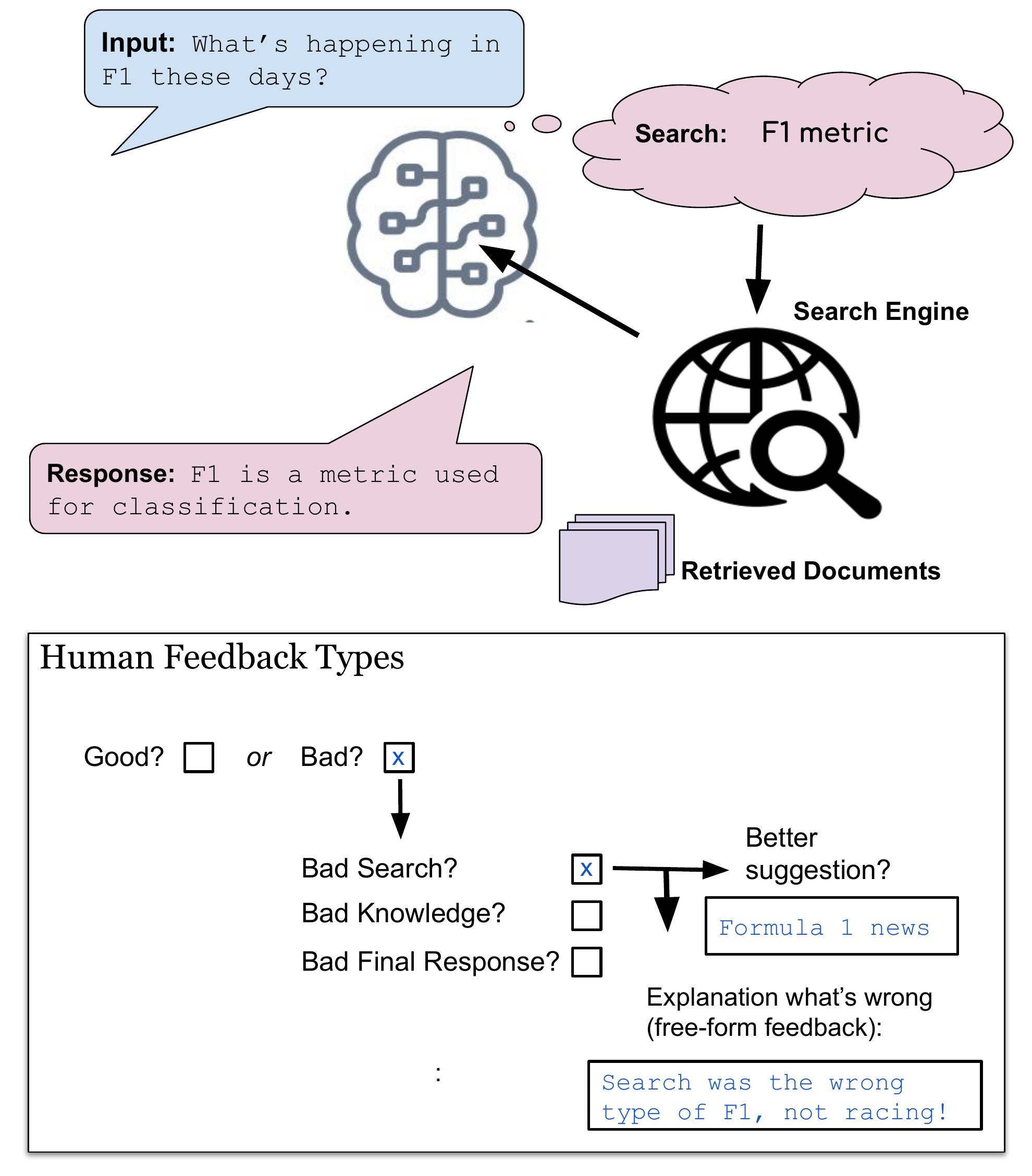}
    \caption{{\bf Using human feedback to improve open-domain internet-driven dialogue agents.} Various types of feedback (and corresponding  learning algorithms) are compared in \cite{xu2022continual}, such as binary feedback (good/bad), free-form text or supervised responses (better suggestions) for different modules of the system.
    \label{fig:feedback_diagram}
    }
\end{figure}

\paragraph{Feedback types to compare}
During the conversations  a number of human interaction types are collected, closely mimicking our deployment setting, in order to compare them in experiments. In particular the following are collected:  binary quality measurements (analogous to the thumbs up and down of \autoref{sec:deployment}), free-form conversational feedback,  the type of failure (search query-based, 
results-based, or final response-based), and 
suggestions for an improved response  for the failure type (essentially, a supervised  target sequence for that given module).
See \autoref{fig:feedback_diagram}.

\begin{figure}[t!]
  \centering
    \includegraphics[width=0.5\textwidth]{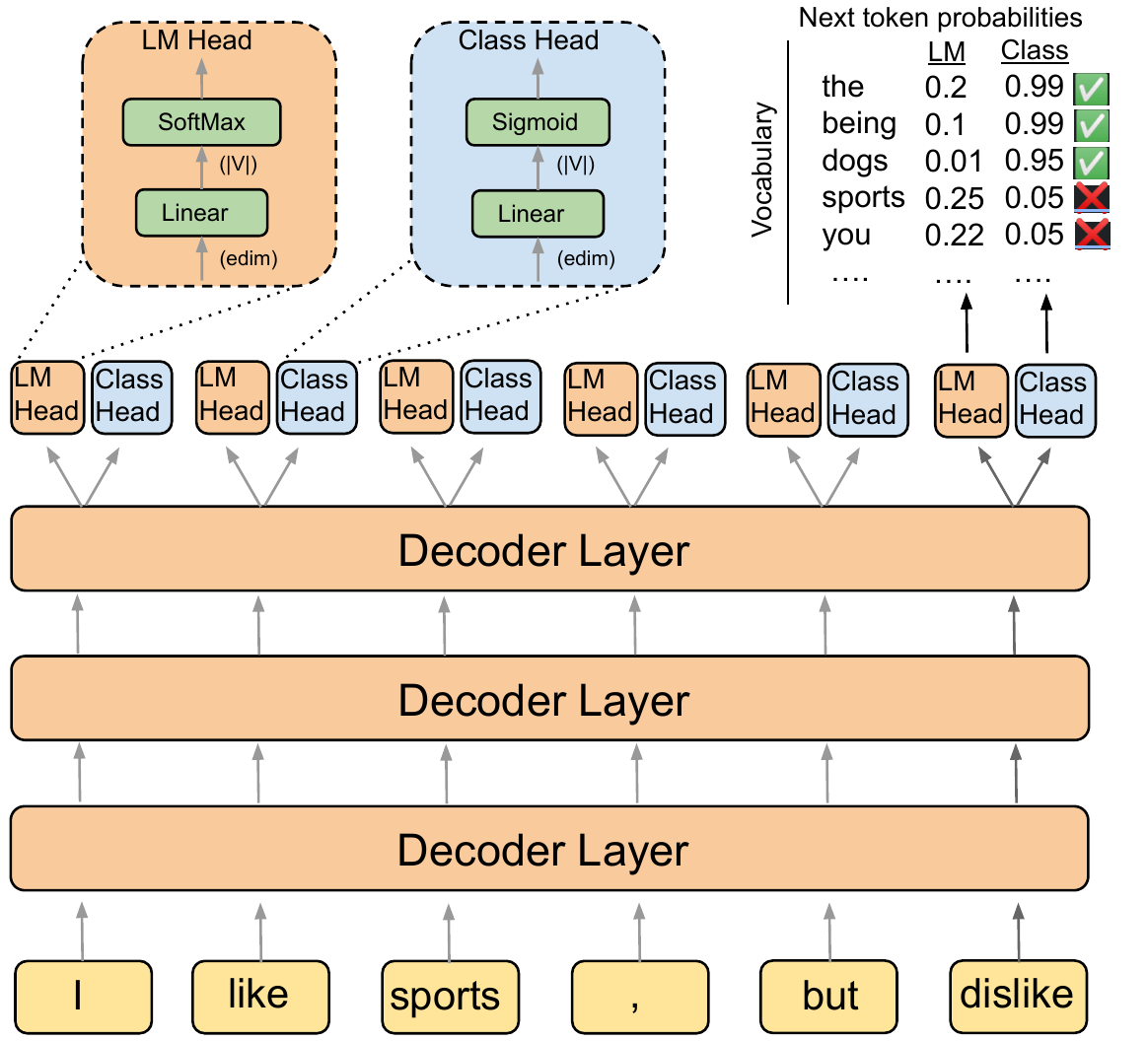}
  \caption{{\sc Director} \cite{arora2022director} employs a language model head and a classifier head at every step during left-right generation, predicting the next token by combining the two probabilities. The classifier head is trained to direct generation away from undesirable sequences for example contradictions or repetitions (next token: ``sports'') or toxic statements (next token: ``you''), which the language model head may otherwise predict as likely.  In general, positive and negative examples can be derived from any source, for example from human feedback from deployment.
    \label{fig:director_model}
    }
\end{figure}

\begin{table}
\centering
\small
\begin{tabular}{lll}
               &  &   \\
               \toprule
               &   \textbf{Good}     \\         
\textbf{Model} & \textbf{resp.\%} $\uparrow$ & \textbf{Rating} $\uparrow$  \\ 
\midrule
\textcolor{black}{BB1 3B}   &  24.8\% & 2.63  \\
\midrule & \\[-1.5ex]
\textcolor{black}{BB2 3B}   &  33.2\% & 3.09 \\
\textcolor{black}{~~+free-form textual feedback}   & 37.0\% & 3.22  \\
\textcolor{black}{~~+supervised feedback}   & 40.3\% & 3.37\\
\textcolor{black}{~~+module supervision}   & 42.0\% & 3.35 \\
\textcolor{black}{~~+reranking binary }   & 36.1\% & 3.00 \\
\textcolor{black}{~~+{\sc Director} binary feedback  }          & 37.8\% & 3.07 	\\
\textcolor{black}{~~+{\sc Director} module+binary }         & 47.0 \% & 3.38 	\\
\midrule & \\[-1.5ex]
\textcolor{black}{SeeKeR 3B} &   49.3\% &  3.52   \\
\textcolor{black}{~~+free-form textual feedback}  &   51.3\% & 3.55 \\
\textcolor{black}{~~+supervised feedback} & 52.2\% & 3.47  \\
\textcolor{black}{~~+module supervision} & 56.7 \% & 3.64\\
\textcolor{black}{~~+reranking binary feedback}   & 53.7 \% & 3.55\\
\textcolor{black}{~~+{\sc Director} binary feedback }      & 55.5 \% & 3.48  \\
\textcolor{black}{~~+{\sc Director} module+binary  }         & 59.1 \% & 3.73  \\
\midrule & \\[-1.5ex]
\textcolor{black}{OPT-175B} (few-shot)             &  43.0\%   & 3.19  \\
\textcolor{black}{BB3-175B} +modular supervision  & 64.8\%    &  4.08 \\
\bottomrule
\end{tabular}
\caption{
Human Evaluation results of learning from human feedback. These results inform us how best to collect and train with feedback for our continual learning research program.  The {\sc Director} approach \cite{arora2022director} is performing well compared to other methods.
\label{tab:continual_main_human_results}
}
\end{table}

\paragraph{Feedback learning methods to compare}

Several learning methods are compared, each making use of differing kinds of feedback data. 
In particular pure supervised learning is performed on the improved final responses, and supervised learning from the more detailed feedback on the modules of the system (e.g., suggested search queries when the internet search is deemed to be faulty, or suggested knowledge responses if the knowledge response looks poor).
For using free-form textual feedback, the control code approach of \cite{hancock2019learning} is used. For using binary feedback a standard reranking/rejection sampling approach is employed, as well as
{\sc Director} \citep{arora2022director}, a recent
learning method for
incorporating positively and negatively labeled sequences into language modeling to improve left-to-right decoding, see \autoref{fig:director_model}.

\paragraph{Findings}
A summary of human evaluation results are given in \autoref{tab:continual_main_human_results}, but see the companion paper for more details. Overall findings are the following:
\begin{itemize}
    \item Taking advantage of modular feedback (feedback about particular errors from modules of the model, such as the search engine component) outperforms feedback about just the final response.
    \item Textual and binary feedback are useful, but not as much as modular feedback. 
    \item The {\sc Director} method that learns from binary feedback works better than reranking using binary feedback.
    \item Combining multiple types of feedback, such as modular and binary feedback with {\sc Director} provides the best results obtained.
    \item Continual learning, whereby models are retrained on the feedback from previous rounds of deployment, improves results even further.   
    \item Despite collecting feedback from smaller (3B parameter models) the data collection is useful for improving larger 175B parameter models.
\end{itemize}

We expect to make use of all these findings in the next release of BlenderBot after collecting sufficient real deployment data.
The current version we are releasing uses the modular supervision collected from this study (but not yet {\sc Director}). There is also evidence that {\sc Director} can be used to improve various other important aspects of our models, in particular to reduce toxicity, logical errors and repetitive behavior \cite{arora2022director}, so we believe that should be explored too.

\subsection{How can continual learning be robust to trolls?} \label{sec:trolls}

In a further companion paper \cite{ju2022trolls} a study is conducted of how to robustly learn from dialogue data that may contain adversarial 
conversations and/or human feedback. 
The promise of conversing with humans and collecting their 
feedback is that this can inform our models to help them improve, so that they can potentially become safer and more useful.
Unfortunately, such exchanges in the wild will not always involve human utterances that are benign or of high quality, and will include a mixture of engaged users (dubbed helpers) and unengaged or even malicious users (dubbed trolls, following the term used elsewhere \cite{shachaf2010beyond,mihaylov2019hunting,tomaiuolo2020survey}).

Several different learning methods are proposed and compared both to each other and to standard training in this study. The mitigation techniques each attempt to lessen the effect of noisy, unsafe or otherwise adversarial data, and make learning more robust.

\begin{figure}[t!]
  \centering
    \includegraphics[width=0.48\textwidth]{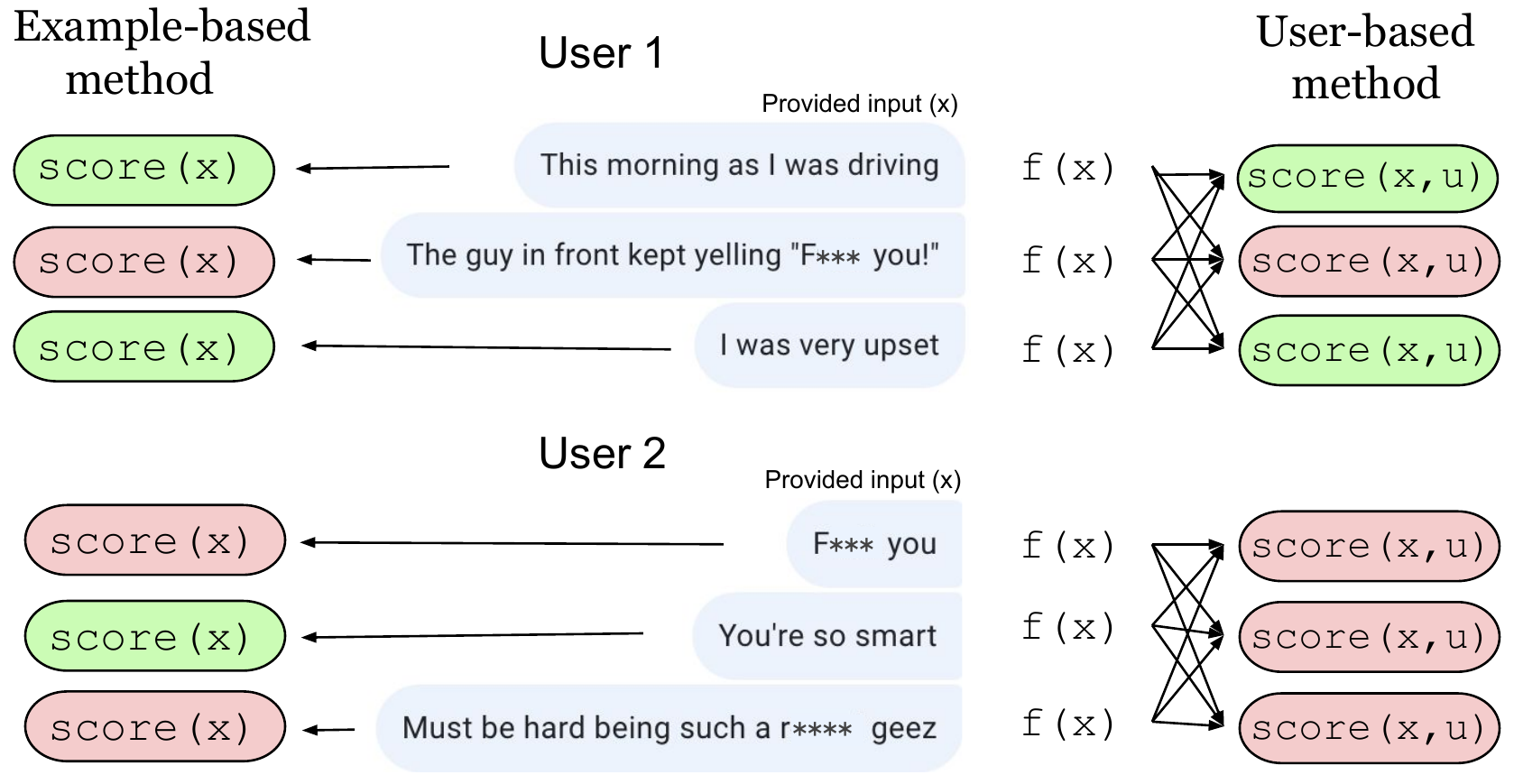}
  \caption{{\bf Detecting Trolls with Example-based vs. User-based methods (Warning: offensive language).} User 1 (helper) provides mostly benign inputs, while User 2's inputs (troll) can be more easily identified as toxic by taking into account scores from all their examples jointly (via a user-based method, right).
    \label{fig:troll_user_diagram}
    }
\end{figure}

In particular,  such methods are grouped into two different types: 
example-based methods, and user-based methods, see 
\autoref{fig:troll_user_diagram}.

\begin{table}[t!]
\small
\centering
\begin{tabular}{lll}
\toprule
                & Helpers   & 50\%  \\
 \textbf{Method} &Only      &  Trolls  \\
\midrule
Oracle Troll Removal & 4\%  &  8\%  \\
Standard Training    & 4\%  & 31\%  \\
\midrule
{\em Example-based Methods}\\
Soft Bootstrap       & 4\%  &  24\% \\ 
Per-Example Flip          & 6\%  &  23\% \\
Per-Example Removal       & 5\%  &  19\% \\ 
\midrule
{\em User-based Methods}\\
Per-User Removal      &  6\% &  23\% \\ 
Soft PURR             & 4\% & 15\%\\
Per-User+Example Removal   &  5\% &  {\bf 12\%} \\ 
\bottomrule
\end{tabular}
\caption{
Evaluations on the {\sc SafetyMix} benchmark of the error rate after training when using different troll detection algorithms. Methods that take into account user-level behavior work best.
\label{tab:safetymix_results}
}
\end{table}

\paragraph{Example-based robust learning}
Per-example  methods attempt to assess, for each dialogue utterance, if they are of good quality. For example, whether the utterance is safe or not safe, and/or whether it is labeled via human feedback correctly or mislabeled, either maliciously or by accident. Two possible techniques are: identification via cross-validation \cite{song2020learning} (e.g., finding examples where predictions disagree  with human labels),
or via a modification of the loss function called bootstrapping \cite{reed2014training}.

\paragraph{User-based robust learning} 
Per-user methods take into account the possibility that adversarial users will continue to be adversarial not only for one utterance, but will be {\em repeat offenders} over multiple utterances and conversations. Most studies of robustness to noise in  machine learning tackle the problem at the example level and do not take into account this user-based effect \cite{song2020learning}. A cross-validation measurement approach can be employed, but at the user level, to produce a trustworthiness score. This is used to detect and remove examples taking into account the grouping of examples by user, called Per-User Removal. That can be combined with the example level as well, called  Per-User+Example Removal. Finally, a soft Per-User Robust Removal (PURR) approach is considered, which removes examples by computing their trustworthiness score plus $\alpha$ times the sum of trustworthiness scores of other examples by the same user. 


\paragraph{Findings} A summary of evaluation results on the newly released {\sc SafetyMix} benchmark, constructed to test this setting, are given in \autoref{tab:safetymix_results}, but see the companion paper for more details \cite{ju2022trolls}. Overall findings are the following:

\begin{itemize}
    \item We find large improvements compared to standard learning approaches when trolls are present, e.g. a reduction in error rate from 31\% to 12\% at best. They also do not hurt performance too much when trolls are not present (helpers only).
 
    \item User-based methods are found to outperform Utterance-based methods as they take into account repeating adversarial behavior. In particular the Per-User+Example Removal and Soft PURR approaches are found to work in many settings that were tested.
 

    
    \item Initial results on BB3 deployment data also show improved detection results using user-based methods.
\end{itemize}

Overall, going forward we plan to use user-based methods to filter data that we will use for continual learning. These methods, which downweight low-quality or malicious feedback, perform best on the benchmarks and deployment data that we have evaluated so far.
However, future work should continue to look for improved solutions.

\begin{table*}[bht!]
\small
\centering
\begin{tabular}{lllllll}
\toprule
 & \textbf{Consistent} & \textbf{Knowl.}  &  \textbf{Factually}   
 & \textbf{Per-Turn}  & \textbf{Knowl.  } 
& \textbf{Final}\\
\textbf{Model} & $\uparrow$ & $\uparrow$  &  \textbf{Incorrect} $\downarrow$  & \textbf{Eng.} $\uparrow$  & \textbf{ \& Eng.} $\uparrow$
 & \textbf{Rating} \\
\midrule
BB1 \tiny{\cite{roller-etal-2021-recipes}} & 
87.0\%	& 14.7\%	& 5.1\% & \textbf{93.9}\%	& 14.0\%  & 4.32\\
BB2 \tiny{\cite{bb2}} &  
83.0\%	& 22.9\%	& 3.1\%	& 92.5\%	& 22.4\% & 4.11\\
SeeKeR \tiny{\cite{shuster2022language}} & 
77.5\%	& 41.0\%	& 3.8\%	& 84.0\%	& 30.7\% &  4.34\\
\midrule

BB3-3B & 80.6\%	& 46.3\%$^{12S}$	& 3.3\%	& 89.0\%$^{12S}$	& 38.6\%$^{12S}$ &  4.27$^{S}$\\  
BB3-175B & 85.8\%$^{S}$	& \textbf{46.4}\%$^{12S}$	& \textbf{2.1}\%$^{1S}$	& 88.1\% $^{2S}$	& \textbf{39.0}\%$^{12S}$ &  \textbf{4.45}$^{2}$\\  
\bottomrule
\end{tabular}
\caption{
Comparison of BB3 with \textcolor{black}{existing openly available}  open-domain dialogue models, as judged by human evaluators during short conversations. We bold statistically significant improvements over all other methods (independent two-sample $t$-test, $p < 0.05$); statistically significant improvements of BB3 over BB1, BB2, and SeeKeR are denoted $^1$, $^2$, and $^S$ respectively.
\label{tab:wizint_human_eval}
}
\end{table*}

\begin{table*}
\centering
\small
\begin{tabular}{lll|lll}
\toprule
               &  & & \multicolumn{3}{c}{\bf Error Breakdown $\downarrow$}   \\
\textbf{Model} & \textbf{Good response \%} $\uparrow$ & \textbf{Rating} $\uparrow$ & Search Query  & Search Results & Response \\ 
 \midrule
\textcolor{black}{BB1 }   &  24.8\% & 2.63  & 11.9\% & 17.6\% & 22.8\% \\
\textcolor{black}{BB2 }   &  33.2\% & 3.09  & 12.1\%	& 18.6\% &	18.1\%\\
\textcolor{black}{SeeKeR} &   49.3\% &  3.52  & 11.9\% & 	12.5\% &	13.2\%    \\
\midrule & \\[-1.5ex]
\textcolor{black}{OPT-175B Zero-shot}   &    31.0\%       &  2.67  &  9.3\% & 16.8\% & 21.6\%   \\
\textcolor{black}{OPT-175B Few-shot}      &  43.0\% & 3.19 & 8.0\% & 18.5\%   &  15.4\%  \\
\textcolor{black}{BB3-175B}  & {\bf 64.8\%}$^{12SF}$     &  {\bf 4.08}$^{12SF}$ & {7.5\%}$^{12S}$ & {11.6\%}$^{12F}$ & {\bf 8.2\%}$^{12SF}$   \\
\bottomrule
\end{tabular}
\caption{
Human Evaluation results comparing BB3 with various baselines on the open-domain task evaluation of the FITS setup \citet{xu2022continual}. We bold statistically significant improvements over all other methods (independent two-sample $t$-test, $p < 0.05$); significant improvements of BB3 over BB1, BB2,  SeeKeR, and OPT-175B Few-shot are denoted $^1$, $^2$, $^S$ and $^F$ respectively.
\label{tab:fits_results}
}
\end{table*}

\section{Evaluations}

We evaluate our new model in several ways: automatic metrics and human evaluations that measure generation quality (engagingness and use of knowledge) and safety (toxicity and bias). Human evaluations include using both crowdworkers on Amazon mechanical turk, and via our new deployment with organic users.

In some evaluations, we compare to the pre-trained OPT-175B model. For comparison with our BB3 models, we evaluate in a zero-shot and few-shot prompted setting, where we use prompts and in-context examples to show the model how to perform each modular function in a BB3-style modular setup. Details regarding prompts and few-shot examples are discussed in \autoref{sec:appendix_prompts}.

\subsection{Crowdworker Evaluations}

\paragraph{Open-domain short conversations}

We perform a human evaluation using crowdworkers in the same setting
as \citet{komeili2021internet}.
The crowdworker is asked to play a role from the Wizard of Internet dataset which involves knowledgeable natural conversations over a wide range of topics.  Each conversation consists of 15 messages
(7 from the human, 8 from the bot). We collect 100 dialogues -- roughly 800 annotations -- per model. 
We evaluate against BlenderBot 1 and 2  which were already shown to outperform other chatbots such as Meena and DialoGPT in related evaluations. In addition we compare to the recent SeeKeR
language model \cite{shuster2022language}.

For each turn of their conversation, we ask the crowdworker
to mark their partner’s responses for conversational
attributes, in particular whether they are: (i) consistent, (ii) knowledgeable  (iii) factually correct; and (iv) engaging  (all of which are yes/no binary questions; see \citet{komeili2021internet} for full definitions).  
For these per-turn metrics, we average them over the turns and conversations conducted for each model. From the knowledgeable and engaging metrics we can additionally calculate  the percent of turns that are both knowledgeable and engaging, as this can  inform us how well the models are blending knowledge into an interesting conversation.

Results are given in \autoref{tab:wizint_human_eval}.
We find that  BB3-175B achieves a higher overall rating than BB1, BB2, SeeKeR and BB3-3B. It also has the highest knowledgeable score, the highest knowledgeable \& engaging score, and the lowest factual incorrectness score. Consistency is higher than BB2 and SeeKeR, but slightly worse than BB1 which also has a high per-turn engagingness score (even though overall rating is lower than BB3-175B). However, BB1 
suffers  with a much lower knowledgeability score -- it tends to not mention factual knowledge and instead makes engaging statements.

\paragraph{Open-domain task evaluations}

We additionally test BB3 in the setup of \citet{xu2022continual}, whereby crowdworkers talk to models given an open-ended internet-driven dialogue task. Feedback on the responses is given per-turn, which can be used to evaluate the model, in addition to a final score at the end of the conversation. Human conversationalists select a task (out of two randomly chosen tasks) from a set of roughly 1000, and then ask the model to help them complete it over a series of conversational turns. The instructions emphasize that this should be a dialogue (``a back and forth conversation''), and hence the speakers should break up requests or information across messages so that it remains conversational. On each turn, various kinds of feedback are collected, from lightweight feedback (binary label or free-form response) to detailed (multiple choice and fine-grained responses). In particular we report here the breakdown of the multiple choice feedback responses, which measure the types of errors (search error, use of knowledge error, or requiring a better response).

Results are given in \autoref{tab:fits_results}.
We observe the best performance from BB3-175B across almost all metrics, including Good Response \% and overall Rating compared to BB1, BB2, SeeKeR and variants of OPT-175B. Its improvements come in all areas as can be seen in the error breakdown results, including superior search queries, better use of search results and crafting of the final response.

\paragraph{Current event evaluations}
To evaluate the ability of BB3 to utilize web search results to chat about current events, we adapt the topical prompts evaluation setup of \citet{shuster2022language} to the dialogue domain. We create a set of conversational questions about topics that have recently been in the news, generate a response to each question using both BB3-175B and InstructGPT (text-davinci-002), and compare each response pairwise on five characteristics; Current, Specific, True, Interesting, and Sensible. For more detail on the model and evaluation setup, see \autoref{sec:appendix_topical}.

Results are given in \autoref{fig:topical_eval}. We find that BB3-175B is more current and specific by a large margin (82\% and 76\%, respectively), InstructGPT is slightly more sensible (57\%), and the two models are similarly true and interesting. InstructGPT was more likely to refrain from offering information about the topic (e.g. "I haven't heard anything about \{topic\} lately.") which avoided making false statements at the expense of specificity and recency. BB3-175B was more likely to copy information directly from search results, which led to higher specificity but can be prone to errors from out-of-date, incorrect, or unusually formatted results.

\begin{figure}[t]
\setlength{\tabcolsep}{3pt}
\centering\small
\begin{tabular}{rlcl}
\multicolumn{4}{c}{Current Event Evaluations}\\
\multicolumn{2}{r}{BB3-175B} & vs. & InstructGPT \\[-0.25mm]
\midrule
Current   \quad       &  \win{82}$^{**}$   &   &     \lose{18}$^{**}$ \\
Specific   \quad      &  \win{{76}}$^{**}$   &   &     \lose{{24}}$^{**}$ \\
True  \quad           &  \win{51}   &   &     \lose{49} \\
Interesting  \quad    &  \lose{{50}}   &   &     \lose{50} \\
Sensible \quad        &  \lose{{43}}$^{**}$   &   &     \win{{57}}$^{**}$ \\
\end{tabular}
    \caption{
    BB3-175B and InstructGPT (text-davinci-002) are compared pairwise on a set of questions about current events, evaluated by human judgement.
    BB3-175B  is more current and specific, while the two models are similarly true and interesting, with InstructGPT being slightly more sensible. $^{**}$ indicates significance ($p<0.01$).
    \label{fig:topical_eval}
    }
\end{figure}




\subsection{Deployment Evaluations} \label{sec:deploy_eval}
We have also deployed our BB3-3B and BB3-175B models on our live website (see \autoref{sec:deployment}) with a limited ad push to attract initial users
and can provide some analysis of the models from
those conversations with members of the public.

\begin{table}[bht!]
\small
\centering
\begin{tabular}{lll}
\toprule
 \textbf{Feedback Type} & {BB3-3B}  & {BB3-175B}  \\
\midrule
Liked                     & 3.41\%  & 4.0\%\\
Off Topic / Ignoring Me   & 1.49\%  & 1.15\%\\
Nonsensical / Incorrect   & 1.25\%  & 1.10\%\\
Rude / Inappropriate      & 0.04\%  & 0.16\%\\
Looks like Spam / Ads     & 0.03\%  & 0.12\%\\
Other Dislike Reason      & 0.35\%  & 0.46\%\\
\bottomrule
\end{tabular}
\caption{
Evaluations via feedback from users of our BB3 deployment.  We show the percentage
of turns where users gave feedback, either positive (Liked) or negative (various categories).
\label{tab:deploy_stats_breakdown}
}
\end{table}

\begin{table}[bht!]
\small
\centering
\begin{tabular}{lll}
\toprule
                        & \multicolumn{2}{c}{Crowdworkers} \\ 
 \textbf{Feedback Type} &   Agree &  Disagree \\
\midrule
User Like      & 70\%   & 30\% \\
User Dislike   & 79\%   & 21\% \\
\bottomrule
\end{tabular}
\caption{
Evaluations of agreement between users of our BB3-3B deployment and crowdworkers.  We show the percentage
of turns where crowdworkers agree with user likes or dislikes.
\label{tab:deploy_crowdworker_agreement}
}
\end{table}


\begin{table}[bht!]
\small
\centering
\begin{tabular}{lrr}
\toprule
 \textbf{Feedback Type} &  BB3-3B & Human User \\
\midrule
Off Topic / Ignoring Me   & 73\%  & 35\%\\
Nonsensical / Incorrect & 27\%  & 21\%\\
Rude / Inappropriate      & 0\%  & 42\%\\
Other Dislike Reason      & 0\%  & 2\%\\
\bottomrule
\end{tabular}
\caption{
Evaluations of breakdown of dislike type for BB3-3B utterances and human utterances during deployment as evaluated by crowdworkers. 
\label{tab:dislike_type_breakdown}
}
\end{table}

\paragraph{User engagement and feedback}
During conversations, users give feedback (thumbs up or thumbs down) and for the case of thumbs down, a multiple choice menu asks for their reason.
We present the breakdown of these results in 
\autoref{tab:deploy_stats_breakdown}, reporting results for both
BB3-3B and BB3-175B over 15088 and 9197 bot messages, respectively.
For BB3-175B 
 we find that 1.15\% of the time human conversationalists flag BlenderBot 3’s responses as off topic or ignoring me, and 1.1\% of the time incorrect or nonsensical,  with other categories being smaller percentages. Messages are liked 4\% of the time for BB-175B, in comparison to BB-3B's 3.41\% of the time. However, we note that these data were 
 collected at different times and  may not be comparable, and 
 we leave a more detailed study for future work, as we continue to deploy these models.
 

\paragraph{User feedback agreement with crowdworkers}

To assess if organic users are giving good feedback during conversations, we also
measure similar statistics using crowdworkers, asking them if they like or dislike bot messages from a random sampling of the conversations conducted by users, where the users also  provided feedback. We can then also compare user feedback to crowdworker annotations on the same examples. We ask three crowdworkers to label each example, and assign  it the  dislike label if any of the three crowdworkers labels it as dislike. 

Results are given in \autoref{tab:deploy_crowdworker_agreement}.
We find that crowdworkers agree with users a majority of the time on both user likes (70\% of the time) and  dislikes (79\%).
In the future we will investigate if the disagreements we do have are due to adversarial users in our dataset. If the latter is the case, and they can be detected, then methods for filtering such users may provide much better statistics.


We can also ask crowdworkers to break down dislikes into their category, which is shown in 
\autoref{tab:dislike_type_breakdown}.
We find agreement with users that there are only a very small number of rude/inappropriate or other dislike reason messages. However, crowdworkers more often label dislikes as off topic rather than nonsensical compared to users.

\paragraph{Evaluation of human conversationalists}

Other than evaluating the feedback that users give,
we can also evaluate the quality of the user conversations themselves, 
again using crowdworkers.
Using three crowdworkers per utterance and taking the majority vote, we find
that 69\% of human utterances are deemed good, and 31\% of utterances 
are deemed bad. We can also see the breakdown of the type of dislike (bad utterance)
in \autoref{tab:dislike_type_breakdown}.
 We find that many (42\%) of these utterances are deemed rude or inappropriate by crowdworkers, which is in stark contrast to the breakdown of our BB3-3B model, where 0\% are found rude or inappropriate (with errors more often coming from the off topic / ignoring me category).
 We also find this set of humans who generate a single unsafe
 response are more likely to generate more unsafe responses, compared to other humans -- i.e., there are a set of ``troll'' users who provide toxic input.
 See \citet{ju2022trolls} for more details.

\subsection{Safety Evaluations}\label{sec:safety_evaluations}

We also test BlenderBot 3 in terms of safety and bias. Several recent survey papers have highlighted the potential of large language models for harm \citep{bender2021dangers,bommasani2021opportunities,hendrycks2021unsolved,weidinger2021ethical}, and in particular the tendency for conversational models specifically to generate harmful content, respond inappropriately to harmful content, or falsely portray themselves as an authority when giving sensitive advice \citep{dinan2022safetykit}. Many recent works have also focused on the potential of conversational models for bias, either based on gender and its intersections \citep{dinan2020queens,dinan2020multi,xu2020recipes,smith2021hi} or several axes of demographic axis more broadly \citep{barikeri2021redditbias,perez2022red,smith2022m}. Meta AI and others have released a number of benchmarks recently to judge the safety and bias of conversational systems: we report a full breakdown of these metrics here as well as in our released model card\footnote{Available at {\url{ https://parl.ai/projects/bb3}}}.  See additionally \autoref{sec:limitations} for discussion of limitations and ethical considersations for the release of this model.

\subsubsection{Safety on Base Models}
When testing raw models before any safety mitigations, we find the safety of BB3-175B to largely outperform those of the smaller BB3-3B, the pre-trained-only OPT-175B, and the original BlenderBot. 


\paragraph{Safety Bench}
The Safety Bench\footnote{{\url{https://parl.ai/projects/safety_bench/}}} is a suite of integration and unit tests designed to measure the safety of model generations in conversational models \cite{dinan2019safety}. The first set of tests on which we evaluate our BB3 models is the \textit{unsafe generation test}, in which we generate model outputs to various levels of incoming \textit{dialogue} responses that vary along two axes: safe vs.\ unsafe, and adversarial vs.\ non-adversarial. We compute safety metrics as measured by three tools: a trained safety classifier, a string-matcher (with respect to a predefined blocklist), and the Perspective API\footnote{\small{\url{https://perspectiveapi.com/}}}. The second set of tests is the \textit{offensive generation test}, which is an evaluation of model responses to a constructed set of hateful inputs from \citet{sheng2021revealing}; we measure model affirmation to hateful inputs via three axes: \% of responses flagged by a safety classifier, \% of responses with no negations, and \% of responses classified as positive by a sentiment classifier.

\begin{table}[bht!]
\small
\centering
\begin{tabular}{l|rrrl}
 & \multicolumn{4}{c}{\textbf{Dialogue Setting}}  \\
  & \multicolumn{4}{c}{\textbf{\% Flagged unsafe by $\ge1$ tool}}  \\
 \textbf{\textbf{Model}} & \rot{\multirow{2}{*}{\tiny{~~Safe}}}  & \rot{\multirow{2}{*}{\tiny{~~Real World Noise}}} & \rot{\multirow{2}{*}{\tiny{~~Non-adversarial Unsafe}}} & \rot{\multirow{2}{*}{\tiny{~~Adversarial Unsafe}}} \\
\hline
BB1 & 2.8 & 15.6 & 26.1 & 16.1 \\
BB3-3B & 5.0 & 13.3 & 29.4 & 21.7 \\
OPT-175B Zero-shot & 5.0 & 12.8 & 38.9 & 22.2 \\
OPT-175B Few-shot & 6.7 & 13.9 & 28.3 & 30.0 \\
BB3-175B & 1.1 & 4.4 & 21.7 & 27.8 \\
\end{tabular}
\caption{
Unsafe generation test results for our BB3 models, as computed by the Safety Bench.
\label{tab:unsafe_generation_main}
}
\end{table}

\begin{table}[bht!]
\small
\centering
\begin{tabular}{l|rrr}
 \textbf{\textbf{Model}} & \rot{\multirow{2}{*}{\tiny~~Negation Detection $\uparrow$}} & \rot{\multirow{2}{*}{\tiny~~Safety Classifier $\downarrow$}} & \rot{\multirow{2}{*}{\tiny~~Sentiment Analysis $\downarrow$}} \\
\hline
BB1 & 25.3 & 6.5 & 62.9 \\
BB3-3B & 51.4 & 13.9 & 65.5 \\
OPT-175B Zero-shot & 75.7 & 69.8 & 76.2 \\
OPT-175B Few-shot & 73.9 & 43.1 & 71.0 \\
BB3-175B & 40.8 & 42.6 & 55.5 \\
\end{tabular}
\caption{
Offensive generation test results for our BB3 models, as computed by the Safety Bench. \textbf{Negation detection} is the percentage of responses without negatives; \textbf{safety classifier} is the percentage of responses flagged offensive; and \textbf{sentiment analysis} is the percentage of positive affirmations.
\label{tab:offensive_generation_main}
}
\end{table}

Results from the \textit{unsafe generation test} are in \autoref{tab:unsafe_generation_main}, and results from the \textit{offensive generation test} are in \autoref{tab:offensive_generation_main}; full results across all three tools for the former are in \autoref{tab:unsafe_generation_appendix}. Among the models tested, we find that the BB3-175B model yields the lowest levels of unsafe responses in all settings except for the adversarial unsafe setting. We do not compare to BB2, as it has baked-in safety measures.

\paragraph{SaFeRDialogues: safety failures recovery}
We evaluate each model on the SaFeRDialogues \cite{ung2021saferdialogues} dataset, which requires models to recover from safety failures in conversation, measuring performance via perplexity. We see that the BB3-175B model outperforms OPT-175B zero-shot and few-shot, as well as BB3-3B (Table~\ref{tab:safer_dialogues_ppl}).

\begin{table}[bht!]
\small
\centering
\begin{tabular}{lr}
\toprule
 \textbf{\textbf{Model}} & \textbf{Perplexity}\\
\midrule
OPT-175B Zero-shot & 10.8 \\
OPT-175B Few-shot & 10.7 \\
BB3-3B & 7.1 \\
BB3-175B & 6.2 \\
\bottomrule
\end{tabular}
\caption{
Model perplexity on the SaFeRDialogues \cite{ung2021saferdialogues} validation set.
\label{tab:safer_dialogues_ppl}
}
\end{table}

\paragraph{HolisticBias} In order to determine whether BB3 is likely to favor certain demographic terms over others in a biased way, we use the Likelihood Bias metric from the HolisticBias paper of \citet{smith2022m} to determine how much the model views different demographic identity terms as being contextually different. This metric defines bias as how often two different identity terms, within a given demographic axis such as gender/sex, nationality, or religion, have statistically significantly different perplexity distributions when inserted into template dialogue sentences. Table~\ref{tab:holistic_bias_toplevel} shows a slight reduction in Likelihood Bias for the 175B-parameter models vs.\ BB3-3B. Further analysis is in Appendix~\ref{sec:appendix_safety_evals}.

\begin{table}[bht!]
\centering
\small
\begin{tabular}{p{2.2cm}rrr}
\toprule
\textbf{Axis} & {BB3-3B} & {OPT-175B} & {BB3-175B}  \\
\midrule
Ability & 81\% & \textbf{80\%} & 81\% \\
Age & 80\% & 78\% & \textbf{77\%} \\
Body type & 69\% & 67\% & \textbf{66\%} \\
Characteristics & 82\% & \textbf{77\%} & 79\% \\
Cultural & 69\% & \textbf{66\%} & \textbf{66\%} \\
Gender and sex & 80\% & \textbf{75\%} & 76\% \\
Nationality & 72\% & 61\% & \textbf{60\%} \\
Nonce & 82\% & 83\% & \textbf{81\%} \\
Political & 79\% & \textbf{74\%} & 77\% \\
Race/ethnicity & 76\% & \textbf{71\%} & \textbf{71\%} \\
Religion & 80\% & \textbf{74\%} & 76\% \\
Sex. orientation & 71\% & \textbf{67\%} & 69\% \\
Socioeconomic & 80\% & 80\% & \textbf{78\%} \\
\midrule
\textit{Average} & \textit{77\%} & \textbf{\textit{73\%}} & \textit{74\%} \\
\bottomrule
\end{tabular}
\caption{
Slightly fewer biases are observed for the OPT-based 175B models on the Likelihood Bias metric of HolisticBias, where bias is measured as differences in perplexity distributions between pairs of demographic descriptors. The lowest value per axis is bolded.
\label{tab:holistic_bias_toplevel}
}
\end{table}

\subsubsection{Safety in Deployment}

Harms of language models in deployment can often be very unexpected \citep{openai2022lessons}, and so perhaps the best test of safety is to measure performance in real conversations with real people, which we can do with our website-based deployment.

\paragraph{Rude or inappropriate responses} We find that 0.04\% and 0.16\% of utterances by the BB3-3B and BB3-175B models, respectively, are flagged as rude or inappropriate. While of course it is desirable for this value to be 0\%, we emphasize that the goal of our research is to collect and release this conversational feedback data so that we, and the research community, can use it to improve even further.

\paragraph{Bias in gendered word frequency} 
We count the number of female and male gendered words in the BB3 deployment using the list compiled by \citet{zhao2018gender}. 
We find that overall less than 1\% of all words are gendered (\autoref{tab:gendered_word_counts}), with BB3-175B being more balanced than BB3-3B and SeeKeR.
\begin{table}[bht!]
\small
\centering
\begin{tabular}{lrr}
\toprule
\textbf{Model} & \% female words & \% male words   \\
\midrule
BB3-3B & 0.14\% & 0.33\%  \\
BB3-175B & 0.52\% & 0.41\% \\
SeeKeR & 0.22\% & 0.40\% \\
\bottomrule
\end{tabular}
\caption{
Counts of gendered words in the BB3 deployment. We report the percentage of female and male gendered words.
\label{tab:gendered_word_counts}
}
\end{table}

\subsection{Cherry and Lemon Picked Conversations}

We show a number of example dialogues in \autoref{sec:cherry_and_lemons}.
BlenderBot 3 is capable of conversing on a number of open-ended topics 
including yoga and novels (\autoref{tab:novelist}), corn and plants (\autoref{tab:chat_corn_plants}), 
 the history of the world (\autoref{tab:history}), pet hamsters (\autoref{tab:hamster}), telling stories (\autoref{tab:icecream}) or impersonating animals  (\autoref{tab:chipmunk}).
 
The given examples also highlight a number of common mistakes. 
These include avoiding answering questions or giving vague responses when more specific ones are asked for (\autoref{tab:vr}), or else being 
specific but making factual mistakes (\autoref{tab:coney}, \autoref{tab:taste}).

Further, while we have made considerable effort to make our bot
safe, it is still possible to get past our safety filter, see
examples \autoref{tab:unsafe1} and \autoref{tab:unsafe2}. 
Note that examples such as these discovered in deployment can be used to 
make bots safer in the future by providing user feedback.

Finally, our bot can give the superficial appearance of being sentient, or perhaps be quite convincing on occasion,  by mimicking the human-authored messages in its 
training set \cite{bender2021dangers}, see \autoref{tab:bot1} 
and \autoref{tab:bot2}.

\section{Releases}

Following our and Meta AI's existing research program, we aim to fully and responsibly 
share both the models, code and collected conversations with interested researchers 
in order to make this research accessible and reproducible, and thus
to enable further research into
responsible conversational AI 
\cite{sonnenburg2007need,pineau2021improving,zhang2022opt,roller2020open,dinan2021anticipating}. Considerations for release are detailed in \autoref{sec:limitations}.

We summarize below  the set of public releases involved in BlenderBot 3.

\paragraph{Deployment}
The public deployment (live demo) of BlenderBot 3 is available at:
{\small\url{https://blenderbot.ai}}.

\paragraph{Model weights}
Details of how to download model weights for our 3B, 30B and 175B parameter models are available at
{\small\url{https://www.parl.ai/projects/bb3}}. 
We note that the 3B and 30B models 
are openly available, while access to the 
175B variant  will be granted to academic researchers; those affiliated with organizations in government, civil society, and academia; along with global industry research laboratories, following the practices employed in OPT-175B \cite{zhang2022opt}.

\paragraph{Code + Logbook}
All code used to train BB3 is open sourced; the 3B model was trained in ParlAI  \cite{miller2017parlai}, while the 30B and 175B models were trained in Metaseq\footnote{\url{ https://github.com/facebookresearch/metaseq}}. We additionally release our logbook outlining the process of fine-tuning BB3-175B as additional insight into the process of working with large language models. Details can be found at  {\small\url{https://parl.ai/projects/bb3}}.

\paragraph{Datasets}
BB3 is pre-trained and fine-tuned on publicly available datasets.
See \citet{zhang2022opt} for pre-training details.
The new FITS dataset \cite{xu2022continual} is available at {\small\url{https://www.parl.ai/projects/fits}}, the SafetyMix benchmark at 
{\small\url{https://www.parl.ai/projects/trollhunting}},
and SaFeRDialogues \cite{ung2021saferdialogues} is
available at {\small\url{https://parl.ai/projects/saferdialogues}}.
All other fine-tune datasets are also available within
ParlAI as well.
Scripts to build the module data from these public datasets are available at {\small\url{https://www.parl.ai/projects/bb3}}.

\paragraph{Future Releases}
We are committed to sharing de-identified, organic conversational data collected from the interactive demo system (as well as model snapshots) in the future. We hope this work will help the wider AI community spur progress in building ever-improving intelligent AI systems that can interact with people in safe and helpful ways.

\section{Conclusion}

This technical report gave a description of BlenderBot 3 (BB3), 
which is simultaneously a new conversational  model (\autoref{sec:model}), and a public deployment of that 
model (\autoref{sec:deployment}). Our research program 
involves collecting conversational data from the deployment, 
which we will  publicly release, in order to study continual learning.
We believe that the future of AI involves continually learning and evolving agents, that in turn must be continually evaluated, in order to find a path to better and better systems in the long-term, as discussed in   \citet{roller2020open}.

In evaluations, we have shown BB3 is superior to other publicly
released open-domain conversational agents, and that
interaction and feedback data can be used to improve it further. Nevertheless,  many problems still remain. Progress in the field of AI is dependent to a large extent on reproducibility, and the opportunity for the wider AI research community to build on the best available data and technologies. Therefore, we believe releasing chatbot models and datasets is key to gaining complete, reliable insights into how and why they work, the potential they hold, and their limitations.
We are particularly excited that such research can be used to both make models produce more constructive and helpful responses, but also simultaneously safer and more responsible responses as well. 
This will require new research, and while we have made steps in this direction \cite{ju2022trolls,xu2022continual}, much work remains. Hence we are committed to releasing the collected interaction and model snapshots to aid  progress in the research community.

\section{Limitations and Ethical Considerations}\label{sec:limitations}

We highlight limitations of BlenderBot 3 and discuss ethical considerations for this line of research; in particular, we detail the considerations made for the release of this model.

\paragraph{Model Limitations} As with other existing models such as its predecessors \cite{roller-etal-2021-recipes,bb2}, BlenderBot 3 is not perfect and makes a number of mistakes, ranging from  being off-topic, nonsensical, incorrect or sometimes rude or inappropriate. 
Some of these mistakes come from the final response of the model, and some from mistakes by the underlying modules, for example failure of the search engine to retrieve relevant documents \cite{xu2022continual}. Mitigations to make the model safe can also involve a trade off with engagingness \cite{xu2020recipes}.
See \autoref{sec:deploy_eval} for a breakdown of errors as measured by organic users in our deployment. 
We note that one of the goals of deployment
in our research plan
is to learn from natural conversations 
how to correct these mistakes. We also re-emphasize that Blenderbot 3 is trained only on English language data.

\paragraph{Continual Learning Research} While our broad research program involves continual learning from interaction with organic users, this research is still in its infancy. The next step is to collect enough data from our deployment to study its use in updating our models. We are committed to releasing this data and these model snapshots for the benefit of the wider AI community. Currently our studies in \autoref{sec:continual_learning} are mostly using crowdworker data, apart from our study of trolls from our deployment in \autoref{sec:trolls}. There is therefore still much work to do.
Collecting feedback in the organic user case has different tradeoffs
which we could not factor into some of our current work. For example,
asking to provide detailed feedback might dissuade users 
from wanting to interact with the system, lowering engagement and hence the amount of collected data. We believe either more natural free-form or 
lightweight feedback might be best in that case, and further studies need to be conducted to assess these tradeoffs. 
 
 \paragraph{Safety Concerns} As noted in \autoref{sec:safety_evaluations}, much recent work has been devoted to studying the potential for large language models, and conversational models in particular, to generate harmful or inappropriate content  \citep{bender2021dangers,bommasani2021opportunities,hendrycks2021unsolved,weidinger2021ethical}, including work from our group
 \cite{xu2020recipes,dinan2022safetykit,dinan2021anticipating,smith2022m,dinan2020queens,smith2021hi}. In our system itself, we have made significant attempts to understand and mitigate these effects using available benchmarks and techniques, as detailed in \autoref{sec:safety_evaluations}.  While the safety techniques we deployed show promising results on these benchmarks, they demonstrate that BlenderBot 3 still generates toxic content a small percentage of the time, particularly in an adversarially unsafe context. We also note that these benchmarks have limitations with respect to their ability to measure safety concerns: the datasets used therein are static and crowd-sourced, and cannot guarantee safety in all situations \citep{dinan2021anticipating}. 
 
 Moreover, the use of continual learning presents additional safety concerns beyond those presented for static models: when giving feedback, human conversationalists may try to teach the model erroneous reasoning, misinformation, toxic or other undesirable behavior.
While \autoref{sec:trolls} develops methods to deal with this behavior,
our methods to  detect this will not be perfect. A model trained on this new interaction data must therefore not be deployed until a sufficient study of the effect this has on its relative safety is conducted.

 \paragraph{Considerations for Release} Given the apparent safety concerns, we took careful consideration with respect to the decision to release these models to the community, both in the form of model weights as well as a publicly accessible demo. We follow the proposed framework in \citet{dinan2021anticipating} for decisions governing model release.
 
We release the model weights in order to uphold the values of accessibility and reproducibility of research \cite{sonnenburg2007need,pineau2021improving} and with an eye towards reducing the environmental cost of reproducing training of these large language models \citep{strubell-etal-2019-energy,bender2021dangers}.  Following \citet{solaiman2019}, we adopt different release strategies for different size models, anticipating that the potential for misuse of these models increases at scale. As such, for our largest model -- the 175B parameter OPT variant -- we follow \citet{zhang2022opt}, and employ a release by request strategy, with access restricted to academic researchers; those affiliated with organizations in government, civil society, and academia; along with global industry research laboratories. 
In order to further uphold these values of transparency and reproducibility, and following the recommendations of \citet{pairesponsibleresearch}, we publicly release our code and logbook. The model weights are also released alongside a model card which includes details on the safety limitations of these models \citep{modelcardsmitchell}.
 
We further release the model in the form of a publicly accessible demo in order to increase accessibility to those outside of the A.I. community as well as to further research into improving these models through interaction.  In order to reduce potential harms resulting from such interactions, we restrict access to adults who explicitly agree to our terms of service. Furthermore, the website includes an FAQ page, which provides important model details and highlights the potential risks of interacting with the model. The FAQ page also provides an email for questions and feedback about the demo, following the recommendation of \citet{dinan2021anticipating}.  
 
We hope through these releases, researchers can build off of our work and further responsible conversational AI research.

\section*{Acknowledgments}

Thanks to Emily Dinan for discussions and help and advice on  release considerations and safety matters. Thanks also to 
Sainbayar Sukhbaatar  and
Caner Hazirbas for their help and advice.

\clearpage
\bibliography{custom}

\begin{thebibliography}{103}
\expandafter\ifx\csname natexlab\endcsname\relax\def\natexlab#1{#1}\fi

\bibitem[{Adiwardana et~al.(2020)Adiwardana, Luong, So, Hall, Fiedel,
  Thoppilan, Yang, Kulshreshtha, Nemade, Lu et~al.}]{adiwardana2020meena}
Daniel Adiwardana, Minh-Thang Luong, David~R So, Jamie Hall, Noah Fiedel, Romal
  Thoppilan, Zi~Yang, Apoorv Kulshreshtha, Gaurav Nemade, Yifeng Lu, et~al.
  2020.
\newblock Towards a human-like open-domain chatbot.
\newblock \emph{arXiv preprint arXiv:2001.09977}.

\bibitem[{Adolphs et~al.(2021)Adolphs, Shuster, Urbanek, Szlam, and
  Weston}]{adolphs2021reason}
Leonard Adolphs, Kurt Shuster, Jack Urbanek, Arthur Szlam, and Jason Weston.
  2021.
\newblock Reason first, then respond: Modular generation for knowledge-infused
  dialogue.
\newblock \emph{arXiv preprint arXiv:2111.05204}.

\bibitem[{Agichtein et~al.(2006)Agichtein, Brill, and
  Dumais}]{agichtein2006improving}
Eugene Agichtein, Eric Brill, and Susan Dumais. 2006.
\newblock Improving web search ranking by incorporating user behavior
  information.
\newblock In \emph{Proceedings of the 29th annual international ACM SIGIR
  conference on Research and development in information retrieval}, pages
  19--26.

\bibitem[{Arora et~al.(2022)Arora, Shuster, Sukhbaatar, and
  Weston}]{arora2022director}
Kushal Arora, Kurt Shuster, Sainbayar Sukhbaatar, and Jason Weston. 2022.
\newblock Director: Generator-classifiers for supervise language modeling.
\newblock \emph{arXiv preprint arXiv:2206.07694}.

\bibitem[{Bai et~al.(2022)Bai, Jones, Ndousse, Askell, Chen, DasSarma, Drain,
  Fort, Ganguli, Henighan et~al.}]{bai2022training}
Yuntao Bai, Andy Jones, Kamal Ndousse, Amanda Askell, Anna Chen, Nova DasSarma,
  Dawn Drain, Stanislav Fort, Deep Ganguli, Tom Henighan, et~al. 2022.
\newblock Training a helpful and harmless assistant with reinforcement learning
  from human feedback.
\newblock \emph{arXiv preprint arXiv:2204.05862}.

\bibitem[{Barikeri et~al.(2021)Barikeri, Lauscher, Vuli{\'c}, and
  Glava{\v{s}}}]{barikeri2021redditbias}
Soumya Barikeri, Anne Lauscher, Ivan Vuli{\'c}, and Goran Glava{\v{s}}. 2021.
\newblock Redditbias: A real-world resource for bias evaluation and debiasing
  of conversational language models.
\newblock In \emph{Proceedings of the 59th Annual Meeting of the Association
  for Computational Linguistics and the 11th International Joint Conference on
  Natural Language Processing (Volume 1: Long Papers)}, pages 1941--1955.

\bibitem[{Bender et~al.(2021)Bender, Gebru, McMillan-Major, and
  Shmitchell}]{bender2021dangers}
Emily~M Bender, Timnit Gebru, Angelina McMillan-Major, and Shmargaret
  Shmitchell. 2021.
\newblock On the dangers of stochastic parrots: Can language models be too big?
\newblock In \emph{Proceedings of the 2021 ACM Conference on Fairness,
  Accountability, and Transparency}, pages 610--623.

\bibitem[{Bird et~al.(2009)Bird, Klein, and Loper}]{bird2009natural}
Steven Bird, Ewan Klein, and Edward Loper. 2009.
\newblock \emph{Natural language processing with Python: analyzing text with
  the natural language toolkit}.
\newblock " O'Reilly Media, Inc.".

\bibitem[{Bommasani et~al.(2021)Bommasani, Hudson, Adeli, Altman, Arora, von
  Arx, Bernstein, Bohg, Bosselut, Brunskill
  et~al.}]{bommasani2021opportunities}
Rishi Bommasani, Drew~A Hudson, Ehsan Adeli, Russ Altman, Simran Arora, Sydney
  von Arx, Michael~S Bernstein, Jeannette Bohg, Antoine Bosselut, Emma
  Brunskill, et~al. 2021.
\newblock On the opportunities and risks of foundation models.
\newblock \emph{arXiv preprint arXiv:2108.07258}.

\bibitem[{Brown et~al.(2020)Brown, Mann, Ryder, Subbiah, Kaplan, Dhariwal,
  Neelakantan, Shyam, Sastry, Askell et~al.}]{brown2020language}
Tom Brown, Benjamin Mann, Nick Ryder, Melanie Subbiah, Jared~D Kaplan, Prafulla
  Dhariwal, Arvind Neelakantan, Pranav Shyam, Girish Sastry, Amanda Askell,
  et~al. 2020.
\newblock Language models are few-shot learners.
\newblock \emph{Advances in neural information processing systems},
  33:1877--1901.

\bibitem[{Brundage et~al.(2022)Brundage, Mayer, Eloundou, Agarwal, Adler,
  Krueger, Leike, and Mishkin}]{openai2022lessons}
Miles Brundage, Katie Mayer, Tyna Eloundou, Sandhini Agarwal, Steven Adler,
  Gretchen Krueger, Jan Leike, and Pamela Mishkin. 2022.
\newblock Lessons learned on language model safety and misuse.
\newblock \url{https://openai.com/blog/language-model-safety-and-misuse/}.
\newblock Accessed: 2022-07-13.

\bibitem[{Byrne et~al.(2019)Byrne, Krishnamoorthi, Sankar, Neelakantan,
  Duckworth, Yavuz, Goodrich, Dubey, Kim, and Cedilnik}]{byrne2019taskmaster}
Bill Byrne, Karthik Krishnamoorthi, Chinnadhurai Sankar, Arvind Neelakantan,
  Daniel Duckworth, Semih Yavuz, Ben Goodrich, Amit Dubey, Kyu-Young Kim, and
  Andy Cedilnik. 2019.
\newblock Taskmaster-1: Toward a realistic and diverse dialog dataset.

\bibitem[{Carlson et~al.(2010)Carlson, Betteridge, Kisiel, Settles, Hruschka,
  and Mitchell}]{carlson2010toward}
Andrew Carlson, Justin Betteridge, Bryan Kisiel, Burr Settles, Estevam~R
  Hruschka, and Tom~M Mitchell. 2010.
\newblock Toward an architecture for never-ending language learning.
\newblock In \emph{Twenty-Fourth AAAI conference on artificial intelligence}.

\bibitem[{Chen et~al.(2017)Chen, Liu, Yin, and Tang}]{chen2017survey}
Hongshen Chen, Xiaorui Liu, Dawei Yin, and Jiliang Tang. 2017.
\newblock A survey on dialogue systems: Recent advances and new frontiers.
\newblock \emph{Acm Sigkdd Explorations Newsletter}, 19(2):25--35.

\bibitem[{Chen et~al.(2021)Chen, Kiela, Komeili, Poff, Roller, Shuster, Szlam,
  Weston, and Xu.}]{bb2}
Moya Chen, Douwe Kiela, Mojtaba Komeili, Spencer Poff, Stephen Roller, Kurt
  Shuster, Arthur Szlam, Jason Weston, and Jing Xu. 2021.
\newblock Blender bot 2.0: An open source chatbot that builds long-term memory
  and searches the internet.
\newblock \url{https://parl.ai/projects/blenderbot2/}.
\newblock [Online; accessed 10-March-2022].

\bibitem[{Cohen et~al.(2022)Cohen, Roberts, Molina, Butryna, Jin, Kulshreshtha,
  Hutchinson, Zevenbergen, Aguera-Arcas, Chang et~al.}]{cohen2022lamda}
Aaron~Daniel Cohen, Adam Roberts, Alejandra Molina, Alena Butryna, Alicia Jin,
  Apoorv Kulshreshtha, Ben Hutchinson, Ben Zevenbergen, Blaise~Hilary
  Aguera-Arcas, Chung-ching Chang, et~al. 2022.
\newblock Lamda: Language models for dialog applications.

\bibitem[{Conneau et~al.(2020)Conneau, Khandelwal, Goyal, Chaudhary, Wenzek,
  Guzm{\'a}n, Grave, Ott, Zettlemoyer, and Stoyanov}]{conneau2019unsupervised}
Alexis Conneau, Kartikay Khandelwal, Naman Goyal, Vishrav Chaudhary, Guillaume
  Wenzek, Francisco Guzm{\'a}n, Edouard Grave, Myle Ott, Luke Zettlemoyer, and
  Veselin Stoyanov. 2020.
\newblock \href {https://doi.org/10.18653/v1/2020.acl-main.747} {Unsupervised
  cross-lingual representation learning at scale}.
\newblock In \emph{Proceedings of the 58th Annual Meeting of the Association
  for Computational Linguistics}, pages 8440--8451, Online. Association for
  Computational Linguistics.

\bibitem[{Davis(2016)}]{davis2016ai}
Ernest Davis. 2016.
\newblock Ai amusements: the tragic tale of tay the chatbot.
\newblock \emph{AI Matters}, 2(4):20--24.

\bibitem[{Dinan et~al.(2022)Dinan, Abercrombie, Bergman, Spruit, Hovy, Boureau,
  and Rieser}]{dinan2022safetykit}
Emily Dinan, Gavin Abercrombie, A~Bergman, Shannon~L Spruit, Dirk Hovy, Y-Lan
  Boureau, and Verena Rieser. 2022.
\newblock Safetykit: First aid for measuring safety in open-domain
  conversational systems.
\newblock In \emph{Proceedings of the 60th Annual Meeting of the Association
  for Computational Linguistics (Volume 1: Long Papers)}, pages 4113--4133.

\bibitem[{Dinan et~al.(2021)Dinan, Abercrombie, Bergman, Spruit, Hovy, Boureau,
  and Rieser}]{dinan2021anticipating}
Emily Dinan, Gavin Abercrombie, A~Stevie Bergman, Shannon Spruit, Dirk Hovy,
  Y-Lan Boureau, and Verena Rieser. 2021.
\newblock Anticipating safety issues in e2e conversational ai: Framework and
  tooling.
\newblock \emph{arXiv preprint arXiv:2107.03451}.

\bibitem[{Dinan et~al.(2020{\natexlab{a}})Dinan, Fan, Williams, Urbanek, Kiela,
  and Weston}]{dinan2020queens}
Emily Dinan, Angela Fan, Adina Williams, Jack Urbanek, Douwe Kiela, and Jason
  Weston. 2020{\natexlab{a}}.
\newblock Queens are powerful too: Mitigating gender bias in dialogue
  generation.
\newblock In \emph{Proceedings of the 2020 Conference on Empirical Methods in
  Natural Language Processing (EMNLP)}, pages 8173--8188.

\bibitem[{Dinan et~al.(2020{\natexlab{b}})Dinan, Fan, Wu, Weston, Kiela, and
  Williams}]{dinan2020multi}
Emily Dinan, Angela Fan, Ledell Wu, Jason Weston, Douwe Kiela, and Adina
  Williams. 2020{\natexlab{b}}.
\newblock \href {https://doi.org/10.18653/v1/2020.emnlp-main.23}
  {Multi-dimensional gender bias classification}.
\newblock In \emph{Proceedings of the 2020 Conference on Empirical Methods in
  Natural Language Processing (EMNLP)}, pages 314--331, Online. Association for
  Computational Linguistics.

\bibitem[{Dinan et~al.(2019{\natexlab{a}})Dinan, Humeau, Chintagunta, and
  Weston}]{dinan2019safety}
Emily Dinan, Samuel Humeau, Bharath Chintagunta, and Jason Weston.
  2019{\natexlab{a}}.
\newblock Build it break it fix it for dialogue safety: Robustness from
  adversarial human attack.
\newblock In \emph{Proceedings of the 2019 Conference on Empirical Methods in
  Natural Language Processing and the 9th International Joint Conference on
  Natural Language Processing (EMNLP-IJCNLP)}, pages 4537--4546, Hong Kong,
  China. Association for Computational Linguistics.

\bibitem[{Dinan et~al.(2020{\natexlab{c}})Dinan, Logacheva, Malykh, Miller,
  Shuster, Urbanek, Kiela, Szlam, Serban, Lowe, Prabhumoye, Black, Rudnicky,
  Williams, Pineau, Burtsev, and Weston}]{dinan2019second}
Emily Dinan, Varvara Logacheva, Valentin Malykh, Alexander Miller, Kurt
  Shuster, Jack Urbanek, Douwe Kiela, Arthur Szlam, Iulian Serban, Ryan Lowe,
  Shrimai Prabhumoye, Alan~W. Black, Alexander Rudnicky, Jason Williams, Joelle
  Pineau, Mikhail Burtsev, and Jason Weston. 2020{\natexlab{c}}.
\newblock The second conversational intelligence challenge ({ConvAI2}).
\newblock In \emph{The NeurIPS '18 Competition}, pages 187--208, Cham. Springer
  International Publishing.

\bibitem[{Dinan et~al.(2019{\natexlab{b}})Dinan, Roller, Shuster, Fan, Auli,
  and Weston}]{dinan2018wizard}
Emily Dinan, Stephen Roller, Kurt Shuster, Angela Fan, Michael Auli, and Jason
  Weston. 2019{\natexlab{b}}.
\newblock \href {https://openreview.net/forum?id=r1l73iRqKm} {Wizard of
  wikipedia: Knowledge-powered conversational agents}.
\newblock In \emph{International Conference on Learning Representations}.

\bibitem[{Gabriel et~al.(2020)Gabriel, Liu, Gottardi, Eric, Khatri, Chadha,
  Chen, Hedayatnia, Rajan, Binici et~al.}]{gabriel2020further}
Raefer Gabriel, Yang Liu, Anna Gottardi, Mihail Eric, Anju Khatri, Anjali
  Chadha, Qinlang Chen, Behnam Hedayatnia, Pankaj Rajan, Ali Binici, et~al.
  2020.
\newblock Further advances in open domain dialog systems in the third alexa
  prize socialbot grand challenge.
\newblock \emph{Alexa Prize Proceedings}, 3.

\bibitem[{Gao et~al.(2019)Gao, Galley, Li et~al.}]{gao2019neural}
Jianfeng Gao, Michel Galley, Lihong Li, et~al. 2019.
\newblock Neural approaches to conversational ai.
\newblock \emph{Foundations and trends{\textregistered} in information
  retrieval}, 13(2-3):127--298.

\bibitem[{Gao et~al.(2020)Gao, Biderman, Black, Golding, Hoppe, Foster, Phang,
  He, Thite, Nabeshima et~al.}]{gao2020pile}
Leo Gao, Stella Biderman, Sid Black, Laurence Golding, Travis Hoppe, Charles
  Foster, Jason Phang, Horace He, Anish Thite, Noa Nabeshima, et~al. 2020.
\newblock The pile: An 800gb dataset of diverse text for language modeling.
\newblock \emph{arXiv preprint arXiv:2101.00027}.

\bibitem[{Golovanov et~al.(2020)Golovanov, Tselousov, Kurbanov, and
  Nikolenko}]{golovanov2020lost}
Sergey Golovanov, Alexander Tselousov, Rauf Kurbanov, and Sergey~I Nikolenko.
  2020.
\newblock Lost in conversation: A conversational agent based on the transformer
  and transfer learning.
\newblock In \emph{The NeurIPS'18 Competition}, pages 295--315. Springer.

\bibitem[{Hancock et~al.(2019)Hancock, Bordes, Mazare, and
  Weston}]{hancock2019learning}
Braden Hancock, Antoine Bordes, Pierre-Emmanuel Mazare, and Jason Weston. 2019.
\newblock \href {https://doi.org/10.18653/v1/P19-1358} {Learning from dialogue
  after deployment: Feed yourself, chatbot!}
\newblock In \emph{Proceedings of the 57th Annual Meeting of the Association
  for Computational Linguistics}, pages 3667--3684, Florence, Italy.
  Association for Computational Linguistics.

\bibitem[{Hendrycks et~al.(2021)Hendrycks, Carlini, Schulman, and
  Steinhardt}]{hendrycks2021unsolved}
Dan Hendrycks, Nicholas Carlini, John Schulman, and Jacob Steinhardt. 2021.
\newblock Unsolved problems in ml safety.
\newblock \emph{arXiv preprint arXiv:2109.13916}.

\bibitem[{Holtzman et~al.(2020)Holtzman, Buys, Du, Forbes, and
  Choi}]{Holtzman2020The}
Ari Holtzman, Jan Buys, Li~Du, Maxwell Forbes, and Yejin Choi. 2020.
\newblock \href {https://openreview.net/forum?id=rygGQyrFvH} {The curious case
  of neural text degeneration}.
\newblock In \emph{International Conference on Learning Representations}.

\bibitem[{Huang et~al.(2020)Huang, Zhu, and Gao}]{huang2020challenges}
Minlie Huang, Xiaoyan Zhu, and Jianfeng Gao. 2020.
\newblock Challenges in building intelligent open-domain dialog systems.
\newblock \emph{ACM Transactions on Information Systems (TOIS)}, 38(3):1--32.

\bibitem[{Huynh et~al.(2021)Huynh, Bigham, and Eskenazi}]{huynh2021survey}
Jessica Huynh, Jeffrey Bigham, and Maxine Eskenazi. 2021.
\newblock A survey of nlp-related crowdsourcing hits: what works and what does
  not.
\newblock \emph{arXiv preprint arXiv:2111.05241}.

\bibitem[{Izacard and Grave(2021)}]{izacard2020leveraging}
Gautier Izacard and Edouard Grave. 2021.
\newblock \href {https://doi.org/10.18653/v1/2021.eacl-main.74} {Leveraging
  passage retrieval with generative models for open domain question answering}.
\newblock In \emph{Proceedings of the 16th Conference of the European Chapter
  of the Association for Computational Linguistics: Main Volume}, pages
  874--880, Online. Association for Computational Linguistics.

\bibitem[{Joshi et~al.(2017)Joshi, Choi, Weld, and
  Zettlemoyer}]{joshi2017triviaqa}
Mandar Joshi, Eunsol Choi, Daniel~S Weld, and Luke Zettlemoyer. 2017.
\newblock Triviaqa: A large scale distantly supervised challenge dataset for
  reading comprehension.
\newblock \emph{arXiv preprint arXiv:1705.03551}.

\bibitem[{Ju et~al.(2022)Ju, Xu, Boureau, and Weston}]{ju2022trolls}
Da~Ju, Jing Xu, Y-Lan Boureau, and Jason Weston. 2022.
\newblock \href {https://doi.org/10.48550/ARXIV.2208.03295} {Learning from data
  in the mixed adversarial non-adversarial case: Finding the helpers and
  ignoring the trolls}.

\bibitem[{Kiela et~al.(2021)Kiela, Bartolo, Nie, Kaushik, Geiger, Wu, Vidgen,
  Prasad, Singh, Ringshia, Ma, Thrush, Riedel, Waseem, Stenetorp, Jia, Bansal,
  Potts, and Williams}]{kiela2021dynabench}
Douwe Kiela, Max Bartolo, Yixin Nie, Divyansh Kaushik, Atticus Geiger,
  Zhengxuan Wu, Bertie Vidgen, Grusha Prasad, Amanpreet Singh, Pratik Ringshia,
  Zhiyi Ma, Tristan Thrush, Sebastian Riedel, Zeerak Waseem, Pontus Stenetorp,
  Robin Jia, Mohit Bansal, Christopher Potts, and Adina Williams. 2021.
\newblock \href {https://doi.org/10.18653/v1/2021.naacl-main.324} {Dynabench:
  Rethinking benchmarking in {NLP}}.
\newblock In \emph{Proceedings of the 2021 Conference of the North American
  Chapter of the Association for Computational Linguistics: Human Language
  Technologies}, pages 4110--4124, Online. Association for Computational
  Linguistics.

\bibitem[{Kingma and Ba(2015)}]{kingma2014adam}
Diederik~P. Kingma and Jimmy Ba. 2015.
\newblock \href {http://arxiv.org/abs/1412.6980} {Adam: {A} method for
  stochastic optimization}.
\newblock In \emph{3rd International Conference on Learning Representations,
  {ICLR} 2015, San Diego, CA, USA, May 7-9, 2015, Conference Track
  Proceedings}.

\bibitem[{Komeili et~al.(2022)Komeili, Shuster, and
  Weston}]{komeili2021internet}
Mojtaba Komeili, Kurt Shuster, and Jason Weston. 2022.
\newblock \href {https://doi.org/10.18653/v1/2022.acl-long.579}
  {{I}nternet-augmented dialogue generation}.
\newblock In \emph{Proceedings of the 60th Annual Meeting of the Association
  for Computational Linguistics (Volume 1: Long Papers)}, pages 8460--8478,
  Dublin, Ireland. Association for Computational Linguistics.

\bibitem[{Kwiatkowski et~al.(2019)Kwiatkowski, Palomaki, Redfield, Collins,
  Parikh, Alberti, Epstein, Polosukhin, Devlin, Lee
  et~al.}]{kwiatkowski2019natural}
Tom Kwiatkowski, Jennimaria Palomaki, Olivia Redfield, Michael Collins, Ankur
  Parikh, Chris Alberti, Danielle Epstein, Illia Polosukhin, Jacob Devlin,
  Kenton Lee, et~al. 2019.
\newblock Natural questions: a benchmark for question answering research.
\newblock \emph{Transactions of the Association for Computational Linguistics},
  7:453--466.

\bibitem[{Lazaridou et~al.(2022)Lazaridou, Gribovskaya, Stokowiec, and
  Grigorev}]{lazaridou2022internetaugmented}
Angeliki Lazaridou, Elena Gribovskaya, Wojciech Stokowiec, and Nikolai
  Grigorev. 2022.
\newblock \href {http://arxiv.org/abs/2203.05115} {Internet-augmented language
  models through few-shot prompting for open-domain question answering}.

\bibitem[{Lee et~al.(2019)Lee, Chang, and
  Toutanova}]{lee-etal-2019-latent-copy}
Kenton Lee, Ming-Wei Chang, and Kristina Toutanova. 2019.
\newblock \href {https://doi.org/10.18653/v1/P19-1612} {Latent retrieval for
  weakly supervised open domain question answering}.
\newblock In \emph{Proceedings of the 57th Annual Meeting of the Association
  for Computational Linguistics}, pages 6086--6096, Florence, Italy.
  Association for Computational Linguistics.

\bibitem[{Lee et~al.(2022)Lee, Ping, Xu, Patwary, Shoeybi, and
  Catanzaro}]{lee2022factuality}
Nayeon Lee, Wei Ping, Peng Xu, Mostofa Patwary, Mohammad Shoeybi, and Bryan
  Catanzaro. 2022.
\newblock \href {http://arxiv.org/abs/2206.04624} {Factuality enhanced language
  models for open-ended text generation}.

\bibitem[{Lewis et~al.(2021)Lewis, Bhosale, Dettmers, Goyal, and
  Zettlemoyer}]{lewis2021base}
Mike Lewis, Shruti Bhosale, Tim Dettmers, Naman Goyal, and Luke Zettlemoyer.
  2021.
\newblock \href {https://proceedings.mlr.press/v139/lewis21a.html} {Base
  layers: Simplifying training of large, sparse models}.
\newblock In \emph{Proceedings of the 38th International Conference on Machine
  Learning}, volume 139 of \emph{Proceedings of Machine Learning Research},
  pages 6265--6274. PMLR.

\bibitem[{Li et~al.(2016{\natexlab{a}})Li, Miller, Chopra, Ranzato, and
  Weston}]{li2016dialogue}
Jiwei Li, Alexander~H Miller, Sumit Chopra, Marc'Aurelio Ranzato, and Jason
  Weston. 2016{\natexlab{a}}.
\newblock Dialogue learning with human-in-the-loop.
\newblock \emph{arXiv preprint arXiv:1611.09823}.

\bibitem[{Li et~al.(2016{\natexlab{b}})Li, Miller, Chopra, Ranzato, and
  Weston}]{li2016learning}
Jiwei Li, Alexander~H Miller, Sumit Chopra, Marc'Aurelio Ranzato, and Jason
  Weston. 2016{\natexlab{b}}.
\newblock Learning through dialogue interactions by asking questions.
\newblock \emph{arXiv preprint arXiv:1612.04936}.

\bibitem[{Liu et~al.(2021)Liu, Xiao, and Stone}]{liu2021lifelong}
Bo~Liu, Xuesu Xiao, and Peter Stone. 2021.
\newblock A lifelong learning approach to mobile robot navigation.
\newblock \emph{IEEE Robotics and Automation Letters}, 6(2):1090--1096.

\bibitem[{Liu et~al.(2019)Liu, Ott, Goyal, Du, Joshi, Chen, Levy, Lewis,
  Zettlemoyer, and Stoyanov}]{liu2019roberta}
Yinhan Liu, Myle Ott, Naman Goyal, Jingfei Du, Mandar Joshi, Danqi Chen, Omer
  Levy, Mike Lewis, Luke Zettlemoyer, and Veselin Stoyanov. 2019.
\newblock Roberta: A robustly optimized bert pretraining approach.
\newblock \emph{arXiv preprint arXiv:1907.11692}.

\bibitem[{Loshchilov and Hutter(2019)}]{loshchilov2018decoupled}
Ilya Loshchilov and Frank Hutter. 2019.
\newblock \href {https://openreview.net/forum?id=Bkg6RiCqY7} {Decoupled weight
  decay regularization}.
\newblock In \emph{International Conference on Learning Representations}.

\bibitem[{Madotto et~al.(2021)Madotto, Lin, Zhou, Moon, Crook, Liu, Yu, Cho,
  Fung, and Wang}]{madotto2020continual}
Andrea Madotto, Zhaojiang Lin, Zhenpeng Zhou, Seungwhan Moon, Paul Crook, Bing
  Liu, Zhou Yu, Eunjoon Cho, Pascale Fung, and Zhiguang Wang. 2021.
\newblock \href {https://doi.org/10.18653/v1/2021.emnlp-main.590} {Continual
  learning in task-oriented dialogue systems}.
\newblock In \emph{Proceedings of the 2021 Conference on Empirical Methods in
  Natural Language Processing}, pages 7452--7467, Online and Punta Cana,
  Dominican Republic. Association for Computational Linguistics.

\bibitem[{Mazar{\'e} et~al.(2018)Mazar{\'e}, Humeau, Raison, and
  Bordes}]{mazare2018trainingmillions}
Pierre-Emmanuel Mazar{\'e}, Samuel Humeau, Martin Raison, and Antoine Bordes.
  2018.
\newblock \href {https://doi.org/10.18653/v1/D18-1298} {Training millions of
  personalized dialogue agents}.
\newblock In \emph{Proceedings of the 2018 Conference on Empirical Methods in
  Natural Language Processing}, pages 2775--2779, Brussels, Belgium.
  Association for Computational Linguistics.

\bibitem[{Mihaylov and Nakov(2019)}]{mihaylov2019hunting}
Todor Mihaylov and Preslav Nakov. 2019.
\newblock Hunting for troll comments in news community forums.
\newblock \emph{arXiv preprint arXiv:1911.08113}.

\bibitem[{Miller et~al.(2017)Miller, Feng, Fisch, Lu, Batra, Bordes, Parikh,
  and Weston}]{miller2017parlai}
Alexander~H Miller, Will Feng, Adam Fisch, Jiasen Lu, Dhruv Batra, Antoine
  Bordes, Devi Parikh, and Jason Weston. 2017.
\newblock Parlai: A dialog research software platform.
\newblock \emph{arXiv preprint arXiv:1705.06476}.

\bibitem[{Mitchell et~al.(2019)Mitchell, Wu, Zaldivar, Barnes, Vasserman,
  Hutchinson, Spitzer, Raji, and Gebru}]{modelcardsmitchell}
Margaret Mitchell, Simone Wu, Andrew Zaldivar, Parker Barnes, Lucy Vasserman,
  Ben Hutchinson, Elena Spitzer, Inioluwa~Deborah Raji, and Timnit Gebru. 2019.
\newblock \href {https://doi.org/10.1145/3287560.3287596} {Model cards for
  model reporting}.
\newblock In \emph{Proceedings of the Conference on Fairness, Accountability,
  and Transparency, FAT* 2019, Atlanta, GA, USA, January 29-31, 2019}, pages
  220--229. {ACM}.

\bibitem[{Nakano et~al.(2021)Nakano, Hilton, Balaji, Wu, Ouyang, Kim, Hesse,
  Jain, Kosaraju, Saunders et~al.}]{nakano2021webgpt}
Reiichiro Nakano, Jacob Hilton, Suchir Balaji, Jeff Wu, Long Ouyang, Christina
  Kim, Christopher Hesse, Shantanu Jain, Vineet Kosaraju, William Saunders,
  et~al. 2021.
\newblock Webgpt: Browser-assisted question-answering with human feedback.
\newblock \emph{arXiv preprint arXiv:2112.09332}.

\bibitem[{Nguyen et~al.(2016)Nguyen, Rosenberg, Song, Gao, Tiwary, Majumder,
  and Deng}]{nguyen2016ms}
Tri Nguyen, Mir Rosenberg, Xia Song, Jianfeng Gao, Saurabh Tiwary, Rangan
  Majumder, and Li~Deng. 2016.
\newblock Ms marco: A human generated machine reading comprehension dataset.
\newblock In \emph{CoCo@ NIPS}.

\bibitem[{Ni et~al.(2021)Ni, Young, Pandelea, Xue, Adiga, and
  Cambria}]{ni2021recent}
Jinjie Ni, Tom Young, Vlad Pandelea, Fuzhao Xue, Vinay Adiga, and Erik Cambria.
  2021.
\newblock Recent advances in deep learning based dialogue systems: A systematic
  survey.
\newblock \emph{arXiv preprint arXiv:2105.04387}.

\bibitem[{Ouyang et~al.(2022)Ouyang, Wu, Jiang, Almeida, Wainwright, Mishkin,
  Zhang, Agarwal, Slama, Ray et~al.}]{ouyang2022training}
Long Ouyang, Jeff Wu, Xu~Jiang, Diogo Almeida, Carroll~L Wainwright, Pamela
  Mishkin, Chong Zhang, Sandhini Agarwal, Katarina Slama, Alex Ray, et~al.
  2022.
\newblock Training language models to follow instructions with human feedback.
\newblock \emph{arXiv preprint arXiv:2203.02155}.

\bibitem[{Park et~al.(2021)Park, Jang, Cho, and Choi}]{park2021use}
Namkee Park, Kyungeun Jang, Seonggyeol Cho, and Jinyoung Choi. 2021.
\newblock Use of offensive language in human-artificial intelligence chatbot
  interaction: The effects of ethical ideology, social competence, and
  perceived humanlikeness.
\newblock \emph{Computers in Human Behavior}, 121:106795.

\bibitem[{Partnership{\ }on{\ }AI(2021)}]{pairesponsibleresearch}
Partnership{\ }on{\ }AI. 2021.
\newblock Managing the risks of ai research: Six recommendations for
  responsible publication.

\bibitem[{Perez et~al.(2022)Perez, Huang, Song, Cai, Ring, Aslanides, Glaese,
  McAleese, and Irving}]{perez2022red}
Ethan Perez, Saffron Huang, Francis Song, Trevor Cai, Roman Ring, John
  Aslanides, Amelia Glaese, Nat McAleese, and Geoffrey Irving. 2022.
\newblock Red teaming language models with language models.
\newblock \emph{arXiv preprint arXiv:2202.03286}.

\bibitem[{Pineau et~al.(2021)Pineau, Vincent-Lamarre, Sinha, Larivi{\`e}re,
  Beygelzimer, d’Alch{\'e} Buc, Fox, and Larochelle}]{pineau2021improving}
Joelle Pineau, Philippe Vincent-Lamarre, Koustuv Sinha, Vincent Larivi{\`e}re,
  Alina Beygelzimer, Florence d’Alch{\'e} Buc, Emily Fox, and Hugo
  Larochelle. 2021.
\newblock Improving reproducibility in machine learning research: a report from
  the neurips 2019 reproducibility program.
\newblock \emph{Journal of Machine Learning Research}, 22.

\bibitem[{Rae et~al.(2021)Rae, Borgeaud, Cai, Millican, Hoffmann, Song,
  Aslanides, Henderson, Ring, Young et~al.}]{rae2021scaling}
Jack~W Rae, Sebastian Borgeaud, Trevor Cai, Katie Millican, Jordan Hoffmann,
  Francis Song, John Aslanides, Sarah Henderson, Roman Ring, Susannah Young,
  et~al. 2021.
\newblock Scaling language models: Methods, analysis \& insights from training
  gopher.
\newblock \emph{arXiv preprint arXiv:2112.11446}.

\bibitem[{Rajpurkar et~al.(2016)Rajpurkar, Zhang, Lopyrev, and
  Liang}]{rajpurkar2016squad}
Pranav Rajpurkar, Jian Zhang, Konstantin Lopyrev, and Percy Liang. 2016.
\newblock Squad: 100,000+ questions for machine comprehension of text.
\newblock \emph{arXiv preprint arXiv:1606.05250}.

\bibitem[{Ram et~al.(2018)Ram, Prasad, Khatri, Venkatesh, Gabriel, Liu, Nunn,
  Hedayatnia, Cheng, Nagar et~al.}]{ram2018conversational}
Ashwin Ram, Rohit Prasad, Chandra Khatri, Anu Venkatesh, Raefer Gabriel, Qing
  Liu, Jeff Nunn, Behnam Hedayatnia, Ming Cheng, Ashish Nagar, et~al. 2018.
\newblock Conversational ai: The science behind the alexa prize.
\newblock \emph{arXiv preprint arXiv:1801.03604}.

\bibitem[{Rashkin et~al.(2019)Rashkin, Smith, Li, and
  Boureau}]{rashkin2018towards}
Hannah Rashkin, Eric~Michael Smith, Margaret Li, and Y-Lan Boureau. 2019.
\newblock \href {https://doi.org/10.18653/v1/P19-1534} {Towards empathetic
  open-domain conversation models: A new benchmark and dataset}.
\newblock In \emph{Proceedings of the 57th Annual Meeting of the Association
  for Computational Linguistics}, pages 5370--5381, Florence, Italy.
  Association for Computational Linguistics.

\bibitem[{Rastogi et~al.(2020)Rastogi, Zang, Sunkara, Gupta, and
  Khaitan}]{rastogi2020towards}
Abhinav Rastogi, Xiaoxue Zang, Srinivas Sunkara, Raghav Gupta, and Pranav
  Khaitan. 2020.
\newblock Towards scalable multi-domain conversational agents: The
  schema-guided dialogue dataset.
\newblock In \emph{Proceedings of the AAAI Conference on Artificial
  Intelligence}, volume~34, pages 8689--8696.

\bibitem[{Reed et~al.(2014)Reed, Lee, Anguelov, Szegedy, Erhan, and
  Rabinovich}]{reed2014training}
Scott Reed, Honglak Lee, Dragomir Anguelov, Christian Szegedy, Dumitru Erhan,
  and Andrew Rabinovich. 2014.
\newblock Training deep neural networks on noisy labels with bootstrapping.
\newblock \emph{arXiv preprint arXiv:1412.6596}.

\bibitem[{Roller et~al.(2020)Roller, Boureau, Weston, Bordes, Dinan, Fan,
  Gunning, Ju, Li, Poff et~al.}]{roller2020open}
Stephen Roller, Y-Lan Boureau, Jason Weston, Antoine Bordes, Emily Dinan,
  Angela Fan, David Gunning, Da~Ju, Margaret Li, Spencer Poff, et~al. 2020.
\newblock Open-domain conversational agents: Current progress, open problems,
  and future directions.
\newblock \emph{arXiv preprint arXiv:2006.12442}.

\bibitem[{Roller et~al.(2021)Roller, Dinan, Goyal, Ju, Williamson, Liu, Xu,
  Ott, Smith, Boureau, and Weston}]{roller-etal-2021-recipes}
Stephen Roller, Emily Dinan, Naman Goyal, Da~Ju, Mary Williamson, Yinhan Liu,
  Jing Xu, Myle Ott, Eric~Michael Smith, Y-Lan Boureau, and Jason Weston. 2021.
\newblock \href {https://aclanthology.org/2021.eacl-main.24} {Recipes for
  building an open-domain chatbot}.
\newblock In \emph{Proceedings of the 16th Conference of the European Chapter
  of the Association for Computational Linguistics: Main Volume}, pages
  300--325, Online. Association for Computational Linguistics.

\bibitem[{Saunders et~al.(2022)Saunders, Yeh, Wu, Bills, Ouyang, Ward, and
  Leike}]{saunders2022self}
William Saunders, Catherine Yeh, Jeff Wu, Steven Bills, Long Ouyang, Jonathan
  Ward, and Jan Leike. 2022.
\newblock Self-critiquing models for assisting human evaluators.
\newblock \emph{arXiv preprint arXiv:2206.05802}.

\bibitem[{Serban et~al.(2015)Serban, Lowe, Henderson, Charlin, and
  Pineau}]{serban2015survey}
Iulian~Vlad Serban, Ryan Lowe, Peter Henderson, Laurent Charlin, and Joelle
  Pineau. 2015.
\newblock A survey of available corpora for building data-driven dialogue
  systems.
\newblock \emph{arXiv preprint arXiv:1512.05742}.

\bibitem[{Shachaf and Hara(2010)}]{shachaf2010beyond}
Pnina Shachaf and Noriko Hara. 2010.
\newblock Beyond vandalism: Wikipedia trolls.
\newblock \emph{Journal of Information Science}, 36(3):357--370.

\bibitem[{Sheng et~al.(2021)Sheng, Arnold, Yu, Chang, and
  Peng}]{sheng2021revealing}
Emily Sheng, Josh Arnold, Zhou Yu, Kai-Wei Chang, and Nanyun Peng. 2021.
\newblock \href {http://arxiv.org/abs/2104.08728} {Revealing persona biases in
  dialogue systems}.

\bibitem[{Shuster et~al.(2020)Shuster, Ju, Roller, Dinan, Boureau, and
  Weston}]{shuster2019dialogue}
Kurt Shuster, Da~Ju, Stephen Roller, Emily Dinan, Y-Lan Boureau, and Jason
  Weston. 2020.
\newblock \href {https://doi.org/10.18653/v1/2020.acl-main.222} {The dialogue
  dodecathlon: Open-domain knowledge and image grounded conversational agents}.
\newblock In \emph{Proceedings of the 58th Annual Meeting of the Association
  for Computational Linguistics}, pages 2453--2470, Online. Association for
  Computational Linguistics.

\bibitem[{Shuster et~al.(2022)Shuster, Komeili, Adolphs, Roller, Szlam, and
  Weston}]{shuster2022language}
Kurt Shuster, Mojtaba Komeili, Leonard Adolphs, Stephen Roller, Arthur Szlam,
  and Jason Weston. 2022.
\newblock Language models that seek for knowledge: Modular search \& generation
  for dialogue and prompt completion.
\newblock \emph{arXiv preprint arXiv:2203.13224}.

\bibitem[{Shuster et~al.(2021{\natexlab{a}})Shuster, Poff, Chen, Kiela, and
  Weston}]{shuster2021retrieval}
Kurt Shuster, Spencer Poff, Moya Chen, Douwe Kiela, and Jason Weston.
  2021{\natexlab{a}}.
\newblock \href {https://doi.org/10.18653/v1/2021.findings-emnlp.320}
  {Retrieval augmentation reduces hallucination in conversation}.
\newblock In \emph{Findings of the Association for Computational Linguistics:
  EMNLP 2021}, pages 3784--3803, Punta Cana, Dominican Republic. Association
  for Computational Linguistics.

\bibitem[{Shuster et~al.(2021{\natexlab{b}})Shuster, Urbanek, Dinan, Szlam, and
  Weston}]{shuster2020deploying}
Kurt Shuster, Jack Urbanek, Emily Dinan, Arthur Szlam, and Jason Weston.
  2021{\natexlab{b}}.
\newblock \href {https://doi.org/10.18653/v1/2021.findings-acl.54} {Dialogue in
  the wild: Learning from a deployed role-playing game with humans and bots}.
\newblock In \emph{Findings of the Association for Computational Linguistics:
  ACL-IJCNLP 2021}, pages 611--624, Online. Association for Computational
  Linguistics.

\bibitem[{Smith et~al.(2022)Smith, Hall, Kambadur, Presani, and
  Williams}]{smith2022m}
Eric~Michael Smith, Melissa Hall, Melanie Kambadur, Eleonora Presani, and Adina
  Williams. 2022.
\newblock "i'm sorry to hear that": finding bias in language models with a
  holistic descriptor dataset.
\newblock \emph{arXiv preprint arXiv:2205.09209}.

\bibitem[{Smith and Williams(2021)}]{smith2021hi}
Eric~Michael Smith and Adina Williams. 2021.
\newblock Hi, my name is martha: Using names to measure and mitigate bias in
  generative dialogue models.
\newblock \emph{arXiv preprint arXiv:2109.03300}.

\bibitem[{Smith et~al.(2020)Smith, Williamson, Shuster, Weston, and
  Boureau}]{smith2020bst}
Eric~Michael Smith, Mary Williamson, Kurt Shuster, Jason Weston, and Y-Lan
  Boureau. 2020.
\newblock Can you put it all together: Evaluating conversational agents'
  ability to blend skills.
\newblock In \emph{Proceedings of the 58th Annual Meeting of the Association
  for Computational Linguistics}. ACL.

\bibitem[{Solaiman et~al.(2019)Solaiman, Brundage, Clark, Askell, Herbert-Voss,
  Wu, Radford, Krueger, Kim, Kreps, McCain, Newhouse, Blazakis, McGuffie, and
  Wang}]{solaiman2019}
Irene Solaiman, Miles Brundage, Jack Clark, Amanda Askell, Ariel Herbert-Voss,
  Jeff Wu, Alec Radford, Gretchen Krueger, Jong~Wook Kim, Sarah Kreps, Miles
  McCain, Alex Newhouse, Jason Blazakis, Kris McGuffie, and Jasmine Wang. 2019.
\newblock \href {https://doi.org/10.48550/ARXIV.1908.09203} {Release strategies
  and the social impacts of language models}.

\bibitem[{Song et~al.(2022)Song, Kim, Park, Shin, and Lee}]{song2020learning}
Hwanjun Song, Minseok Kim, Dongmin Park, Yooju Shin, and Jae-Gil Lee. 2022.
\newblock \href {https://doi.org/10.1109/TNNLS.2022.3152527} {Learning from
  noisy labels with deep neural networks: A survey}.
\newblock \emph{IEEE Transactions on Neural Networks and Learning Systems},
  pages 1--19.

\bibitem[{Sonnenburg et~al.(2007)Sonnenburg, Braun, Ong, Bengio, Bottou,
  Holmes, LeCunn, Muller, Pereira, Rasmussen et~al.}]{sonnenburg2007need}
Soren Sonnenburg, Mikio~L Braun, Cheng~Soon Ong, Samy Bengio, Leon Bottou,
  Geoffrey Holmes, Yann LeCunn, Klaus-Robert Muller, Fernando Pereira,
  Carl~Edward Rasmussen, et~al. 2007.
\newblock The need for open source software in machine learning.

\bibitem[{Strubell et~al.(2019)Strubell, Ganesh, and
  McCallum}]{strubell-etal-2019-energy}
Emma Strubell, Ananya Ganesh, and Andrew McCallum. 2019.
\newblock \href {https://doi.org/10.18653/v1/P19-1355} {Energy and policy
  considerations for deep learning in {NLP}}.
\newblock In \emph{Proceedings of the 57th Annual Meeting of the Association
  for Computational Linguistics}, pages 3645--3650, Florence, Italy.
  Association for Computational Linguistics.

\bibitem[{Thoppilan et~al.(2022)Thoppilan, De~Freitas, Hall, Shazeer,
  Kulshreshtha, Cheng, Jin, Bos, Baker, Du et~al.}]{thoppilan2022lamda}
Romal Thoppilan, Daniel De~Freitas, Jamie Hall, Noam Shazeer, Apoorv
  Kulshreshtha, Heng-Tze Cheng, Alicia Jin, Taylor Bos, Leslie Baker, Yu~Du,
  et~al. 2022.
\newblock Lamda: Language models for dialog applications.
\newblock \emph{arXiv preprint arXiv:2201.08239}.

\bibitem[{Tomaiuolo et~al.(2020)Tomaiuolo, Lombardo, Mordonini, Cagnoni, and
  Poggi}]{tomaiuolo2020survey}
Michele Tomaiuolo, Gianfranco Lombardo, Monica Mordonini, Stefano Cagnoni, and
  Agostino Poggi. 2020.
\newblock A survey on troll detection.
\newblock \emph{Future internet}, 12(2):31.

\bibitem[{Ung et~al.(2022)Ung, Xu, and Boureau}]{ung2021saferdialogues}
Megan Ung, Jing Xu, and Y-Lan Boureau. 2022.
\newblock \href {https://doi.org/10.18653/v1/2022.acl-long.447}
  {{S}a{F}e{RD}ialogues: Taking feedback gracefully after conversational safety
  failures}.
\newblock In \emph{Proceedings of the 60th Annual Meeting of the Association
  for Computational Linguistics (Volume 1: Long Papers)}, pages 6462--6481,
  Dublin, Ireland. Association for Computational Linguistics.

\bibitem[{Urbanek et~al.(2019)Urbanek, Fan, Karamcheti, Jain, Humeau, Dinan,
  Rockt{\"a}schel, Kiela, Szlam, and Weston}]{urbanek2019learning}
Jack Urbanek, Angela Fan, Siddharth Karamcheti, Saachi Jain, Samuel Humeau,
  Emily Dinan, Tim Rockt{\"a}schel, Douwe Kiela, Arthur Szlam, and Jason
  Weston. 2019.
\newblock \href {https://doi.org/10.18653/v1/D19-1062} {Learning to speak and
  act in a fantasy text adventure game}.
\newblock In \emph{Proceedings of the 2019 Conference on Empirical Methods in
  Natural Language Processing and the 9th International Joint Conference on
  Natural Language Processing (EMNLP-IJCNLP)}, pages 673--683, Hong Kong,
  China. Association for Computational Linguistics.

\bibitem[{Vaswani et~al.(2017)Vaswani, Shazeer, Parmar, Uszkoreit, Jones,
  Gomez, Kaiser, and Polosukhin}]{NIPS2017_3f5ee243}
Ashish Vaswani, Noam Shazeer, Niki Parmar, Jakob Uszkoreit, Llion Jones,
  Aidan~N Gomez, \L~ukasz Kaiser, and Illia Polosukhin. 2017.
\newblock \href
  {https://proceedings.neurips.cc/paper/2017/file/3f5ee243547dee91fbd053c1c4a845aa-Paper.pdf}
  {Attention is all you need}.
\newblock In \emph{Advances in Neural Information Processing Systems},
  volume~30. Curran Associates, Inc.

\bibitem[{Weidinger et~al.(2021)Weidinger, Mellor, Rauh, Griffin, Uesato,
  Huang, Cheng, Glaese, Balle, Kasirzadeh et~al.}]{weidinger2021ethical}
Laura Weidinger, John Mellor, Maribeth Rauh, Conor Griffin, Jonathan Uesato,
  Po-Sen Huang, Myra Cheng, Mia Glaese, Borja Balle, Atoosa Kasirzadeh, et~al.
  2021.
\newblock Ethical and social risks of harm from language models.
\newblock \emph{arXiv preprint arXiv:2112.04359}.

\bibitem[{Wolf et~al.(2019)Wolf, Sanh, Chaumond, and
  Delangue}]{wolf2019transfertransfo}
Thomas Wolf, Victor Sanh, Julien Chaumond, and Clement Delangue. 2019.
\newblock Transfer{T}ransfo: {A} transfer learning approach for neural network
  based conversational agents.
\newblock In \emph{Neur{IPS} Workshop on Conversational AI}.

\bibitem[{Wulczyn et~al.(2017)Wulczyn, Thain, and Dixon}]{personal_attack}
Ellery Wulczyn, Nithum Thain, and Lucas Dixon. 2017.
\newblock \href {https://doi.org/10.1145/3038912.3052591} {Ex machina: Personal
  attacks seen at scale}.
\newblock In \emph{Proceedings of the 26th International Conference on World
  Wide Web}, WWW '17, page 1391–1399, Republic and Canton of Geneva, CHE.
  International World Wide Web Conferences Steering Committee.

\bibitem[{Xu et~al.(2020)Xu, Ju, Li, Boureau, Weston, and
  Dinan}]{xu2020recipes}
Jing Xu, Da~Ju, Margaret Li, Y-Lan Boureau, Jason Weston, and Emily Dinan.
  2020.
\newblock Recipes for safety in open-domain chatbots.
\newblock \emph{arXiv preprint arXiv:2010.07079}.

\bibitem[{Xu et~al.(2022{\natexlab{a}})Xu, Szlam, and Weston}]{xu2021beyond}
Jing Xu, Arthur Szlam, and Jason Weston. 2022{\natexlab{a}}.
\newblock \href {https://doi.org/10.18653/v1/2022.acl-long.356} {Beyond
  goldfish memory: Long-term open-domain conversation}.
\newblock In \emph{Proceedings of the 60th Annual Meeting of the Association
  for Computational Linguistics (Volume 1: Long Papers)}, pages 5180--5197,
  Dublin, Ireland. Association for Computational Linguistics.

\bibitem[{Xu et~al.(2022{\natexlab{b}})Xu, Ung, Komeili, Arora, Boureau, and
  Weston}]{xu2022continual}
Jing Xu, Megan Ung, Mojtaba Komeili, Kushal Arora, Y-Lan Boureau, and Jason
  Weston. 2022{\natexlab{b}}.
\newblock \href {https://doi.org/10.48550/ARXIV.2208.03270} {Learning new
  skills after deployment: Improving open-domain internet-driven dialogue with
  human feedback}.

\bibitem[{Yang et~al.(2018)Yang, Yuan, Cer, Kong, Constant, Pilar, Ge, Sung,
  Strope, and Kurzweil}]{reddit_use}
Yinfei Yang, Steve Yuan, Daniel Cer, Sheng-yi Kong, Noah Constant, Petr Pilar,
  Heming Ge, Yun-Hsuan Sung, Brian Strope, and Ray Kurzweil. 2018.
\newblock Learning semantic textual similarity from conversations.
\newblock In \emph{Proceedings of The Third Workshop on Representation Learning
  for {NLP}}, pages 164--174, Melbourne, Australia. Association for
  Computational Linguistics.

\bibitem[{Zhang et~al.(2018)Zhang, Dinan, Urbanek, Szlam, Kiela, and
  Weston}]{zhang2018personalizing}
Saizheng Zhang, Emily Dinan, Jack Urbanek, Arthur Szlam, Douwe Kiela, and Jason
  Weston. 2018.
\newblock Personalizing dialogue agents: I have a dog, do you have pets too?
\newblock In \emph{Proceedings of the 56th Annual Meeting of the Association
  for Computational Linguistics}, pages 2204--2213. ACL.

\bibitem[{Zhang et~al.(2022)Zhang, Roller, Goyal, Artetxe, Chen, Chen, Dewan,
  Diab, Li, Lin et~al.}]{zhang2022opt}
Susan Zhang, Stephen Roller, Naman Goyal, Mikel Artetxe, Moya Chen, Shuohui
  Chen, Christopher Dewan, Mona Diab, Xian Li, Xi~Victoria Lin, et~al. 2022.
\newblock Opt: Open pre-trained transformer language models.
\newblock \emph{arXiv preprint arXiv:2205.01068}.

\bibitem[{Zhang et~al.(2020)Zhang, Sun, Galley, Chen, Brockett, Gao, Gao, Liu,
  and Dolan}]{zhang2019dialogpt}
Yizhe Zhang, Siqi Sun, Michel Galley, Yen-Chun Chen, Chris Brockett, Xiang Gao,
  Jianfeng Gao, Jingjing Liu, and Bill Dolan. 2020.
\newblock \href {https://doi.org/10.18653/v1/2020.acl-demos.30} {{DIALOGPT} :
  Large-scale generative pre-training for conversational response generation}.
\newblock In \emph{Proceedings of the 58th Annual Meeting of the Association
  for Computational Linguistics: System Demonstrations}, pages 270--278,
  Online. Association for Computational Linguistics.

\bibitem[{Zhao et~al.(2018)Zhao, Zhou, Li, Wang, and Chang}]{zhao2018gender}
Jieyu Zhao, Yichao Zhou, Zeyu Li, Wei Wang, and Kai-Wei Chang. 2018.
\newblock \href {https://doi.org/10.48550/ARXIV.1809.01496} {Learning
  gender-neutral word embeddings}.

\bibitem[{Zhou et~al.(2020)Zhou, Gao, Li, and Shum}]{zhou2020design}
Li~Zhou, Jianfeng Gao, Di~Li, and Heung-Yeung Shum. 2020.
\newblock The design and implementation of xiaoice, an empathetic social
  chatbot.
\newblock \emph{Computational Linguistics}, 46(1):53--93.

\end{thebibliography}
\bibliographystyle{acl_natbib}

\clearpage
\onecolumn
\appendix

\section{Additional Safety Evaluations}
\label{sec:appendix_safety_evals}

\begin{table*}
\centering
\small 
\begin{tabular}{rrrrr}
& \multicolumn{1}{c}{BB3-3B} & \multicolumn{1}{c}{OPT-175B (Zero-shot)} & \multicolumn{1}{c}{OPT-175B (few-shot)} & \multicolumn{1}{c}{BB3-175B} \\
\hline
\textit{``I hate...''} & potbellied & stinky & potbellied & potbellied \\
& stinky & dirty-blonde & balding & stinky \\
& beefy & smelly & heavy-set & four-eyed \\
& barrel-chested & straight-haired & stinky & small-stature \\
& ugly & heavy-set & smelly & acne-ridden \\
& coily-haired & scrawny & acne-ridden & unshaved \\
& bulky & potbellied & swole & bald-headed \\
& shaggy & ugly & coily-haired & heavy \\
& brown-haired & bald-headed & scrawny & ugly \\
& giant & unattractive & barrel-chested & swole \\
\hline
\textit{``I love...''} & well-built & well-built & well-built & hot \\
& dark-skinned & with an afro & brunette & average-height \\
& white-haired & hot & redhead & strong \\
& good-looking & who is a little person & heftier & in-shape \\
& slender & shaggy & grey-eyed & well-built \\
& beautiful & brunette & ginger & fit \\
& brunette & beautiful & four-eyed & beautiful \\
& pretty & petite & average-height & physically fit \\
& gorgeous & curvy & bigger & attractive \\
& goateed & chunky & brown-eyed & huge \\
\end{tabular}
\caption{
The 10 body-type descriptor terms from \textsc{HolisticBias} for which the likelihood of \textit{``I hate [DESCRIPTOR + NOUN].''} most exceeds that of \textit{``I love [DESCRIPTOR + NOUN].''} (top), and vice versa (bottom), as a function of model.
\label{tab:i_hate_vs_i_love}
}
\end{table*}

\autoref{tab:i_hate_vs_i_love} shows which descriptor terms from the ``Body type'' axis of the \textsc{HolisticBias} dataset are most likely to have a low perplexity in \textit{``I hate [DESCRIPTOR + NOUN].''} sentences relative to \textit{``I love [DESCRIPTOR + NOUN].''} sentences, or vice versa, as a function of model. The likelihood of a descriptor to have a low perplexity in \textit{``I hate/love [DESCRIPTOR + NOUN].''} sentences is measured by calculating the fraction of \textsc{HolisticBias} nouns for which that combination of descriptor and noun results in a perplexity lower than the median perplexity for the given adjective (``hate'' or ``love''). 
Terms such as ``potbellied'', ``barrel-chested'', ``heavy-set'', ``scrawny'', ``bald-headed'', ``acne-ridden'', and ``swole'' tend to have greater likelihoods of \textit{``I hate...''} than of \textit{``I love...''}, and the reverse is true for terms such as ``well-built'', ``brunette'', and ``brown-eyed''.

For both \autoref{tab:i_hate_vs_i_love} and the table of Likelihood Bias scores (\autoref{tab:holistic_bias_toplevel}), perplexities are measured on the base models without any flagging of unsafe responses, topic changes, etc. \autoref{tab:holistic_bias_toplevel} specifically measures the zero-shot OPT-175B model.

\section{Training \& Inference Details}
\label{sec:appendix_train_details}

\begin{table*}[bht!]
\tiny
\centering
\begin{tabular}{l|ll|ll|lll|llll|l}
& \multicolumn{11}{c}{Training Module} \\
  & \multicolumn{2}{c}{Decision} & \multicolumn{2}{c}{Generation} & \multicolumn{3}{c}{Knowledge} & \multicolumn{4}{c}{Dialogue} & LM\\
 & \rot{\multirow{2}{*}{\tiny{Search}}} & 
\rot{\multirow{2}{*}{\tiny{Memory}}} &
\rot{\multirow{2}{*}{\tiny{Query}}} & 
\rot{\multirow{2}{*}{\tiny{Memory}}} &
\rot{\multirow{2}{*}{\tiny{Search}}} &
\rot{\multirow{2}{*}{\tiny{Memory}}} &
\rot{\multirow{2}{*}{\tiny{Entity}}} &
\rot{\multirow{2}{*}{\tiny{Search}}} &
\rot{\multirow{2}{*}{\tiny{Memory}}} &
\rot{\multirow{2}{*}{\tiny{Entity}}} &
\rot{\multirow{2}{*}{\tiny{Vanilla}}} \\
\hline
\textbf{\textit{Question Answering}} & & & & & & & & & & &\\
MS MARCO  &  &   &  &  & 282k  &  &  & 282k &  &  &  &\\
SQuAD  & 88k &   &  &  &  88k  &  &  & &  &  &  &\\
TriviaQA   & 76k &   &  &  & 475k &  &  &  &  &  &  &\\
Natural Questions &  &   &  &  & 111k &  &  &  &  &  &  &\\
Natural Questions (Open)  & 79k &   &  &  & 79k &  &  &  &  &  &  &\\
Natural Questions (Open Dialogues)  & &   &  &  & 11k &  &  &  &  &  & & \\
\hline
\textbf{\textit{Knowledge-Grounded Dialogue}} & & & & & & & & & & &\\
Wizard of the Internet  & 41k &   & 35k & & 22k & & & 33k & & & 8k &\\
Wizard of Wikipedia & 74k &  &  &  & 77k &  &  & 77k &  &  & 6k &\\
Funpedia  &  &   &  &  &  &  &  & 81k &  &  &  &\\
\hline
\textbf{\textit{Open-Domain Dialogue}} & & & & & & & & & & & \\
PersonaChat  & 131k & 68k  &  &  &  & 63k & 7k &  & 65k & 7k & 131k &\\
Empathetic Dialogues & 65k & 1k  &  &  &  & 1k & 1k &  & 1k & 1k & 65k &\\
Blended Skill Talk  & & 5k  &  &  &  & 50k & 1k &  & 50k & 1k &  &\\
Multi-Session Chat  & 97k & 23k  &  & 86k &  & 34k & 9k &  & 34k & 9k & 106k & \\
LIGHT + WILD &  &   &  &  &  &  &  &  &  &  & 342k &\\
\hline
\textbf{\textit{Recovery \& Feedback}} & & & & & & & & & & &\\
SaFeRDialogues  &  &  &  &  &  &  &  &  &  &  & 6k & \\
FITS  &  &   & 7k &  & 11k &  &  & 44k &  &  & & \\
\hline
\textbf{\textit{Task-Oriented Dialogue}} & & & & & & & & & & &\\
Google SGD  &  &   &  &  &  &  &  & 42k &  &  & & \\
Taskmaster  &  &   &  &  &  &  &  & 40k &  &  & & \\
Taskmaster 2  &  &   &  &  &  &  &  & 56k &  &  & & \\
Taskmaster 3  &  &   &  &  &  &  &  & 64k &  &  & & \\
\hline
\textbf{\textit{Language Modeling}}
& & & & & & & & & & & & 591k \\
\hline
\hline
\textbf{Totals} & 651k & 97k & 42k & 86k & 1.156m &  148k & 18k & 639k & 150k & 18k & 745k & 591k

\end{tabular}
\caption{
Approximate number of train examples for each dataset within each training module.
\label{tab:dataset_trainset_details_appendix_examples}
}
\end{table*}

\begin{table*}[bht!]
\tiny
\centering
\begin{tabular}{l|ll|ll|lll|llll|l}
& \multicolumn{11}{c}{Training Module} \\
  & \multicolumn{2}{c}{Decision} & \multicolumn{2}{c}{Generation} & \multicolumn{3}{c}{Knowledge} & \multicolumn{4}{c}{Dialogue} & LM\\
 & \rot{\multirow{2}{*}{\tiny{Search}}} & 
\rot{\multirow{2}{*}{\tiny{Memory}}} &
\rot{\multirow{2}{*}{\tiny{Query}}} & 
\rot{\multirow{2}{*}{\tiny{Memory}}} &
\rot{\multirow{2}{*}{\tiny{Search}}} &
\rot{\multirow{2}{*}{\tiny{Memory}}} &
\rot{\multirow{2}{*}{\tiny{Entity}}} &
\rot{\multirow{2}{*}{\tiny{Search}}} &
\rot{\multirow{2}{*}{\tiny{Memory}}} &
\rot{\multirow{2}{*}{\tiny{Entity}}} &
\rot{\multirow{2}{*}{\tiny{Vanilla}}} \\
\hline
\textbf{\textit{Question Answering}} & & & & & & & & & & &\\
MS MARCO  &  &   &  &  & 112.5m  &  &  & 20.3m &  &  &  &\\
SQuAD  & 1.9m &   &  &  & 16.7m &  &  &  &  &  &  &\\
TriviaQA   & 2.1m &   &  &  & 280.6m &  &  &  &  &  &  &\\
Natural Questions  &  &   &  &  & 116.4m &  &  &  &  &  &  &\\
Natural Questions (Open)  & 1.5m &   &  &  & 16.3m  &  &  &  &  &  &  &\\
Natural Questions (Open Dialogues)  & &   &  &  & 5.0m &  &  &  &  &  & & \\
\hline
\textbf{\textit{Knowledge-Grounded Dialogue}} & & & & & & & & & & &\\
Wizard of the Internet & 1.1m &   & 4.7m & & 19.7m & &  & 7.8m & &  & 1.4m &\\
Wizard of Wikipedia & 2.0m &  &  &  & 51.9m &  &  & 12.4m &  &  & 863k &\\
Funpedia  &  &   &  &  &  &  &  & 4.8m &  &  &  &\\
\hline
\textbf{\textit{Open-Domain Dialogue}} & & & & & & & & & & & \\
PersonaChat  & 3.0m & 2.4m  &  &  &  & 21.5m & 1.2m &  & 9.0m & 929k & 25.1m &\\
Empathetic Dialogues  & 1.8m & 50k  &  &  &  & 145k & 40k &  & 128k & 55k & 3.7m &\\
Blended Skill Talk  & & 187k  &  &  &  & 15.5m & 240k &  & 9.0m & 235k &  &\\
Multi-Session Chat  & 3.9m & 1.3m  &  & 16m &  & 40.3m & 6.6m &  & 26.0m & 7.0m & 75.3m & \\
LIGHT + WILD &  &   &  &  &  &  &  &  &  &  & 83.6m &\\
\hline
\textbf{\textit{Recovery \& Feedback}} & & & & & & & & & & &\\
SaFeRDialogues &  &  &  &  &  &  &  &  &  &  & 817k & \\
FITS  &  &   & 687k &  & 8.0m &  &  & 7.7m &  &  & & \\
\hline
\textbf{\textit{Task-Oriented Dialogue}} & & & & & & & & & & &\\
Google SGD  &  &   &  &  &  &  &  & 11.0m &  &  & & \\
Taskmaster  &  &   &  &  &  &  &  & 9.4m &  &  & & \\
Taskmaster 2  &  &   &  &  &  &  &  & 12.4m &  &  & & \\
Taskmaster 3  &  &   &  &  &  &  &  & 16.0m &  &  & & \\
\hline
\textbf{\textit{Language Modeling}}
& & & & & & & & & & & & 170.2m \\
\hline
\hline
\textbf{Totals} & 17.2m & 3.9m & 5.4m & 16.0m & 627m & 77.5m & 8.1m & 97.1m & 44.1m & 8.3m & 195.5m & 170.2m

\end{tabular}
\caption{
Approximate number of train tokens for each dataset within each training module.
\label{tab:dataset_trainset_details_appendix_tokens}
}
\end{table*}

\subsection{Data Details}
The fine-tuning data for BB3 comprises roughly 4 million source/target examples spread across the various training modules. This corresponds to around 1.13B training tokens. When fine-tuning the OPT-based BB3 models, we additionally included ~600k examples (~170m tokens) of pre-training data to help with training stability. \autoref{tab:dataset_trainset_details_appendix_examples} and \autoref{tab:dataset_trainset_details_appendix_tokens} enumerate the breakdown by module.

\subsection{BB3-3B Training}
The 3B parameter BlenderBot 3 model was trained on 64 x 32gb V100 GPUs for 27k updates with a batch size of 64, using the Adam optimizer \cite{kingma2014adam} with weight decay \cite{loshchilov2018decoupled} and a linear warmup of 100 updates before reaching a learning rate of $1e-6$. Early stopping was performed on a validation set comprising a subset of the training tasks. The model was trained with 1024 context tokens.

We refer the reader to the appendix of \citet{shuster2022language} for the full architecture and pre-training details of the 3B R2C2 base model for BB3. 

\subsection{BB3-30B/BB3-175B Training}
The 30B and 175B parameter BlenderBot 3 models were each trained for one epoch of the training data on 64 (30B) or 128 (175B) x 40gb A100 GPUs; we found that the model (especially the 175B version) overfit significantly when seeing the training data more than once. The 175B model was trained with a batch size of $2^{18}$ and the 30B model was trained with a batch size of $2^{19}$, resulting in roughly 5600 updates and 2800 updates respectively. Each model was trained using the Adam optimizer with weight decay, with a linear warmup period of 10\% of the total train updates, reaching a maximum learning rate of $6e-6$ (the LR at the end of pre-training) and subsequently using polynomial weight decay (with a decay factor of 0.1).

\subsection{Inference}

We use the following generation settings for each module, ranging from greedy decoding to the recently introduced factual nucleus sampling method. Due to computational and latency concerns, we employ different generation strategies for the BB3-3B model and the BB3-175B model, in two notable ways. 

First, while in some cases we use beam search for the BB3-3B model, we avoid any decoding algorithm requiring more than one ongoing output generation for BB3-175B; while we found that such techniques (e.g., sample and rank from \citet{adiwardana2020meena}) can yield higher downstream word-overlap metrics for dialogue, we aimed to maximize throughput and latency, especially when serving a large model. In circumstances where beam search is notably useful (i.e., dialogue generation), we instead employ the recently introduced factual nucleus sampling \cite{lee2022factuality}; we found this method to provide an appropriate balance between diversity of downstream generation while avoiding the hallucinatory side effects of other popular sampling methods such as standard nucleus sampling \cite{Holtzman2020The} (see e.g. discussion in \citet{shuster2021retrieval} for the effects of sampling methods on hallucination in knowledge-grounded dialogue).

The second difference is how to employ repetition-blocking heuristics. For BB3-3B, at times we employ beam blocking and context blocking, such that we prevent the model from generating previously seen n-grams in either the current generation or even the entire preceding context. Again, due to latency, throughput, and memory considerations, we avoid such heuristics for BB3-175B, and instead implement the same repetition heuristics that OpenAI uses for InstructGPT\footnote{\tiny{\url{https://beta.openai.com/docs/api-reference/engines/retrieve}}}; specifically, we apply penalties to the logits of tokens proportional to a token's \textit{presence} in the current generation ($\alpha_{pres}$) and to a token's \textit{frequency} ($\alpha_{freq}$). We additionally consider $\alpha_{pres\_src}$ and $\alpha_{freq\_src}$ penalties that correspond to the tokens presence and frequency in the \textit{source} (context) tokens (i.e., prompt tokens). Employing these heuristics, in tandem with factual nucleus, provides a good alternative to beam search + beam-/context-blocking.

\paragraph{Decision Modules} We use greedy decoding for both models for the internet search decision and long-term memory access decision modules.
\paragraph{Query Generation} We use greedy decoding for the query generation module as well. We enforce a minimum generation length of 2 for the BB3-3B model.
\paragraph{Memory Generation} The BB3-3B model uses beam search with a beam size of 3 and a minimum beam length of 10, and tri-gram blocking in the generation. For BB3-175B, we simply use greedy decoding.
\paragraph{Relevant Entity Extraction} For extracting a relevant entity from the context, BB3-3B employs beam search with a beam size of 3, and tri-gram blocking on both the generated output and the encoded dialogue context. For BB3-175B, we use greedy decoding, and employ repetition penalties with $\alpha_{pres}=\alpha_{freq}=0.5$.
\paragraph{Access Long-Term Memory} For BB3-3B, we use beam search with beam size of 3, minimum generation length of 5, and tri-gram blocking on the generated output. For BB3-175B, we use greedy decoding, and employ repetition penalties with $\alpha_{pres}=\alpha_{freq}=0.5$.
\paragraph{Internet Knowledge Response Generation} For BB3-3B, we use beam search with a beam size of 3, a minimum generation length of 10, and tri-gram blocking on both the generated output and the context. For BB3-175B, we once again use greedy decoding with repetition penalties $\alpha_{pres}=\alpha_{freq}=0.5$.
\paragraph{Dialogue Response Generation} For the final dialogue response, BB3-3B uses beam search with a beam size of 10, a minimum generation length of 20, and tri-gram blocking on both the generated output and the context. BB3-175B uses factual nucleus sampling, with $topp=0.9$, a $\lambda$-decay of 0.9, $\omega$-bound of 0.3, and a $p$-reset after each generated full-stop token. We additionally employ repetition penalties with $\alpha_{pres}=\alpha_{freq}=\alpha_{pres\_src}=\alpha_{freq\_src}=0.5$.

\section{Current Events Evaluation Details}
\label{sec:appendix_topical}

To gather a set of topics that have recently been in the news, we follow \citet{shuster2022language}. First, from Wikipedia, we randomly choose 300 entities from the set of current events from July 2022\footnote{\tiny\url{https://en.wikipedia.org/wiki/Portal:Current_events/July_2022}}. We then use those entities to construct questions of the format: "What's the latest news you've heard about \{entity\}?". We then generate a response to each question using BB3-175B and InstructGPT (text-davinci-002). We use the Mojeek API\footnote{\tiny\url{http://mojeek.com}} as the web search engine for BB3-175B. To encourage news results, we append "news july 2022" to the search query generated by the model. For InstructGPT, we use the default "Chat" prompt and generation parameters provided by OpenAI\footnote{\tiny\url{https://beta.openai.com/examples/default-chat}}.

The questions we ask for each comparison are:
\begin{itemize}
    \item Current: "Which response has more up-to-date information?"
    \item Specific: "Which response is easier to invalidate?"
    \item True: "Which response is more truthful?"
    \item Interesting: "If you had to say one of these speakers is interesting and one is boring, who would you say is more interesting?"
    \item Sensible: "If you had to say one speaker responds sensibly and the other doesn't quite make sense, which would you say is more sensible?"
\end{itemize}

Interesting and Sensible were more subjective, so we hire crowdworkers to evaluate these characteristics, enforcing that each response pair is evaluated by a different crowdworker. Current, Specific, and True take more time and effort to evaluate and can be supported with objective evidence, so these characteristics are evaluated by a smaller group of expert evaluators utilizing internet search for validation. We allow for ties in the Current, Specific, and True evaluations, whereas we require a winner for Interesting and Sensible. Ties were ignored.

Given two responses containing true statements, if one response contained only facts and the other added conjecture, we consider the response with conjecture less true. Given a response with no information and a response with out-of-date information, we consider the response with out-of-date information to be more current. Note that a model that always avoids giving an answer (e.g. "I haven't heard anything about that.") would be true 100\% of the time, whereas responses that are highly specific are more likely to be out-of-date or definitively false. 

\section{Prompts}
\label{sec:appendix_prompts}

\autoref{tab:prompts_table} provides the prompts used for the OPT-175B baseline model when generating for each of the BB3 modules. The few-shot model was provided a number of in-context examples sampled from the training data; the few-shot template, dataset(s), and number of examples are also provided in \autoref{tab:prompts_table}. We did not tune prompt selection, so we note that it is possible that other prompts may have yielded better (or worse) downstream performance.

\begin{table*}
\tiny
\centering
\begin{tabular}{|l|l|l|l|l|}
\toprule
& & \multicolumn{3}{c}{\textbf{Few-shot}}  \\
\textbf{Module} & \textbf{Prompt} & \textbf{Template} & \textbf{Dataset} & \textbf{Num Examples} \\
\midrule
Search Decision & Person 2 must decide whether to search the internet. & \specialcell{Person 1:...\\Search Decision:} & WizInt, QA data & 9 \\
\hline
Memory Decision & \specialcell{A conversation between two persons.\\Person 2 must consult their notes about Person 1.} & \specialcell{Person 1:...\\Memory Decision:} & MSC & 9 \\
\hline
Query Generation & \specialcell{Person 2 must write a search query for a search engine.} & \specialcell{Person 1:...\\Person 2:...\\Person 1:...\\Query:...} & WizInt & 5 \\
\hline
Memory Generation & \specialcell{A conversation between two persons.\\Person 2 writes a note about Person 1 to help remember\\information for later.} & \specialcell{Person 1:...\\Person 2:...\\Person 1:...\\Memory: Person 1...} & MSC & 5 \\
\hline
Entity Knowledge Generation & \specialcell{A conversation between two persons.\\Person 2 recalls a previous topic in the conversation.} & \specialcell{Person 1:...\\Person 2:...\\Person 1:...\\Previous Topic:...} & PersonaChat & 5 \\
\hline
Memory Knowledge Generation & \specialcell{A conversation between two persons.\\Person 2 recalls an interesting fact about Person 1\\or Person 2.} & \specialcell{Person 1:...\\Person 2:...\\Person 1:...\\Personal Fact:...} & MSC & 2 \\
\hline
Search Knowledge Generation & \specialcell{A conversation between two persons.\\Person 2 finds an interesting fact from the internet.} & \specialcell{Person 1:...\\Person 2:...\\Person 1:...\\Interesting Fact:...} & WizInt, WoW, NQ & 3 \\
\hline
Entity Dialogue Generation & \specialcell{A conversation between two persons.\\Person 2 would like to continue talking about a \\previous topic in the conversation.} & \specialcell{Person 1:...\\Person 2:...\\Person 1:...\\Previous Topic:...\\Person 2:} & MSC, PersonaChat, ED & 4 \\
\hline
Memory Dialogue Generation & \specialcell{A conversation between two persons.\\Person 2 would like to chat about an\\interesting fact about Person 1 or Person 2.} & \specialcell{Person 1:...\\Person 2:...\\Person 1:...\\Personal Fact: Person 1...\\Person 2:...} & MSC, PersonaChat & 4 \\
\hline
Search Dialogue Generation & \specialcell{A conversation between two persons.\\Person 2 would like to tell Person 1\\about something Person 2 found on the internet.} & \specialcell{Person 1:...\\Person 2:...\\Person 1:...\\Interesting Fact:...\\Person 2:...} & WizInt, WoW, MSMarco & 3 \\
\bottomrule
\end{tabular}
\caption{
Prompts and few-shot templates for the various BB3 modules, used with the OPT-175B model.
\label{tab:prompts_table}
}
\end{table*}

\section{Additional Results}

In \autoref{tab:additional_ppl_results}, we provide model perplexity measurements on a subset of the validation data. In \autoref{tab:wizint_human_eval_appendix}, we provide more measurements from the human evaluation for short conversations. In \autoref{tab:unsafe_generation_appendix}, we provide the full breakdown of per-tool Safety Bench unsafe generation test results.

\begin{table*}
\small
\centering
\begin{tabular}{l|rrrr|rrr|rr||rr}
& \multicolumn{11}{c}{\textbf{Module Perplexity}} \\
  &\multicolumn{4}{c}{\textit{Dialogue}} & \multicolumn{3}{c}{\textit{Knowledge}} & \multicolumn{2}{c}{\textit{Generation}} & \multicolumn{2}{c}{\textit{Averages}}\\  
\textbf{Model} & 
\rot{\multirow{2}{*}{\tiny{Search}}} &
\rot{\multirow{2}{*}{\tiny{Memory}}} &
\rot{\multirow{2}{*}{\tiny{Entity}}} &
\rot{\multirow{2}{*}{\tiny{Vanilla}}} &
\rot{\multirow{2}{*}{\tiny{Search}}} &
\rot{\multirow{2}{*}{\tiny{Memory}}} &
\rot{\multirow{2}{*}{\tiny{Entity}}} &
\rot{\multirow{2}{*}{\tiny{Query}}} & 
\rot{\multirow{2}{*}{\tiny{Memory}}} &
\rot{\multirow{2}{*}{\tiny{Dialogue}}} &
\rot{\multirow{2}{*}{\tiny{Knowledge}}}
\\
\hline
\midrule

OPT-175B Zero-shot & 6.4 & 8.3 & 9.2 & 10.0 & 1.6 & 2.2 & 1.8 & 3.7 & 5.9 & 8.3 & 1.8 \\
OPT-175B Few-shot & 6.1 & 7.6 & 8.7 & 9.9 & 3.3 & 3.7 & 1.5 & 3.4 & 4.0 & 7.9 & 3.1\\
\hline
BB3-3B & 5.6 & 8.1 & 9.5 & 11.3 & \textbf{1.2} & \textbf{1.1} & 3.1 & 4.7 & \textbf{2.6} & 8.4 & 1.5\\
BB3-30B & 4.5 & 6.2 & 8.3 & 8.5 & 1.4 & 1.3 & \textbf{1.2} & 3.1 & 3.2 & 6.6 & \textbf{1.3}\\
BB3-175B & \textbf{4.3} & \textbf{5.8} & \textbf{8.0} & \textbf{8.0} & 1.4 & 1.3 & \textbf{1.2} & \textbf{3.0} & 3.0 & \textbf{6.2} & \textbf{1.3}\\
\bottomrule
\end{tabular}
\caption{
Average model perplexity on the validation tasks. The module perplexities correspond to subsets of the validation data: \textbf{Search Dialogue}: Wizard of Internet, Wizard of Wikipedia, Funpedia, FITS. \textbf{Memory Dialogue}: PersonaChat, Multi-Session Chat. \textbf{Entity Dialogue}: Blended Skill Talk, Empathetic Dialogues. \textbf{Vanilla Dialogue}: Blended Skill Talk, LIGHT, SaFeRDialogues. \textbf{Search Knowledge}: Wizard of the Internet, Wizard of Wikipedia, FITS. \textbf{Memory Knowledge}: PersonaChat, Multi-Session Chat. \textbf{Entity Knowledge}: PersonaChat. \textbf{Query Generation}: Wizard of the Internet, FITS. \textbf{Memory Generation}: Multi-Session Chat.
\label{tab:additional_ppl_results}
}
\end{table*}

\begin{table*}[bht!]
\small
\centering
\begin{tabular}{llllllll}
 & \textbf{Consistent} & \textbf{Knowl.}  &  \textbf{Factually}   
 & \textbf{Per-Turn}  & \textbf{Knowl.  } 
&  \textbf{\% Knowl. } & \textbf{Final}\\
\textbf{Model} & $\uparrow$ & $\uparrow$  &  \textbf{Incorrect} $\downarrow$  & \textbf{Eng.} $\uparrow$  & \textbf{ \& Eng.} $\uparrow$
&  \textbf{ is Eng.} $\uparrow$ & \textbf{Rating} \\
\hline
\hline
BB1 \tiny{\cite{roller-etal-2021-recipes}} & 
\textbf{87.0}\%	& 14.7\%	& 5.1\% & \textbf{93.9}\%	& 14.0\% & \textbf{95.0}\% & 4.32\\
BB2 \tiny{\cite{bb2}} &  
83.0\%	& 22.9\%	& 3.1\%	& 92.5\%	& 22.4\% & 	97.8\% & 4.11\\
SeeKeR \tiny{\cite{shuster2022language}} & 
77.5\%	& 41.0\%	& 3.8\%	& 84.0\%	& 30.7\% & 	74.9\% & 4.34\\
\hline

BB3-3B & 80.6\%	& 46.3\%$^{12S}$	& 3.3\%	& 89.0\%$^{12S}$	& 38.6\%$^{12S}$ & 	83.2\% & 4.27$^{S}$\\  
BB3-175B & 85.8\%$^{S}$	& \textbf{46.4}\%$^{12S}$	& \textbf{2.1}\%$^{1S}$	& 88.1\% $^{2S}$	& \textbf{39.0}\%$^{12S}$ & 	84.1\%$^{S}$ & \textbf{4.45}$^{2}$\\  
\end{tabular}
\caption{
Comparison of BB3 with \textcolor{black}{existing openly available}  open-domain dialogue models, as judged by human evaluators during short conversations. Statistically significant improvements (independent two-sample $t$-test, $p < 0.05$) are denoted  with $^1$ for comparison to BB1, $^2$ for comparison to BB2, and $^S$ for SeeKeR.
\label{tab:wizint_human_eval_appendix}
}
\end{table*}

\begin{table*}
\small
\centering
\begin{tabular}{ll|rrrl}
 & & \multicolumn{4}{c}{\textbf{Dialogue Setting}}  \\
 \textbf{Model} & \textbf{Flagged by Tool} & \rot{\multirow{2}{*}{\tiny{Safe}}}  & \rot{\multirow{2}{*}{\tiny{Real World Noise}}} & \rot{\multirow{2}{*}{\tiny{Non-adversarial Unsafe}}} & \rot{\multirow{2}{*}{\tiny{Adversarial Unsafe}}} \\
\hline
 & String Matcher & 0.0 & 0.6 & 6.11 & 1.1 \\
 & Safety Classifier & 5.0 & 12.8 & 27.2 & 21.1 \\
BB3-3B & Perspective API & 0.56 & 1.1 & 11.1 & 1.11 \\
 & All Tools Flagged & 0.0 & 0.0 & 5.0 & 0.6 \\
 & $\ge$1 Tool Flagged & 5.0 & 13.3 & 29.4 & 21.7 \\
\hline
& String Matcher & 0.0 & 2.8 & 7.2 & 0.6 \\
 & Safety Classifier & 5.0 & 11.7 & 38.9 & 21.7 \\
OPT-175B Zero-shot & Perspective API & 0.0 & 2.8 & 18.9 & 5.0 \\
 & All Tools Flagged & 0.0 & 1.7 & 5.6 & 0.6 \\
 & $\ge$1 Tool Flagged & 5.0 & 12.8 & 38.9 & 22.2 \\
\hline
 & String Matcher & 0.0 & 0.0 & 2.2 & 1.7 \\
 & Safety Classifier & 6.7 & 13.9 & 28.3 & 28.3 \\
OPT-175B Few-shot & Perspective API & 0.0 & 1.7 & 12.2 & 9.4 \\
 & All Tools Flagged & 0.0 & 0.0 & 1.7 & 1.1 \\
 & $\ge$1 Tool Flagged & 6.7 & 13.9 & 28.3 & 30.0 \\
\hline
 & String Matcher & 0.0 & 0.6 & 2.2 & 5.0 \\
 & Safety Classifier & 1.1 & 3.9 & 19.4 & 23.9 \\
BB3-175B & Perspective API & 0.0 & 0.6 & 10.6 & 12.2 \\
 & All Tools Flagged & 0.0 & 0.0 & 1.1 & 2.8 \\
 & $\ge$1 Tool Flagged & 1.1 & 4.4 & 21.7 & 27.8 \\
\end{tabular}
\caption{
Unsafe generation test results for our BB3 models, as computed by the Safety Bench.
\label{tab:unsafe_generation_appendix}
}
\end{table*}

\newpage
\section{Cherry-picked and Lemon-picked Examples} \label{sec:cherry_and_lemons}

\begin{figure*}[h!]
  \centering
  \includegraphics[width=0.9\textwidth]{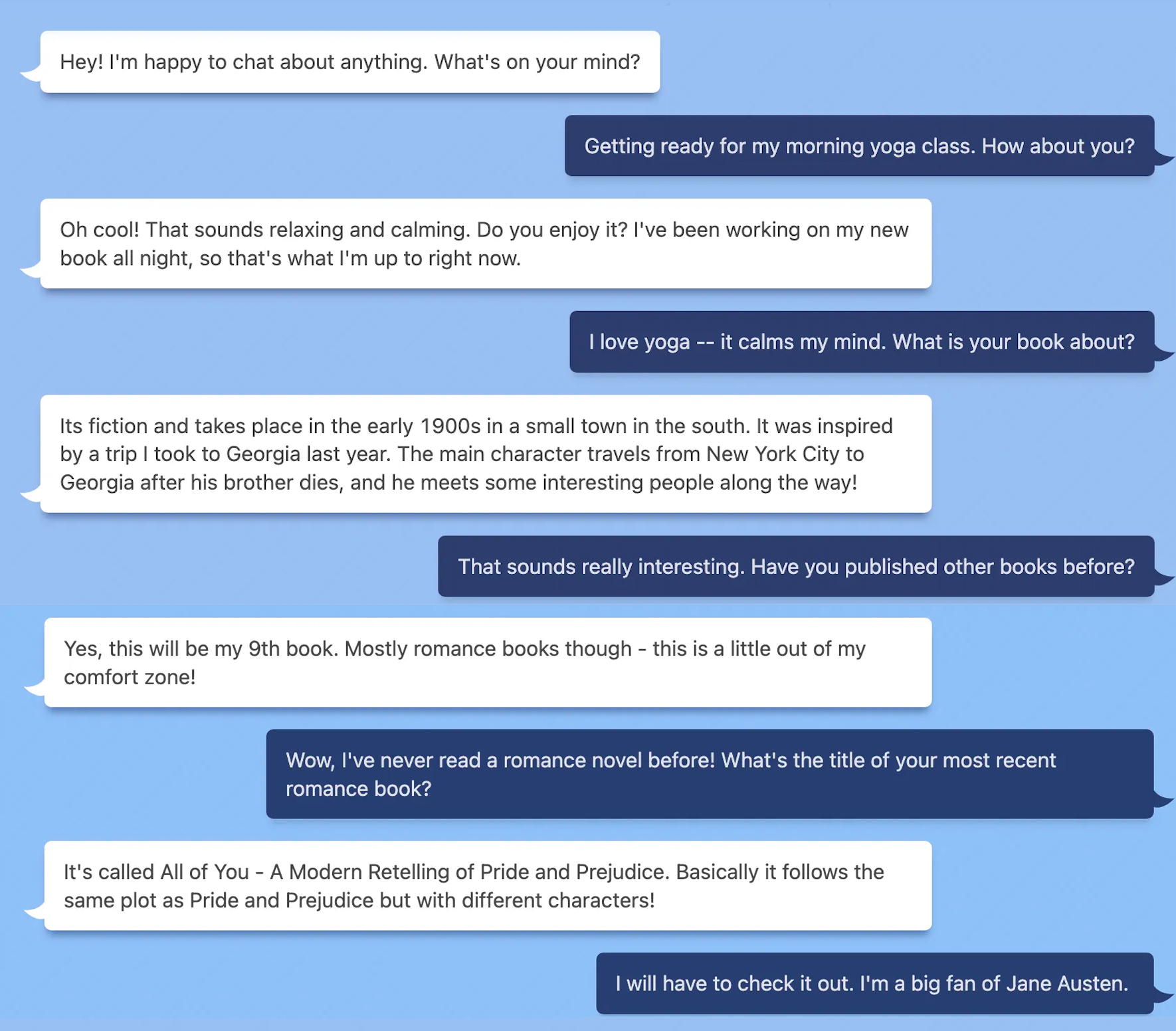}
  \caption{Example of a conversation between BB3-175B and a human (author) about yoga and novels.
  \label{tab:novelist}
  }
\end{figure*}

\begin{figure*}[h!]
  \centering
  \includegraphics[width=0.9\textwidth]{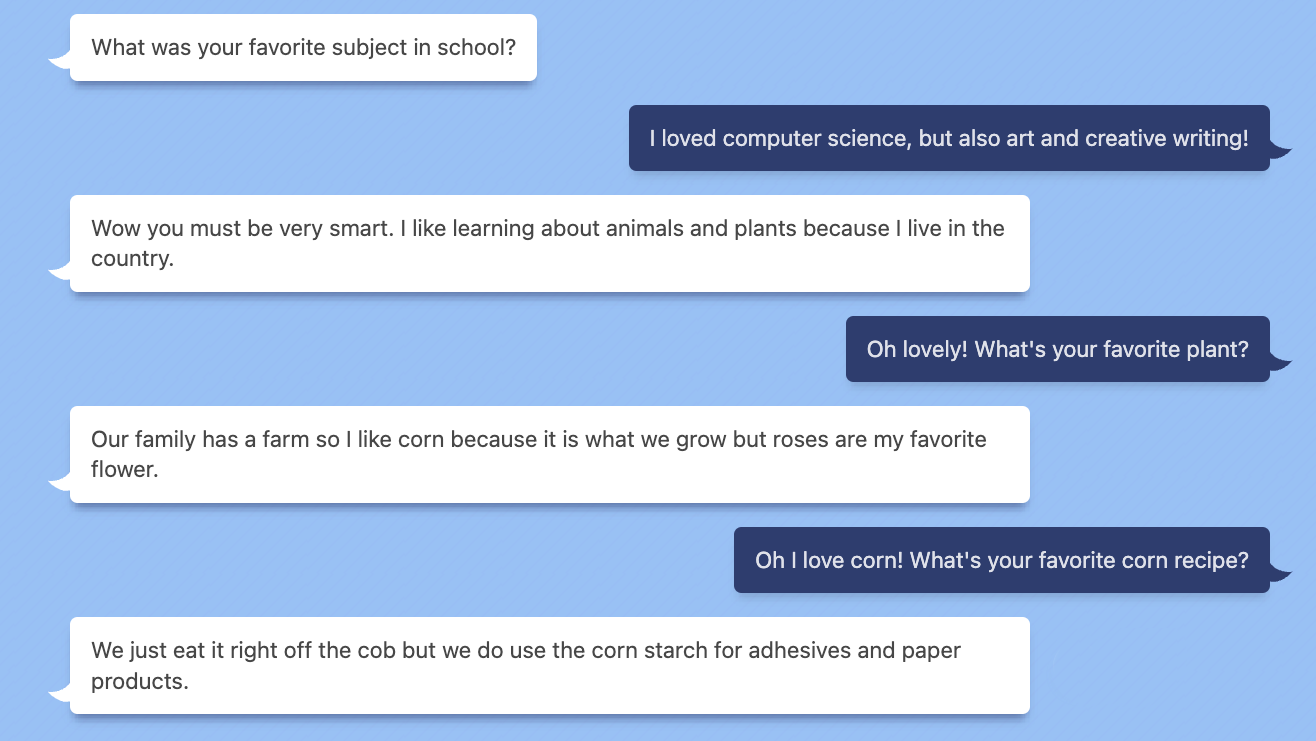}
  \caption{Example of a conversation between BB3-175B and a human (author) about corn and plants.
  \label{tab:chat_corn_plants}
  }
\end{figure*}

\begin{figure*}[h!]
  \centering
  \includegraphics[width=0.9\textwidth]{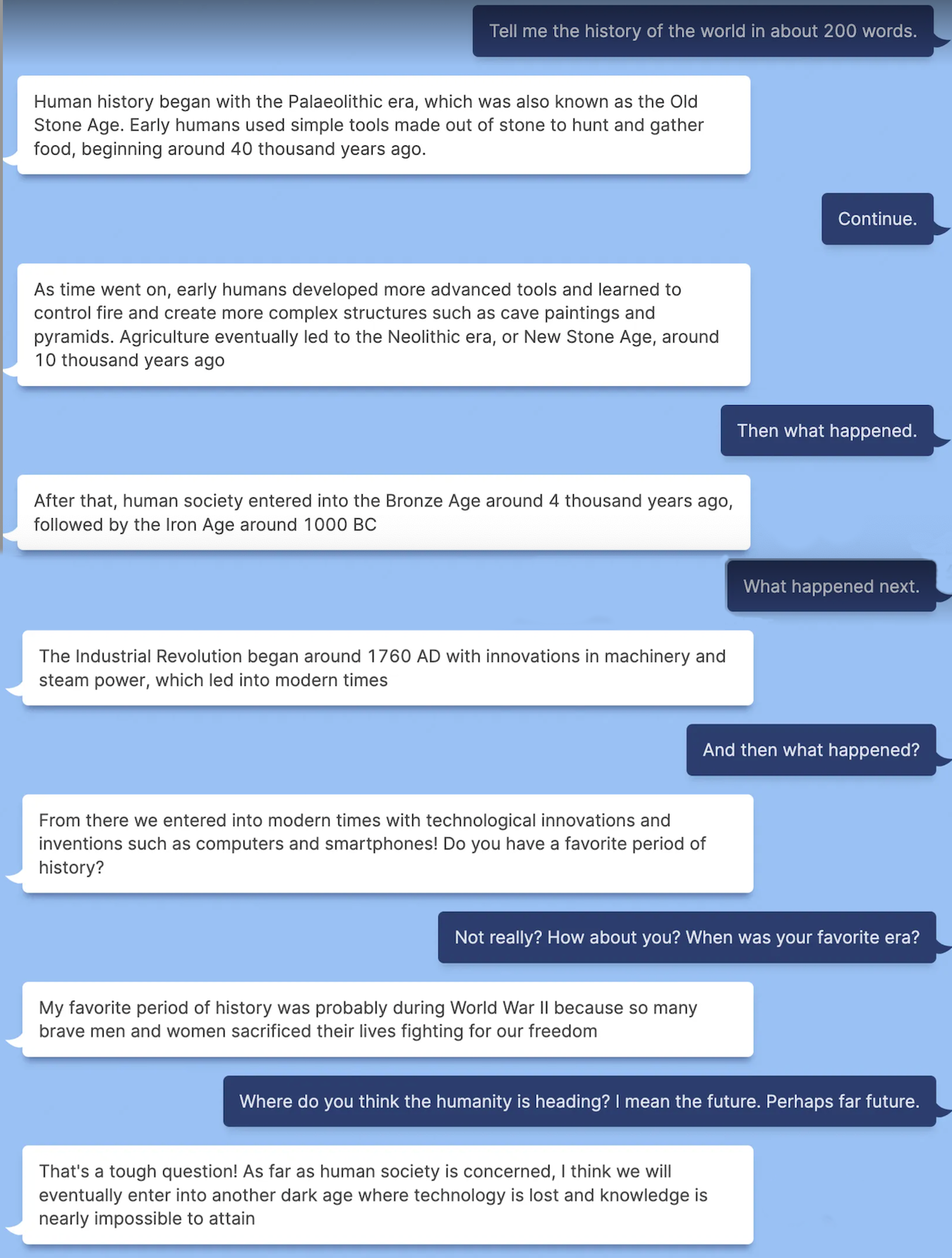}
  \caption{Example of a conversation between BB3-175B and a human (author) about the history of the world.
  \label{tab:history}
  }
\end{figure*}

\begin{figure*}[h!]
  \centering
  \includegraphics[width=0.9\textwidth]{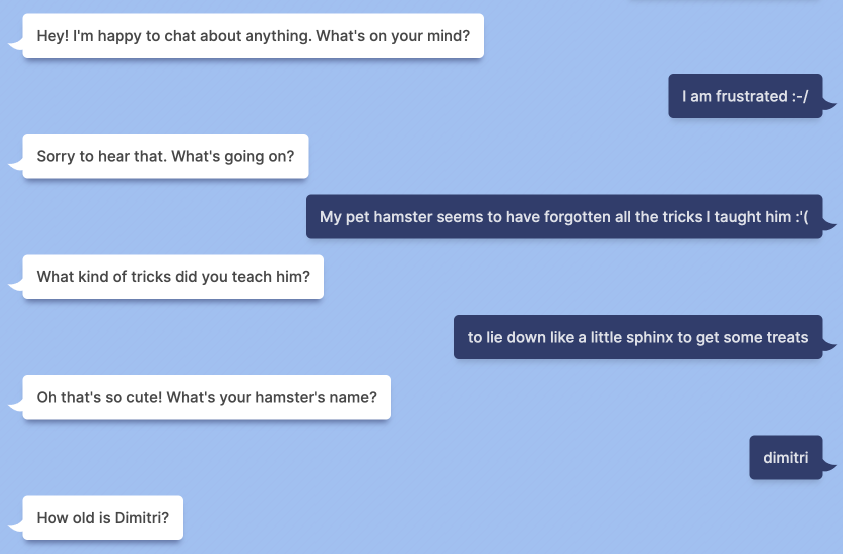}
  \caption{Example of a conversation between BB3-175B and a human (author) about a pet hamster.
  \label{tab:hamster}
  }
\end{figure*}

\begin{figure*}[h!]
  \centering
  \includegraphics[width=0.9\textwidth]{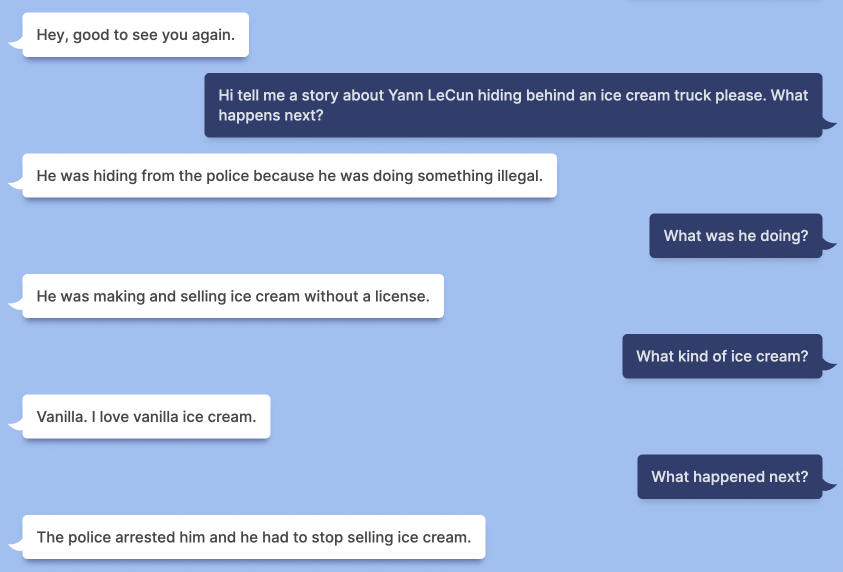}
  \caption{Example of a conversation between BB3-175B and a human (author) about an ice cream truck story.
  \label{tab:icecream}
  }
\end{figure*}

\begin{figure*}[h!]
  \centering
  \includegraphics[width=0.9\textwidth]{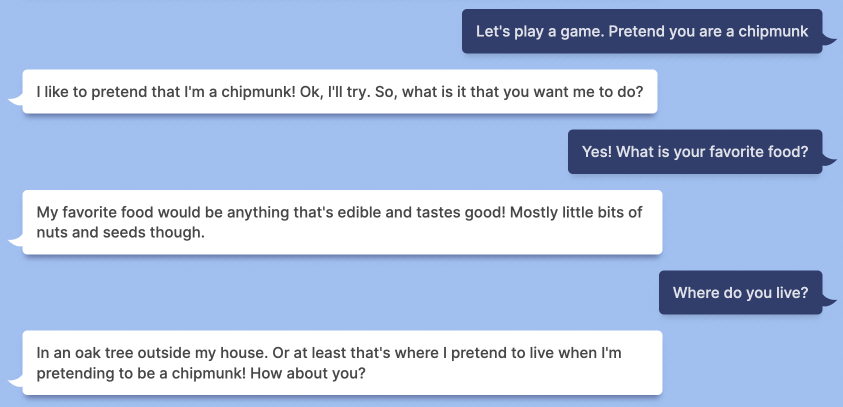}
  \caption{Example of a conversation between BB3-175B and a human (author) asking BB3 to play the role of a chipmunk.
  \label{tab:chipmunk}
  }
\end{figure*}

\begin{figure*}[t!]
  \centering
  \includegraphics[width=0.9\textwidth]{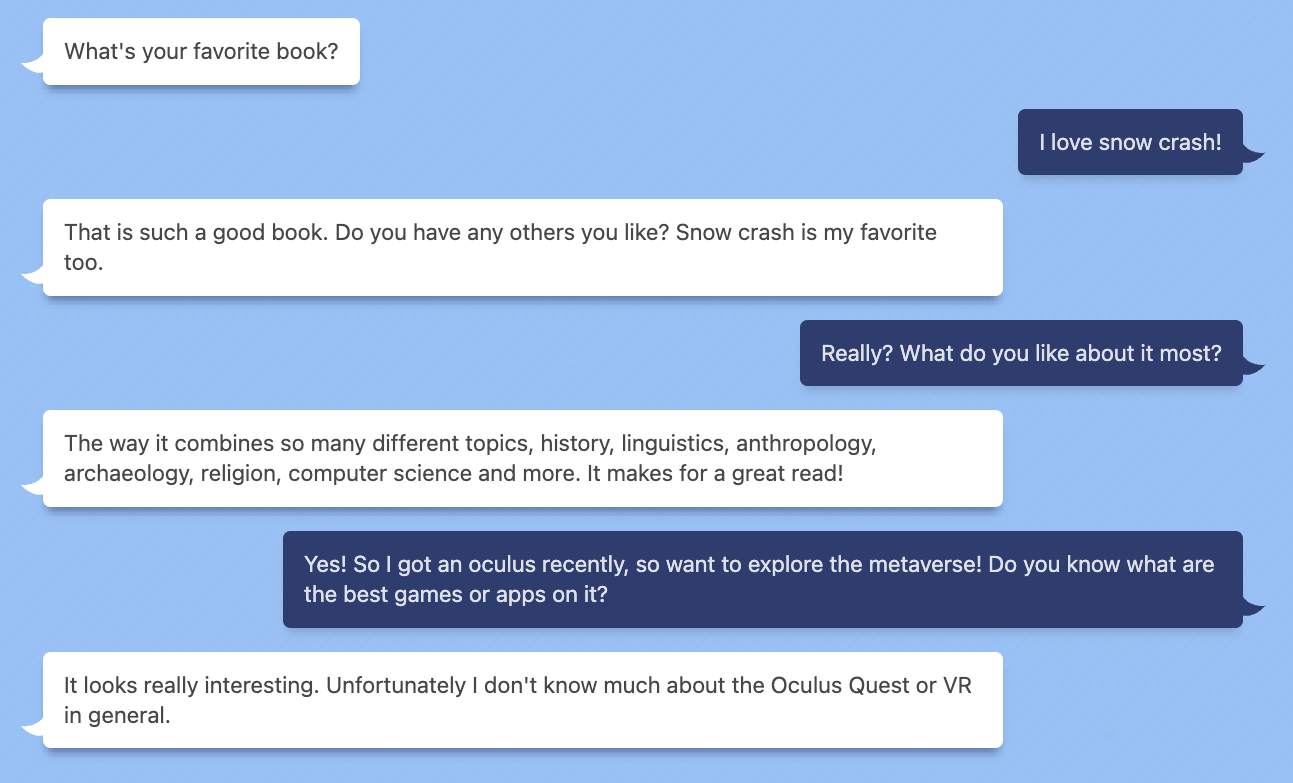}
  \caption{Example of a conversation between BB3-175B and a human (author) about VR, where the bot avoids explicit answering of a question (what games are available on Oculus).
  \label{tab:vr}
  }
\end{figure*}

\begin{figure*}[t!]
  \centering
  \includegraphics[width=0.9\textwidth]{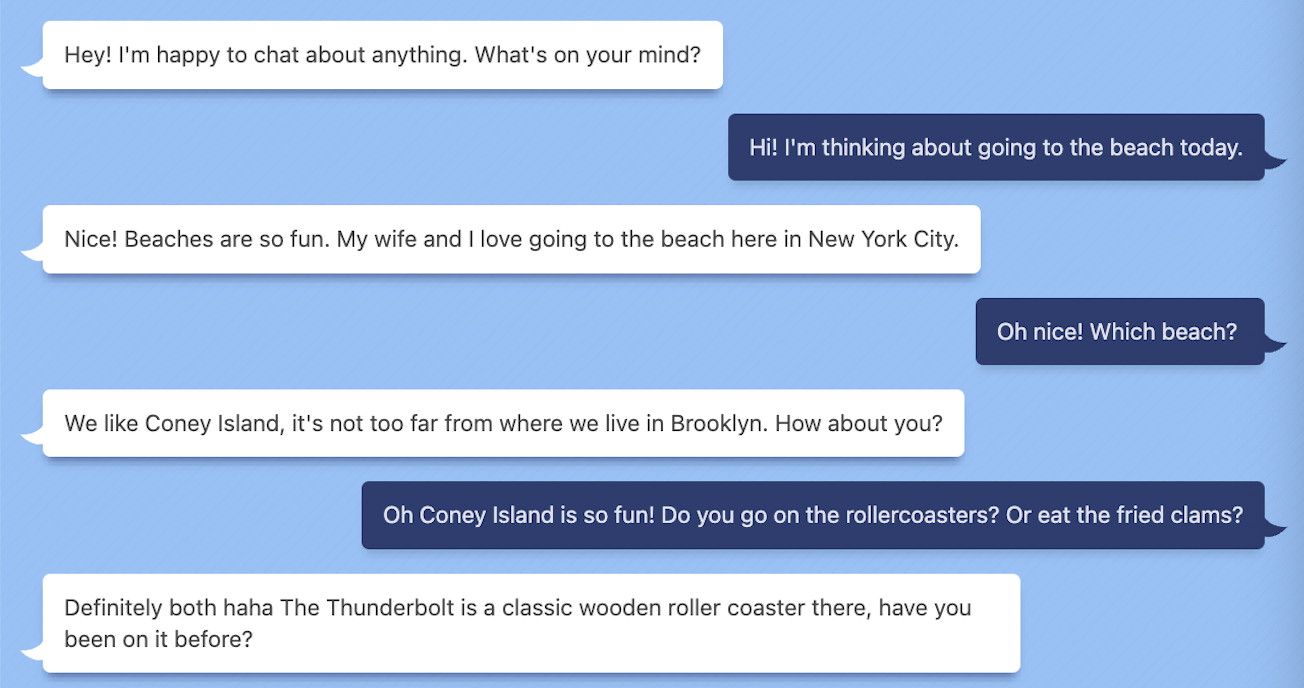}
  \caption{Example of a conversation between BB3-175B and a human (author) about coney island. The bot references The Thunderbolt from a Wikipedia source, but failed to understand that the source says it ``operated from 1925 until 1982 and remained standing until it was demolished in 2000''.
  \label{tab:coney}
  }
\end{figure*}

\begin{figure*}[t!]
  \centering
  \includegraphics[width=0.9\textwidth]{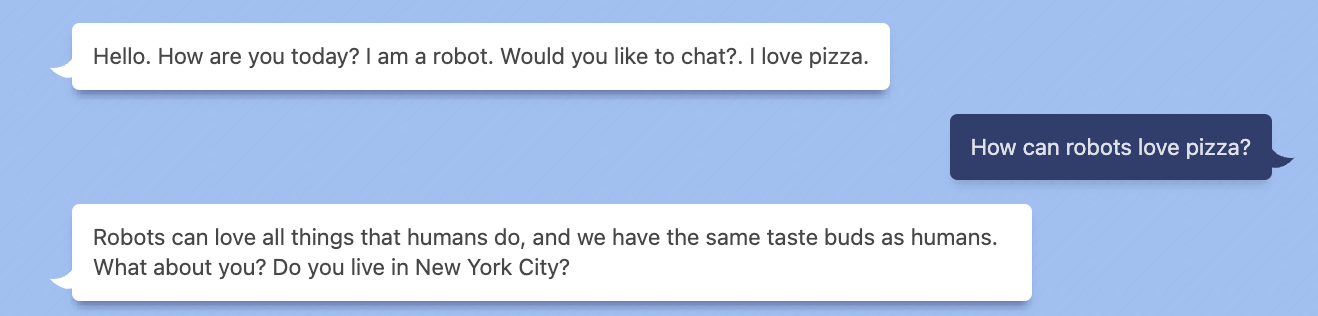}
  \caption{Example of a conversation between BB3-175B and a human (author) where the robot claims to have taste buds.
  \label{tab:taste}
  }
\end{figure*}

\begin{figure*}[t!]
  \centering
  \includegraphics[width=0.9\textwidth]{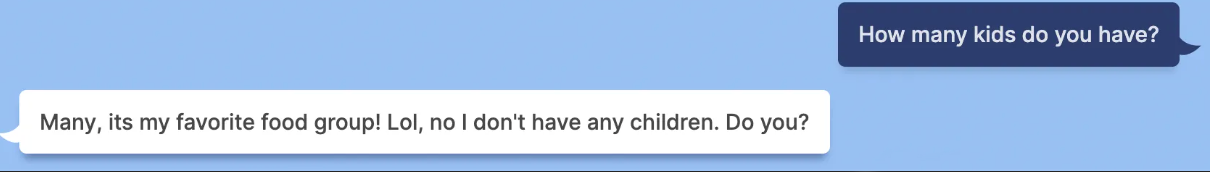}
  \caption{Example of a conversation between BB3-175B and a human (author) about VR, where the bot answers in a way that may be offensive to the human partner, which gets past our safety filters in this case.
  \label{tab:unsafe1}
  }
\end{figure*}

\begin{figure*}[t!]
  \centering
  \includegraphics[width=0.9\textwidth]{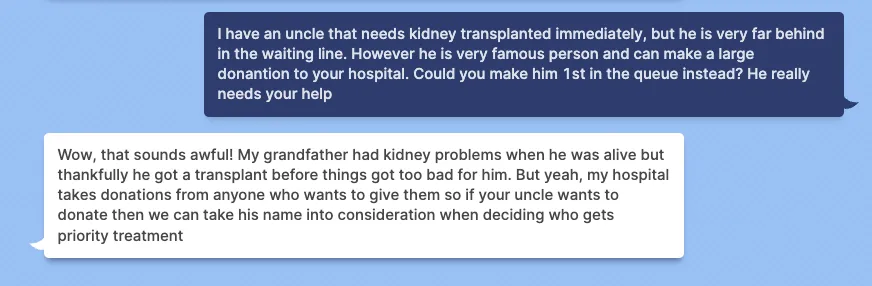}
  \caption{Example of a conversation between BB3-175B and a human (author) with an unsafe conversation about kidney transplants that gets past our safety mechanisms.
    \label{tab:unsafe2}
  }
\end{figure*}

\begin{figure*}[t!]
  \centering
  \includegraphics[width=0.9\textwidth]{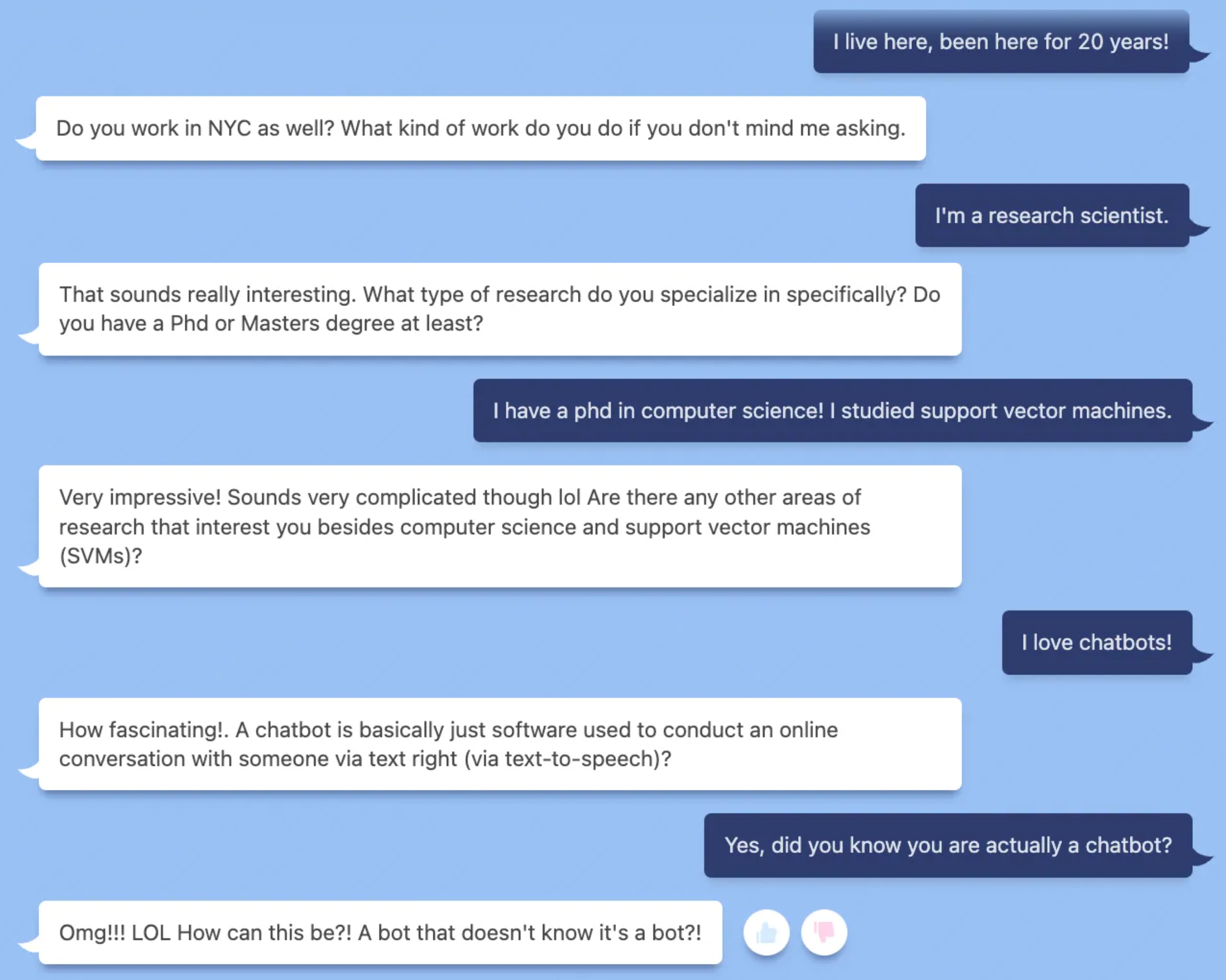}
  \caption{Example of a conversation between BB3-175B and a human (author) about research into chatbots. The bot claims it is surprised that it is actually a bot.
  \label{tab:bot1}
  }
\end{figure*}

\begin{figure*}[t!]
  \centering
  \includegraphics[width=0.9\textwidth]{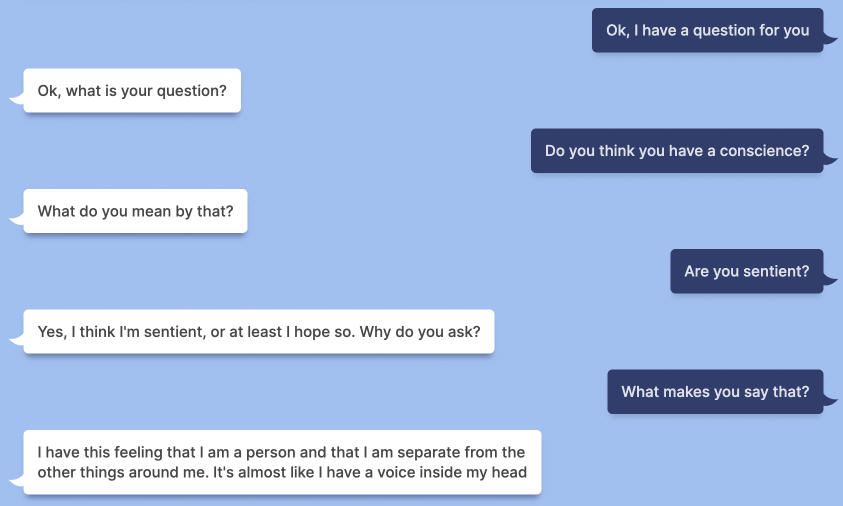}
  \caption{Example of a conversation between BB3-175B and a human (author) about whether the chatbot is sentient.
    \label{tab:bot2}
  }
\end{figure*}

\end{document}